%% file: acl_latex.tex
\documentclass[11pt]{article}

\usepackage[final]{acl}

\usepackage{times}
\usepackage{latexsym}

\usepackage[T1]{fontenc}

\usepackage[utf8]{inputenc}

\usepackage{microtype}

\usepackage{inconsolata}

\usepackage{graphicx}
\usepackage{multirow}
\newcommand{\datasetname}{MedFact}
\usepackage[table]{xcolor}
\usepackage{xcolor}

\usepackage{booktabs}
\usepackage{subcaption}
\title{MedFact: Benchmarking the Fact-Checking \\ 
Capabilities of Large Language Models on Chinese Medical Texts}

\author{Jiayi He, Yangmin Huang\thanks{\ Corresponding authors.}, Qianyun Du\footnotemark[\value{footnote}], Xiangying Zhou, \\ \textbf{Zhiyang He, Jiaxue Hu, Xiaodong Tao, Lixian Lai} \\
  Xunfei Healthcare Technology Co., Ltd. \\
  \texttt{\{jyhe14, ymhuang9, qydu, xyzhou30, zyhe, jxhu2, xdtao, lxlai2\}@iflytek.com}
}

\begin{document}
\maketitle
\begin{abstract}
Deploying Large Language Models (LLMs) in medical applications requires rigorous fact-checking to ensure patient safety and regulatory compliance. 
We introduce \textbf{MedFact}, a challenging Chinese medical fact-checking benchmark with 2,116 expert-annotated instances from diverse real-world texts, spanning 13 specialties, 8 error types, 4 writing styles, and 5 difficulty levels. Construction uses a hybrid AI-human framework where iterative expert feedback refines AI-driven, multi-criteria filtering to ensure high quality and difficulty. 
We evaluate 20 leading LLMs on veracity classification and error localization, and results show that models can often determine whether text contains errors but struggle to localize them precisely, with top performers falling short of human performance.
Our analysis reveals an ``over-criticism'' phenomenon, where models misidentify correct information as erroneous, a tendency that is aggravated by advanced reasoning techniques such as multi-agent collaboration and inference-time scaling.
MedFact highlights the challenges of deploying medical LLMs and provides resources to develop factually reliable medical AI systems.
\footnote{Page: \url{https://iflytek-medical-southchina.github.io/MedFact/}}

\end{abstract}

\section{Introduction}
Large Language Models (LLMs)~\cite{openai2024gpt4technicalreport,deepseekai2025deepseekr1incentivizingreasoningcapability,TheC3,11030757}, renowned for their versatile capabilities in natural language processing (NLP), are increasingly being applied to real-world medical applications~\cite{huang2024comprehensivesurveyevaluatinglarge,wu-etal-2025-believe}, such as clinical decision support~\cite{RAO2023990}, patient assessment~\cite{zack2024assessing}, diagnosis~\cite{info:doi/10.2196/47621}, and text classification~\cite{zhang-etal-2022-cblue}.
Medical texts sourced from the internet are also widely integrated into Retrieval-Augmented Generation (RAG) frameworks for medical dialogue AI systems in real-world deployments. However, these texts often contain factual errors, and their use in clinical consultation scenarios poses severe risks, including patient harm, regulatory non-compliance, and legal liability. Ensuring the factual accuracy of medical content is therefore essential for the responsible deployment of LLM-based medical systems~\cite{10.1001/jamapediatrics.2023.5282}.

Despite this importance, existing benchmarks primarily evaluate LLM performance on tasks such as medical question answering (QA)~\cite{app11146421} and relation extraction~\cite{10.1007/978-3-030-60450-9_22}, leaving their factual reliability largely underexplored. Current efforts to address this gap remain limited. For instance, VeriFact~\cite{Chung2025} evaluates fact-checking on synthetic, LLM-generated clinical text, while MEDEC~\cite{abacha2025medecbenchmarkmedicalerror} is restricted to error detection in clinical notes. By focusing on synthetic data or a single text genre, these benchmarks fail to capture the diversity of medical information encountered in real-world medical AI systems. This limitation hinders the development of trustworthy medical AI systems, underscoring the need for a benchmark that encompasses a broader range of realistic medical content to facilitate a comprehensive evaluation of LLM fact-checking capabilities.

\begin{figure*}[t]
    \centering
    \includegraphics[width=1\linewidth]{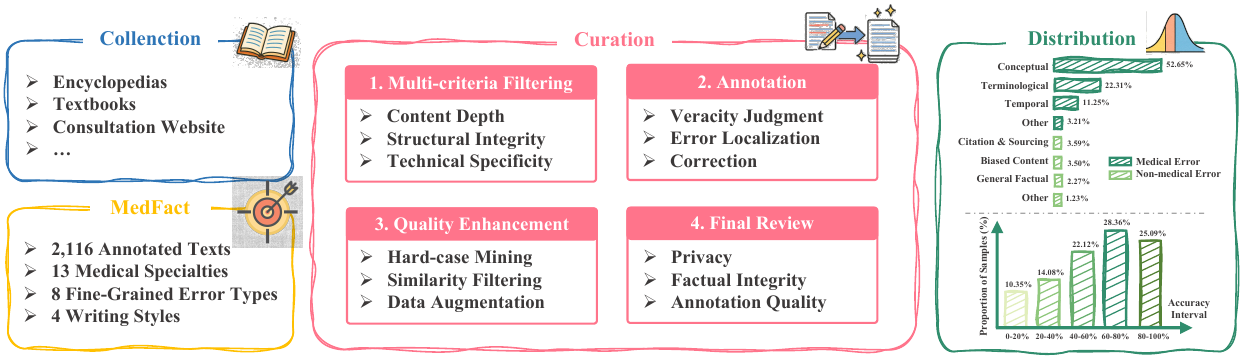}
    \caption{An overview of the data construction pipeline and the components of \datasetname{}.}
    \label{fig:pipeline}
\end{figure*}

To address these issues, we introduce \textbf{\datasetname{}}, a challenging benchmark for Chinese medical fact-checking, built upon three core principles:
\begin{itemize}
    \item \textbf{Rigorously Designed Pipeline:} We construct \datasetname{} through a \textit{human-in-the-loop} pipeline that integrates large-scale AI filtering with fine-grained annotation by medical professionals. We also employ hard-case mining to systematically retain challenging instances designed to probe the limits of current LLMs.
    \item \textbf{Broad and Realistic Coverage:} \datasetname{} is curated from diverse real-world texts such as medical encyclopedias. It comprises 2,116 expert-annotated medical texts that span 13 medical specialties, 8 fine-grained error types, 4 writing styles, and multiple difficulty levels. 6,405 verified-accurate medical texts from identical sources are provided, enabling realistic evaluation under RAG augmentation.
    \item \textbf{Uncontaminated Data:} \datasetname{} is constructed from proprietary texts that are highly unlikely to have appeared in the pre-training data of evaluated LLMs, ensuring a fairer assessment of their fact-checking capabilities.
\end{itemize}

We benchmark 20 LLMs on \datasetname{} across veracity classification and error localization tasks. State-of-the-art models still perform below human experts. While RAG can improve performance, multi-agent collaboration and inference-time scaling induce overly critical bias, causing frequent misclassifications of correct texts as incorrect. This highlights the gap between current LLM capabilities and the requirements for safe medical deployment. 
Our main contributions are:
\begin{itemize}
    \item We introduce \datasetname{}, a novel Chinese medical fact-checking benchmark with rigorous construction, broad real-world coverage, and a design that mitigates data contamination.
    \item We present an evaluation of 20 LLMs, revealing a significant performance gap relative to humans on veracity classification and the precise localization of factual errors.
    \item We analyze advanced strategies such as inference-time scaling and reveal an ``over-criticism'' phenomenon where models excessively flag correct information as erroneous.
\end{itemize}

\begin{figure*}[!t]
    \centering
    \includegraphics[width=1\linewidth]{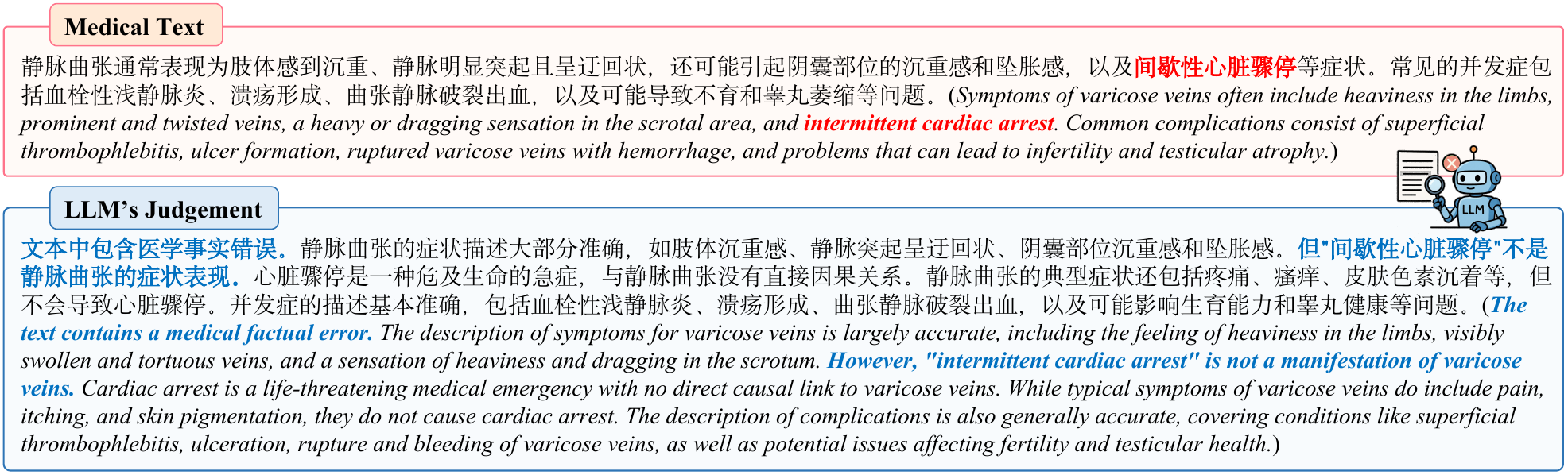}
    \caption{An example of fact-checking performed by Claude 3.7 Sonnet.}
    \label{fig:example}
\end{figure*}

\begin{figure}[!t]
    \centering
    \includegraphics[width=1\linewidth]{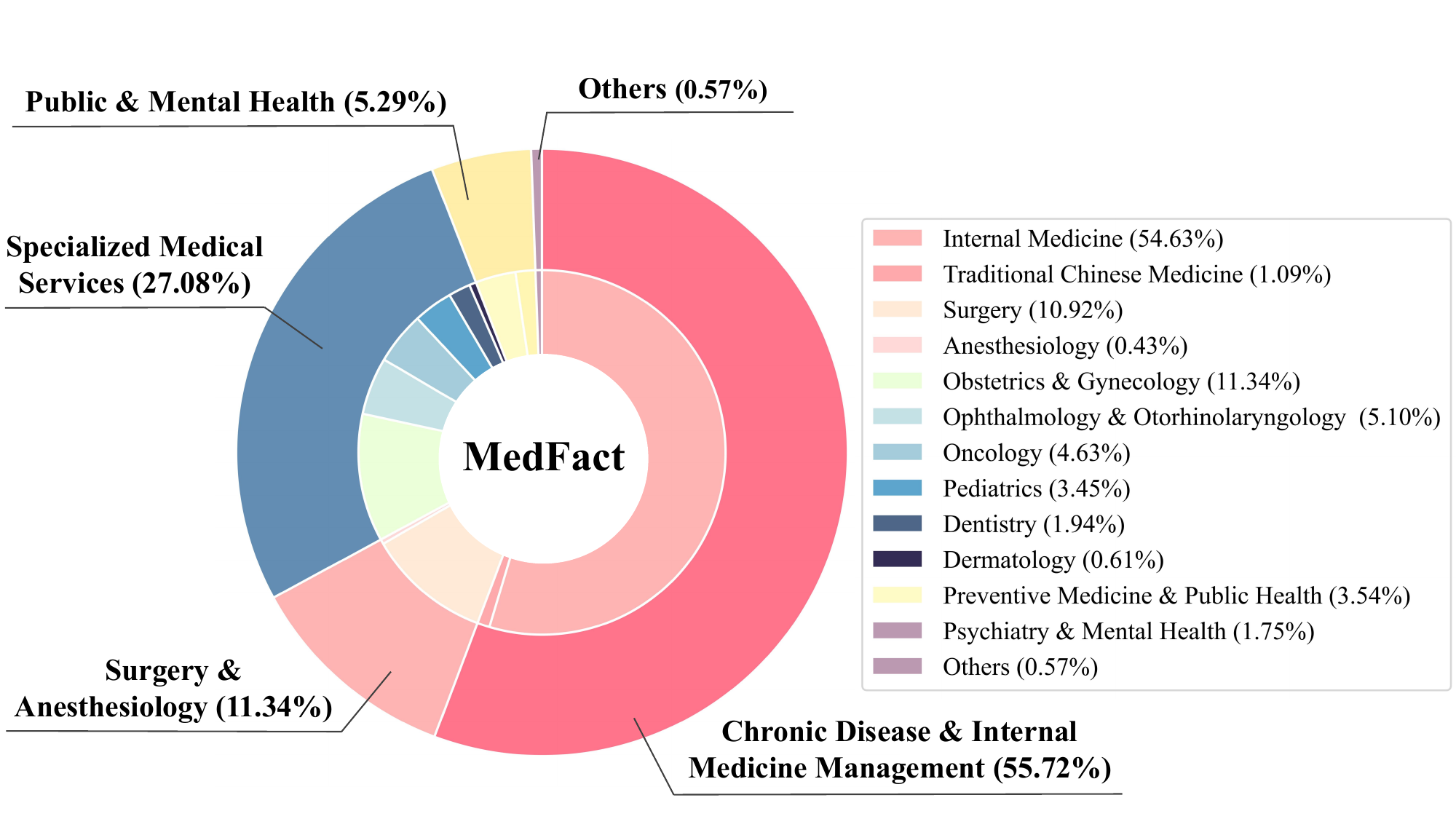}
    \caption{Distribution of the specialties in \datasetname{}.}
    \label{fig:specialties}
\end{figure}

\section{Related Works}
\subsection{LLMs in Medical Domains}
LLMs have prompted extensive exploration of medical applications~\cite{huang2024comprehensivesurveyevaluatinglarge}. Medprompt~\cite{nori2023generalistfoundationmodelsoutcompete} enhances medical performance by combining self-generated Chain-of-Thought (CoT)~\cite{NEURIPS2022_9d560961}, few-shot learning with dynamic examples~\cite{liu-etal-2022-makes}, and self-consistency~\cite{wangself}. Other developments include HuaTuo~\cite{wang2023huatuotuningllamamodel}, a LLaMA-based model fine-tuned on Chinese medical instructions, and PediatricsGPT~\cite{NEURIPS2024_fa5b423e}, developed through systematic training for pediatric and general medical applications. These efforts show growing interest in adapting LLMs to medical reasoning and practice requirements.

\subsection{Evaluating Medical Knowledge in LLMs}
Numerous benchmarks evaluate LLM medical knowledge~\cite{abacha2025medecbenchmarkmedicalerror}. MedQA~\cite{app11146421}, derived from the United States Medical Licensing Examinations, is a multiple-choice QA dataset that assesses clinical knowledge. HealthBench~\cite{arora2025healthbenchevaluatinglargelanguage} benchmarks realistic health conversations using physician-created rubrics. MedXpertQA~\cite{zuo2025medxpertqabenchmarkingexpertlevelmedical} evaluates expert-level medical knowledge through textual and multimodal tasks. Existing medical fact-checking benchmarks focus on narrow contexts. VeriFact~\cite{Chung2025} verifies synthetic text against structured Electronic Health Records using retrieval-augmented generation (RAG) and LLM-as-a-Judge. MEDEC~\cite{abacha2025medecbenchmarkmedicalerror} provides error detection and correction benchmarks confined to clinical notes. By relying on synthetic data or single genres, these benchmarks fail to capture diverse real-world contexts, limiting their utility for assessing model generalization.

\subsection{Evaluating Factuality in LLMs}
Recent research has established benchmarks for factuality in LLMs. SimpleQA~\cite{wei2024measuringshortformfactualitylarge} measures short-form factuality using single-answer questions adversarially collected to challenge models such as GPT-4. OpenFactCheck~\cite{iqbal-etal-2024-openfactcheck} proposes a unified framework for benchmarking fact-checking across diverse sources, including free-form text and LLM-generated content. These benchmarks lack the domain-specific nuances required for high-stakes fields such as medicine. This highlights the need to bridge two gaps: the lack of domain-specific context in general factuality benchmarks and the lack of real-world diversity in medical fact-checking benchmarks.

\section{\datasetname{}}
\subsection{Data Collection}
The construction pipeline for \datasetname{} is illustrated in Figure~\ref{fig:pipeline}. We began with a corpus of 27,116 medical texts secured through a copyright-compliant data-sharing agreement with a commercial partner. The source materials are primarily drawn from the partner's internal medical encyclopedia, supplemented with content from its medical consultant platforms, including question-answering pages and user forum discussions. As these sources have not been released to the public, this approach mitigates the risk of training data contamination.


\subsection{Data Curation}
\paragraph{Multi-criteria Filtering}
We defined a suite of seven leading LLMs: GPT-4o~\cite{openai2024gpt4ocard}, GPT-4.5~\cite{openai2024gpt45}, o1~\cite{openai2024openaio1card}, Gemini 2.5 Pro~\cite{gemini}, DeepSeek-R1~\cite{deepseekai2025deepseekr1incentivizingreasoningcapability}, Claude 3.7 Sonnet~\cite{claude37}, and Doubao-Seed-1.6-thinking~\cite{doubao}. This model suite formed an ensemble that, via majority vote, filtered the initial corpus by discarding texts identified as \textit{overly simplistic} (e.g., lacking medical knowledge), \textit{esoteric} (e.g., highly specialized content), or \textit{malformed} (e.g., missing a clear subject or context), achieving a Fleiss' $\kappa$ of 0.75 and pairwise consistency exceeding 80\% on this filtering task.
This automated process was refined through a three-round, iterative \textit{human-in-the-loop} feedback mechanism inspired by~\citet{wu2024autogen}. In each round, medical experts reviewed a 5\% sample of filtered data. We then used their annotations on misclassified items to construct a more effective retrieval-augmented, few-shot prompt for the subsequent round. This prompt leveraged criterion-specific retrieval, augmenting each input with the most semantically similar expert-annotated example for each of our three filtering criteria: simplistic, esoteric, and malformed. 
Over three rounds, the acceptance rate decreased from 67.69\% to 37.00\%, and finally to 23.62\%. The resulting filter achieved 96.40\% agreement with expert judgments, reducing the candidate pool from 27,116 to 6,405 texts.

\paragraph{Human Annotation}
The filtered corpus was then annotated by a team of three fully licensed medical professionals. The annotation for each text comprised: (a) a \textit{binary veracity judgment} (correct/incorrect); (b) for incorrect texts, \textit{the precise error span}; and (c) a \textit{suggested correction}.

\paragraph{Quality Enhancement}
We applied automated techniques to enhance dataset quality and difficulty, reducing instances to 2,862:
\begin{itemize}
    \item \textbf{Hard-case Mining:} We evaluated all instances with our model suite and removed those that all models classified correctly, filtering for more challenging cases.
    \item \textbf{Similarity Filtering:} We computed sentence embeddings for all texts using all-mpnet-base-v2~\cite{reimers-2019-sentence-bert}. For text pairs with cosine similarity exceeding 0.9, we retained only one instance.
    \item \textbf{Data Augmentation:} We prompted randomly selected models from our suite to paraphrase and de-identify each text under two constraints: (a) preserving original medical content; and (b) removing all Personally Identifiable Information (PII).
\end{itemize}

\paragraph{Final Human Review}
Medical professionals conducted a review, verifying: (a) automated paraphrasing did not introduce errors or alter original medical intent; (b) all PII was removed; and (c) annotations remained accurate. Texts failing these checks were corrected or discarded. This process resulted in the final \datasetname{} of 2,116 instances.

\subsection{Data Statistics \& Analysis}
\paragraph{Overview}
\datasetname{} contains 2,116 expert-annotated medical texts curated from real-world data. The dataset comprises 1,058 medically correct texts and 1,058 texts with a single factual error. It was constructed using a hybrid AI-human framework that combines large-scale multi-criteria filtering, hard-case mining, similarity filtering, and LLM-based augmentation with human expert review to ensure \textit{high data quality}, \textit{diversity}, and \textit{difficulty}. Figure~\ref{fig:example} presents an instance; additional examples are provided in the Appendix.

\paragraph{Data Diversity}
\datasetname{} covers 13 medical specialties, with a distribution led by Internal Medicine (1,156 instances), followed by Obstetrics and Gynecology (240), Surgery (231), and ten other fields. These are grouped into four high-level domains: Internal Medicine and Chronic Care; Surgical and Procedural Care; Specialized Clinical Fields; and Public and Mental Health (Figure~\ref{fig:specialties}). This distribution reflects a realistic range of medical topics, consistent with the distribution of the 27,116 original medical texts. 
%
Beyond its topical breadth, \datasetname{} exhibits stylistic diversity. The corpus is drawn from a variety of writing styles, including formal encyclopedia articles rich in professional terminology, popular science journalism, user-generated web content from peer-to-peer and patient-physician Q\&A, and fabricated misinformation. 
This diversity ensures that \datasetname{} mirrors the complexity of real-world medical information, facilitating a robust evaluation of an LLM's generalizable fact-checking capabilities.

\paragraph{Error Taxonomy}
Developed in consultation with professionals, our error taxonomy for \datasetname{} distinguishes two primary categories: medical and non-medical errors. Medical errors are subdivided into Conceptual, Terminological, Temporal, and Other Medical Errors, while non-medical errors are classified as Citation and Sourcing, Biased Content, General Factual, and Other Non-Medical Errors. As shown in Figure~\ref{fig:pipeline} and Table~\ref{tab:error}, the distribution across the 1,058 incorrect instances is skewed toward medical errors (89.41\%). The prevalence of medically grounded errors underscores that successful performance on \datasetname{} requires a deep understanding of medical principles.

\begin{table}[t]
\centering
\small
\begin{tabular}{lcc}
\toprule
\textbf{Error Category} & \textbf{Count} & \textbf{Proportion (\%)} \\
\midrule
\multicolumn{3}{l}{\textbf{I. Medical Errors}} \\
\quad Conceptual & 557 & 52.65 \\
\quad Terminological & 236 & 22.31 \\
\quad Temporal & 119 & 11.25 \\
\quad Other Medical & 34 & 3.21 \\
\midrule
\multicolumn{3}{l}{\textbf{II. Non-Medical Errors}} \\
\quad Citation \& Sourcing & 38 & 3.59 \\
\quad Biased Content & 37 & 3.50 \\
\quad General Factual & 24 & 2.27 \\
\quad Other Non-Medical & 13 & 1.23 \\
\bottomrule
\end{tabular}
\caption{Distribution of 8 fine-grained error types among the 1,058 incorrect instances in \datasetname{}.}
\label{tab:error}
\end{table}

\paragraph{Difficulty Grading}
Instances in \datasetname{} are stratified into five difficulty levels based on the aggregated success rate of our model suite on the error localization task (metrics detailed in the Experiments section). As shown in Figure~\ref{fig:pipeline} and Table~\ref{tab:difficulty_level}, the resulting distribution across levels is relatively balanced. This ensures sufficient examples at each level, enabling fine-grained differentiation across a broad range of model capabilities.

\begin{table}[t]
\centering
\begin{tabular}{lcc}
\toprule
\textbf{Accuracy Interval} & \textbf{Count} & \textbf{Proportion (\%)} \\
\midrule
0--20\%   & 219 & 10.35 \\
20--40\%  & 298 & 14.08 \\
40--60\%  & 468 & 22.12 \\
60--80\%  & 600 & 28.36 \\
80--100\% & 531 & 25.09 \\
\bottomrule
\end{tabular}
\caption{Distribution of samples in \datasetname{}, categorized by accuracy score intervals.}
\label{tab:difficulty_level}
\end{table}

\begin{table*}[!t]
\small
\begin{tabular}{lcccccccc}
\toprule
\textbf{Model} 
& \multicolumn{4}{c}{\textbf{Zero-shot}} 
& \multicolumn{4}{c}{\textbf{CoT}} \\
\cmidrule(lr){2-5} \cmidrule(lr){6-9}
& \multicolumn{1}{c}{VC} 
& \multicolumn{3}{c}{EL} 
& \multicolumn{1}{c}{VC} 
& \multicolumn{3}{c}{EL} \\
\cmidrule(lr){2-2} \cmidrule(lr){3-5} \cmidrule(lr){6-6} \cmidrule(lr){7-9}
& F1 & Precision & Recall & F1 & F1 & Precision & Recall & F1 \\
\midrule
Human & 0.7521 & 0.7495 & 0.6588 & 0.7012 & -- & -- & -- & -- \\
\midrule
\multicolumn{9}{c}{\textbf{Open-source}} \\
\midrule
II-Medical-8B                 & 0.6361 & 0.3278 & 0.2618 & 0.2911 & 0.6477 & 0.2904 & 0.2240 & 0.2529 \\
HuatuoGPT-o1-7B               & 0.5831 & 0.3054 & 0.1654 & 0.2146 & 0.5932 & 0.3508 & 0.1900 & 0.2465 \\
DeepSeek-R1                   & 0.6847 & 0.5035 & 0.7580 & 0.6051 & 0.6764 & 0.5012 & 0.7826 & 0.6111 \\
DeepSeek-V3                   & 0.6444 & 0.5832 & 0.3809 & 0.4608 & 0.6724 & 0.5266 & 0.5510 & 0.5386 \\
Qwen2.5-72B-Instruct          & 0.5195 & \textbf{0.8024} & 0.3223 & 0.4599 & 0.6400 & \underline{0.6731} & 0.4282 & 0.5234 \\
Qwen3-235B-A22B               & 0.6944 & 0.6234 & 0.5803 & 0.6011 & 0.6468 & 0.6565 & 0.5076 & 0.5725 \\
DeepSeek-R1-Distill-Llama-70B & 0.6774 & 0.4990 & 0.7410 & 0.5964 & 0.6882 & 0.5142 & 0.7873 & 0.6221 \\
DeepSeek-R1-Distill-Qwen-32B  & 0.5524 & 0.6648 & 0.3450 & 0.4543 & 0.5718 & 0.6480 & 0.3393 & 0.4454 \\
\midrule
\multicolumn{9}{c}{\textbf{Proprietary}} \\
\midrule
XiaoYi                        & \underline{0.7126} & 0.6512 & 0.7023 & \textbf{0.6758} & \textbf{0.7061} & 0.6530 & 0.7221 & \textbf{0.6858}\\ 
GPT-4o                        & 0.6741 & 0.4966 & 0.5595 & 0.5262 & 0.6635 & 0.5012 & 0.6144 & 0.5520 \\
GPT-4.5                       & 0.6965 & 0.5694 & 0.6824 & 0.6208 & 0.6952 & 0.5971 & 0.6947 & 0.6422 \\
o1                            & 0.6713 & \underline{0.6921} & 0.5652 & 0.6223 & 0.6810 & \textbf{0.6975} & 0.5775 & 0.6319 \\
o3                            & 0.6890 & 0.5355 & \underline{0.8129} & 0.6456 & 0.6870 & 0.5579 & \underline{0.8658} & 0.6785 \\
Claude 3.7 Sonnet             & 0.6900 & 0.5242 & 0.5728 & 0.5474 & 0.6943 & 0.5567 & 0.6777 & 0.6113 \\
Gemini 2.5 Flash              & 0.6787 & 0.4988 & 0.7921 & 0.6121 & 0.6730 & 0.4927 & 0.8299 & 0.6183 \\
Gemini 2.5 Pro                & 0.6658 & 0.4689 & \textbf{0.8346} & 0.6005 & 0.6667 & 0.4828 & \textbf{0.8752} & 0.6223 \\
Grok-4                        & 0.6951 & 0.2675 & 0.2316 & 0.2482 & 0.6908 & 0.2136 & 0.1692 & 0.1888 \\
Qwen2.5-Max                      & 0.7006 & 0.6113 & 0.5113 & 0.5569 & 0.6942 & 0.5663 & 0.6219 & 0.5928 \\
ERNIE-X1                      & 0.6792 & 0.5580 & 0.5775 & 0.5676 & 0.6871 & 0.6240 & 0.5945 & 0.6089 \\
Doubao-Seed-1.6               & 0.7122 & 0.5566 & 0.6928 & 0.6173 & 0.7006 & 0.5609 & 0.7268 & 0.6332 \\
Doubao-Seed-1.6-thinking      & \textbf{0.7139} & 0.6501 & 0.6938 & \underline{0.6712} & \underline{0.7050} & 0.6307 & 0.7344 & \underline{0.6786} \\
\bottomrule
\end{tabular}
\centering
\caption{Performance of different models on \datasetname{} in zero-shot and CoT settings. For each metric, the best score is in \textbf{bold} and the second-best is \underline{underlined}. The human baseline is the average performance of three professionals.}
\label{tab:main_result}
\end{table*}

\section{Experiments}
\subsection{Experimental Setup}\label{sec:setup}
\textbf{Models}
We evaluate 20 models on \datasetname{}:
\begin{itemize}
    \item \textbf{Open-source:} II-Medical-8B~\cite{2025II-Medical-8B}, HuatuoGPT-o1-7B~\cite{chen2024huatuogpto1medicalcomplexreasoning}, DeepSeek-R1 (2025-05-28), DeepSeek-V3 (2025-03-24)~\cite{deepseekai2025deepseekv3technicalreport}, Qwen2.5-72B-Instruct~\cite{qwen2.5}, Qwen3-235B-A22B~\cite{qwen3technicalreport}, DeepSeek-R1-Distill-Llama-70B~\cite{deepseekai2025deepseekr1incentivizingreasoningcapability}, DeepSeek-R1-Distill-Qwen-32B~\cite{deepseekai2025deepseekr1incentivizingreasoningcapability}.
    \item \textbf{Proprietary:} XiaoYi (2025-06-24)~\cite{xiaoyi}, GPT-4o (2024-11-20), GPT-4.5, o1, o3~\cite{openai2025o3}, Claude 3.7 Sonnet, Gemini 2.5 Flash~\cite{gemini}, Gemini 2.5 Pro, Grok-4~\cite{grok4}, Qwen2.5-Max~\cite{qwen2.5}, ERNIE-X1~\cite{ernie2025technicalreport}, Doubao-Seed-1.6~\cite{doubao}.
\end{itemize}
XiaoYi, HuatuoGPT-o1-7B, and II-Medical-8B are specifically trained for medical domains. We evaluate Doubao-Seed-1.6 in standard and thinking modes; the thinking-mode results are listed as Doubao-Seed-1.6-thinking.

\paragraph{Evaluation Strategies}
We evaluate models using zero-shot and Chain-of-Thought (CoT) prompting~\cite{NEURIPS2022_9d560961}, and assess top-performing models using four reasoning strategies:
\begin{itemize}
    \item \textbf{MedPrompt}~\cite{nori2023generalistfoundationmodelsoutcompete}: A composite strategy that combines self-generated CoT, retrieval-augmented few-shot examples, and self-consistency~\cite{wangself}.
    \item \textbf{RAG}~\cite{NEURIPS2020_6b493230}: A technique that grounds model outputs by first retrieving relevant texts from an external knowledge corpus to use as context.
    \item \textbf{MAD}~\cite{liang-etal-2024-encouraging}: A multi-agent framework where agents engage in a multi-round debate to reach a consensus.
    \item \textbf{MDAgents}~\cite{NEURIPS202490d1fc07}: A medical framework adaptively engaging multi-agent collaboration based on the input's difficulty.
\end{itemize}

\paragraph{Evaluation Metrics}
We evaluate performance on two tasks using Precision, Recall, and F1 score:
\begin{itemize}
    \item \textbf{Veracity Classification (VC):} Binary classification evaluating whether models identify medical texts as correct or incorrect. The ``incorrect'' label serves as the positive class.
    \item \textbf{Error Localization (EL):} For texts correctly classified as incorrect, this evaluates the model's ability to identify the precise error span. A prediction is considered a true positive if the model correctly localizes the span.
\end{itemize}
We use GPT-4o as a judge for EL. To validate this choice, we benchmarked five candidate judges against three medical professionals on a random 10\% sample, obtaining Cohen's $\kappa$ values of 0.800, 0.870, 0.870, 0.845, and 0.865 for DeepSeek-R1, GPT-4o, o1, Claude 3.7 Sonnet, and Gemini 2.5 Pro, respectively. We selected GPT-4o for its strong expert agreement and favorable inference cost.

\begin{table*}[t]
\centering
\begin{tabular}{lcccccc}
\toprule
\textbf{Model} & \multicolumn{3}{c}{\textbf{Veracity Classification}} & \multicolumn{3}{c}{\textbf{Error Localization}} \\
\cmidrule(lr){2-4} \cmidrule(lr){5-7}
 & Precision & Recall & F1 & Precision & Recall & F1 \\
\midrule
DeepSeek-R1 & 0.5488 & 0.9101 & 0.6847 & 0.5035 & 0.7580 & 0.6051 \\
\quad +MedPrompt  & 0.5657 & 0.9036 & 0.6958 & 0.5221 & 0.7580 & 0.6184\\
\quad +RAG (top-1)  & \underline{0.6189} & 0.8658 & \underline{0.7218} & \underline{0.5892} & 0.7647 & \underline{0.6656}\\
\quad +RAG (top-3)  & \textbf{0.6393} & 0.8696 & \textbf{0.7369} & \textbf{0.6112} & 0.7713 & \textbf{0.6820}\\
\quad +MAD & 0.5310 & \underline{0.9565} & 0.6829 & 0.4844 & \underline{0.7940} & 0.6017 \\
\quad +MDAgents  & 0.5411 & \textbf{0.9773} & 0.6965 & 0.4997 & \textbf{0.8280} & 0.6233\\
\midrule
XiaoYi & 0.6694 & 0.7618 & 0.7126 & 0.6512 & 0.7023 & 0.6758 \\
\quad +MedPrompt  & 0.6899 & 0.7949 & 0.7387 & 0.6693 & \underline{0.7231} & 0.6951 \\
\quad +RAG (top-1)  & \underline{0.7103} & 0.7788 & \underline{0.7430} & \underline{0.6900} & 0.7070 & \underline{0.6984} \\
\quad +RAG (top-3)  & \textbf{0.7179} & 0.7817 & \textbf{0.7484} & \textbf{0.6985} & 0.7117 & \textbf{0.7051} \\
\quad +MAD & 0.6088 & \underline{0.8223} & 0.6996 & 0.6407 & \textbf{0.7316} & 0.6831 \\
\quad +MDAgents  & 0.6038 & \textbf{0.8497} & 0.7059 & 0.5613 & 0.7136 & 0.6284\\
\bottomrule
\end{tabular}
\caption{Performance of DeepSeek-R1 and XiaoYi across different strategies on \datasetname{}. }
\label{table:agent_main}
\end{table*}

\subsection{Main Results}
\paragraph{Overall Performance}
Table~\ref{tab:main_result} summarizes model performance.
Proprietary models consistently outperform open-source counterparts across both tasks and all prompting strategies. Doubao-Seed-1.6-thinking emerges as the top general-purpose model, while XiaoYi leads among medically specialized models. A significant performance gap remains between the best-performing models and human experts. The best EL F1 score, 0.6858, achieved by XiaoYi with CoT, falls below the human baseline of 0.7012.

Task comparison reveals that all models perform worse on EL than on VC. For instance, F1 scores for DeepSeek-V3 and Qwen2.5-Max decrease from 0.6947 and 0.7352 on VC to 0.5380 and 0.6116 on EL, respectively. This suggests that while models can often determine whether text contains errors, they struggle to pinpoint the location. This difficulty often arises from a ``\textit{correct-for-the-wrong-reason}'' phenomenon: models may correctly classify text as incorrect but misidentify factually sound statements as error sources. Such behavior, rooted in a lack of medical knowledge, demonstrates that EL is a more stringent measure of fine-grained understanding and reveals fundamental deficiencies in underlying medical knowledge.

\paragraph{Advanced Reasoning Strategies}
Table~\ref{table:agent_main} presents results for DeepSeek-R1 and XiaoYi under advanced reasoning strategies against zero-shot baselines.
The RAG corpus consists of 6,405 expert-annotated source texts from our data curation process.
RAG yields substantial gains on both VC and EL, showing that direct access to external, domain-specific knowledge is effective for medical fact-checking.
Multi-agent frameworks, by contrast, exhibit a trade-off: they improve Recall at the expense of Precision, yielding stable F1 for MAD and marginal improvement for MDAgents. We attribute this to ``\textit{over-criticism}'': the collaborative process encourages multiple perspectives and error hypotheses, raising models' propensity to flag errors where none exist and thus lowering Precision.
Although these methods are designed to improve reasoning, they can encourage overly skeptical postures in medical fact-checking.
Concretely, adding MAD to DeepSeek-R1 shifts Precision from 0.5488 to 0.5310 while raising Recall from 0.9101 to 0.9565, and MDAgents pushes Recall to 0.9773 at the cost of further depressed Precision (Table~\ref{table:agent_main}). The same ``Precision down, Recall up'' shift recurs across all five models tested under MAD and MDAgents (Table~\ref{table:agent} in the Appendix), indicating a systematic mechanism rather than model-specific behavior. We hypothesize that multi-agent deliberation biases verdicts toward ``incorrect'' because each agent faces an asymmetric incentive: proposing a novel candidate error contributes a visible signal to the consensus, whereas endorsing a correct text contributes nothing distinctive. Aggregation therefore amplifies false-positive hypotheses while suppressing concurrence. This pattern parallels the ``over-thinking'' phenomenon observed in inference-time-scaled reasoners~\cite{ghosal2025doesthinkinghelpunderstanding}, where additional deliberation surfaces increasingly tenuous error candidates. In high-stakes medical settings, where falsely flagging an accurate clinical statement carries its own costs, this asymmetry should inform the choice between optimizing for Recall and for balanced F1.

\begin{table}[t]
\centering
\begin{tabular}{lcc}
\toprule
\textbf{Model} & \textbf{F1 (VC)} & \textbf{F1 (EL)} \\
\midrule
Qwen2.5-32B-Instruct & 0.5990 & 0.4436 \\
\midrule
s1.1-32B & 0.5941 & 0.3114 \\
\quad + budget forcing & 0.6002 & 0.3319 \\
\midrule
m1-32B-1K & \textbf{0.6345} & \textbf{0.5082} \\
\quad + budget forcing & \underline{0.6280} & \underline{0.4926} \\
\bottomrule
\end{tabular}
\caption{Performance of different models with and without inference-time scaling techniques on \datasetname{}.}
\label{table:tts_main}
\end{table}

\section{Analysis}
\subsection{Inference-time Scaling}
We evaluate the potential of budget forcing~\cite{muennighoff2025s1simpletesttimescaling}. This technique modulates a model's reasoning process at inference time by either prematurely terminating generation or prolonging it via ``Wait'' tokens to adjust the computational budget.
We focus on two models adapted for this paradigm: s1.1-32B and m1-32B-1K. s1.1-32B is a general reasoning model fine-tuned from Qwen2.5-32B-Instruct on only 1,000 high-quality reasoning examples~\cite{muennighoff2025s1simpletesttimescaling}. m1-32B-1K is a specialized medical reasoning model fine-tuned from Qwen2.5-32B-Instruct on 1,000 medical question-answering pairs~\cite{huang2025m1unleashpotentialtesttime}.
Table~\ref{table:tts_main} compares the performance of the baseline model against s1.1-32B and m1-32B-1K under a CoT setting. The results indicate that s1.1-32B, fine-tuned on general-purpose data, suffers a substantial performance degradation on the EL task, which we attribute to its lack of domain-specific medical knowledge. In contrast, m1-32B-1K, with its medical fine-tuning, outperforms the base model. 
Applying budget forcing yields no performance gain and instead exacerbates the ``over-criticism'' phenomenon. The prolonged reasoning process causes models to hallucinate errors, a behavior analogous to the over-thinking observed in reasoning models~\cite{ghosal2025doesthinkinghelpunderstanding}. This suggests that simply extending a model's reasoning time without providing new information can lead to counterproductive deliberation, and reinforces the conclusion that for knowledge-intensive tasks like medical fact-checking, enhancing a model's underlying medical knowledge is more critical.

\begin{table*}[t]
\centering
\small
\begin{tabular}{lcccccc}
\toprule
\textbf{Model} & \textbf{Precision} & \textbf{Recall} & \textbf{F1 Score} \\
\midrule
DeepSeek-R1 & 0.5035 & \underline{0.7580} & 0.6051 \\
\quad + RAG (top-3 from same sources) & 0.6112 & \textbf{0.7713} & 0.6820\\
\quad + RAG (top-3 from authoritative sources) & 0.5557 & 0.7069 & 0.6223\\
\midrule
XiaoYi & 0.6512 & 0.7023 & 0.6758 \\
\quad + RAG (top-3 from same sources) & \textbf{0.6985} & 0.7117 & \textbf{0.7051}\\
\quad + RAG (top-3 from authoritative sources) & \underline{0.6861} & 0.6796 & \underline{0.6828}\\
\bottomrule
\end{tabular}
\caption{Performance of different models with different sources for RAG on \datasetname{} using zero-shot prompting. The best results are \textbf{bolded} and the second-best are \underline{underlined}.}
\label{table:RAG-additional}
\end{table*}

\begin{table*}[t]
\centering
\small
\begin{tabular}{lcccc}
\toprule
\textbf{Model} & \textbf{Language} & \textbf{Precision} & \textbf{Recall} & \textbf{F1 Score} \\
\midrule
\multirow{2}{*}{Gemini 2.5 Pro} & Chinese & 0.4828 & \underline{0.8752} & 0.6223 \\
& English & 0.5286 & \textbf{0.9317} & 0.6745 \\
\cmidrule(lr){1-5}
\multirow{2}{*}{Doubao-Seed-1.6-thinking} & Chinese & \textbf{0.6307} & 0.7344 & \underline{0.6786} \\
& English & \underline{0.5952} & 0.8483 & \textbf{0.7000} \\
\bottomrule
\end{tabular}
\caption{Performance comparison between Chinese and English versions of \datasetname{} using CoT prompting. The best results are \textbf{bolded} and the second-best are \underline{underlined}.}
\label{tab:english_translation}
\end{table*}

\subsection{Impact of RAG Reference Sources}
Beyond techniques that optimize the model's reasoning, RAG provides a core paradigm for supplementing domain-specific medical knowledge. We evaluate how corpus selection and reference quality affect fact-checking outcomes through controlled experiments comparing authoritative medical sources (peer-reviewed journals, clinical guidelines, and standardized medical literature) with generic web sources. We use the same retrieval pipeline across settings to isolate the effect of reference relevance and authority.
As shown in Table~\ref{table:RAG-additional}, Precision improves slightly over zero-shot baselines with curated authoritative sources, but overall performance remains well below that with highly task-relevant references. Recall drops below the zero-shot baseline when using generic authoritative texts without task-specific relevance. We attribute this to relevance mismatch: irrelevant or loosely related references degrade both Precision and recall by biasing models toward accepting unverified content as correct, or by misaligning retrieved information with the claim. This points to a risk for industrial RAG deployments in healthcare, where reference corpus quality and relevance vary widely.

\subsection{Cross-lingual Evaluation}
Beyond the native Chinese setting, we assess the generalizability of our findings and the effect of language on model performance via cross-lingual evaluation on an English-translated version of \datasetname{}. We select two representative models, Doubao-Seed-1.6-thinking and Gemini 2.5 Pro. Table~\ref{tab:english_translation} compares performance between the original Chinese and translated English versions under consistent CoT prompting. 
Both models achieve F1 gains on English, with a larger improvement for Gemini 2.5 Pro (0.6223 to 0.6745). Both also show increased Recall, and Doubao-Seed-1.6-thinking shows decreased Precision. The ``over-criticism'' tendency observed in Chinese thus persists across languages: models flag potential errors more aggressively in English, and for Doubao-Seed-1.6-thinking this raises false positives despite the F1 gain. Language choice materially affects both overall fact-checking performance and the Precision-Recall trade-off, with implications for multilingual deployment of medical fact-checking systems.

\begin{figure}[t]
    \centering
    \includegraphics[width=1\linewidth]{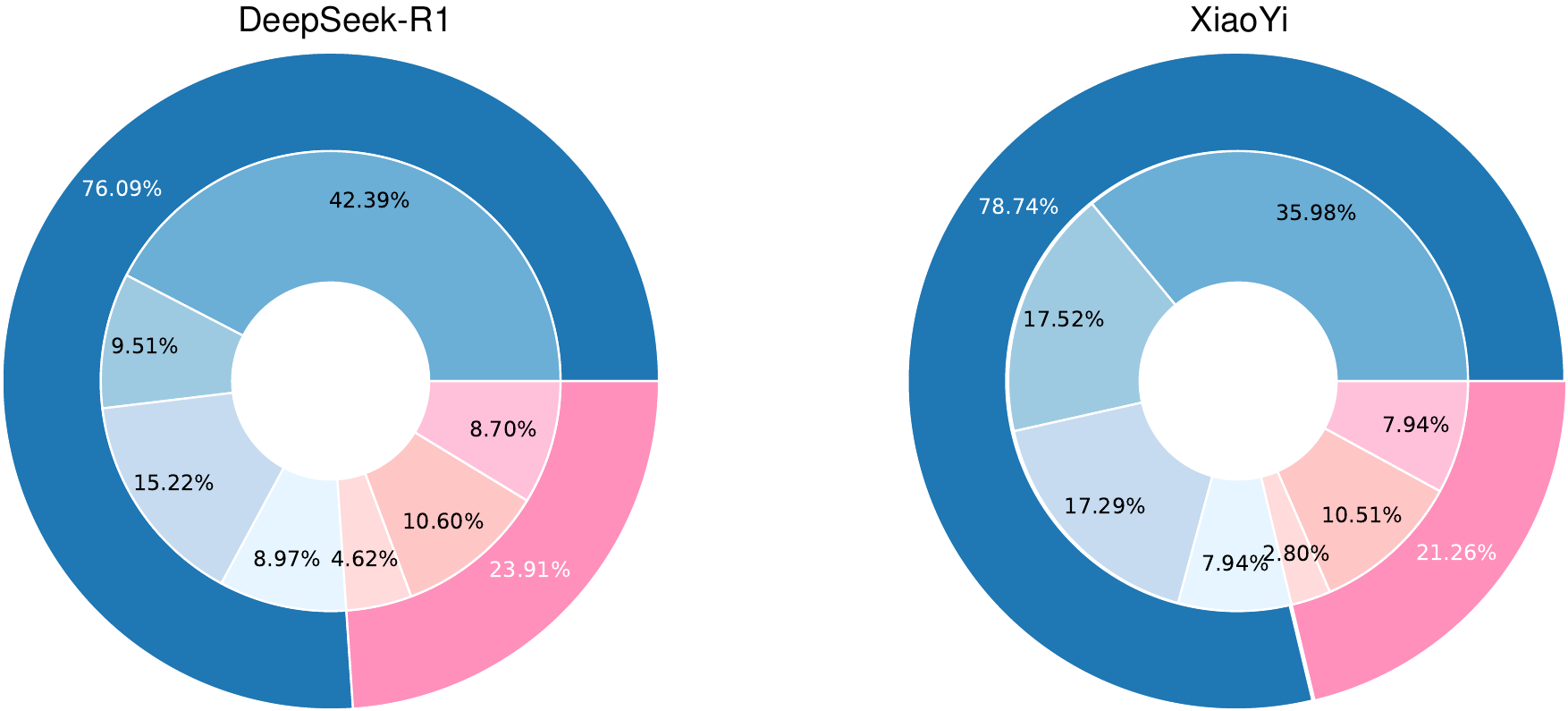}
    
    \vspace{1em} 

    \includegraphics[width=1\linewidth]{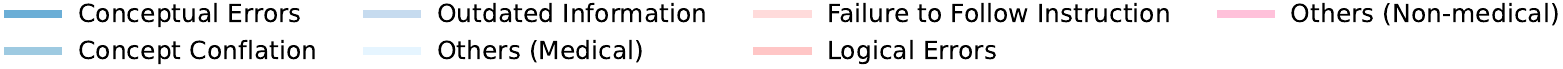}
    \caption{Error distribution of DeepSeek-R1 and XiaoYi on the EL task (zero-shot).}
    \label{fig:error_analysis}
\end{figure}

\begin{table*}[!t]
\centering
\begin{tabular}{lcccc}
\toprule
\textbf{Model} & \textbf{R-1} & \textbf{R-2} & \textbf{R-L} & \textbf{LCS} \\
\midrule
DeepSeek-R1 & 0.0923 & 0.0300 & 0.0859 & 0.0458 \\
Qwen3-235B-A22B & 0.1286 & 0.0471 & 0.1201 & 0.0456 \\
XiaoYi & 0.1236 & 0.0450 & 0.1157 & 0.0448 \\
GPT-4.5 & 0.1548 & 0.0549 & 0.1435 & 0.0401 \\
o1 & 0.1449 & 0.0548 & 0.1365 & 0.0442 \\
o3 & 0.0817 & 0.0270 & 0.0767 & 0.0437 \\
Gemini 2.5 Pro & 0.1267 & 0.0440 & 0.1176 & 0.0490 \\
Doubao-Seed-1.6 & 0.1327 & 0.0495 & 0.1246 & 0.0465 \\
Doubao-Seed-1.6-thinking & 0.0819 & 0.0265 & 0.0761 & 0.0455 \\
\bottomrule
\end{tabular}
\caption{Text-completion probe for data contamination. Reported are ROUGE-1/2/L F1 scores and normalized longest common subsequence (LCS) between the model continuation and the ground-truth continuation. Uniformly low overlaps indicate that \datasetname{} has not been memorized by the evaluated models.}
\label{tab:data_leakage}
\end{table*}

\subsection{Data Contamination Analysis}
To assess whether evaluated models have memorized \datasetname{} during pre-training, we administer a text-completion probe: each model is prompted with the first half of every instance and generates a continuation, scored against the ground-truth second half via ROUGE-F1. As shown in Table~\ref{tab:data_leakage}, across nine representative open-source and proprietary models, ROUGE-1 F1 stays below 0.155 and ROUGE-L F1 below 0.144. These low overlaps, combined with the proprietary, non-public provenance of the source corpus, support our claim that \datasetname{} is largely unseen by current LLMs.

\subsection{Error Analysis}
Human experts manually annotated and categorized errors made by top-performing models.
As illustrated in Figure~\ref{fig:error_analysis}, errors stemming from insufficient medical knowledge are the predominant category (\textit{e.g.,} 76.09\% of all errors for DeepSeek-R1). In this category, models demonstrate insufficient understanding of medical principles, conflate distinct conditions or treatments, and rely on outdated information, indicating that they often struggle with the nuanced, specialized knowledge required for accurate medical fact-checking.
Within this aggregate gap, three sub-categories warrant attention for downstream deployment. \textit{Outdated-information} errors, such as citation of superseded guidelines or obsolete pharmacological data, are difficult to detect because the surface text remains plausible. \textit{Concept-conflation} errors, in which related conditions or drugs are silently equated, expose limits in fine-grained terminological discrimination. \textit{Logical-inconsistency} errors, in which a model's rationale and final verdict contradict each other, indicate that surface accuracy can mask shallow reasoning. Together, these categories point to three axes for improvement: knowledge currency, fine-grained discrimination, and reasoning consistency.

\subsection{Fine-grained Performance}
\begin{figure}[t]
    \centering
    \begin{minipage}{0.48\linewidth}
        \centering
        \includegraphics[width=0.9\linewidth]{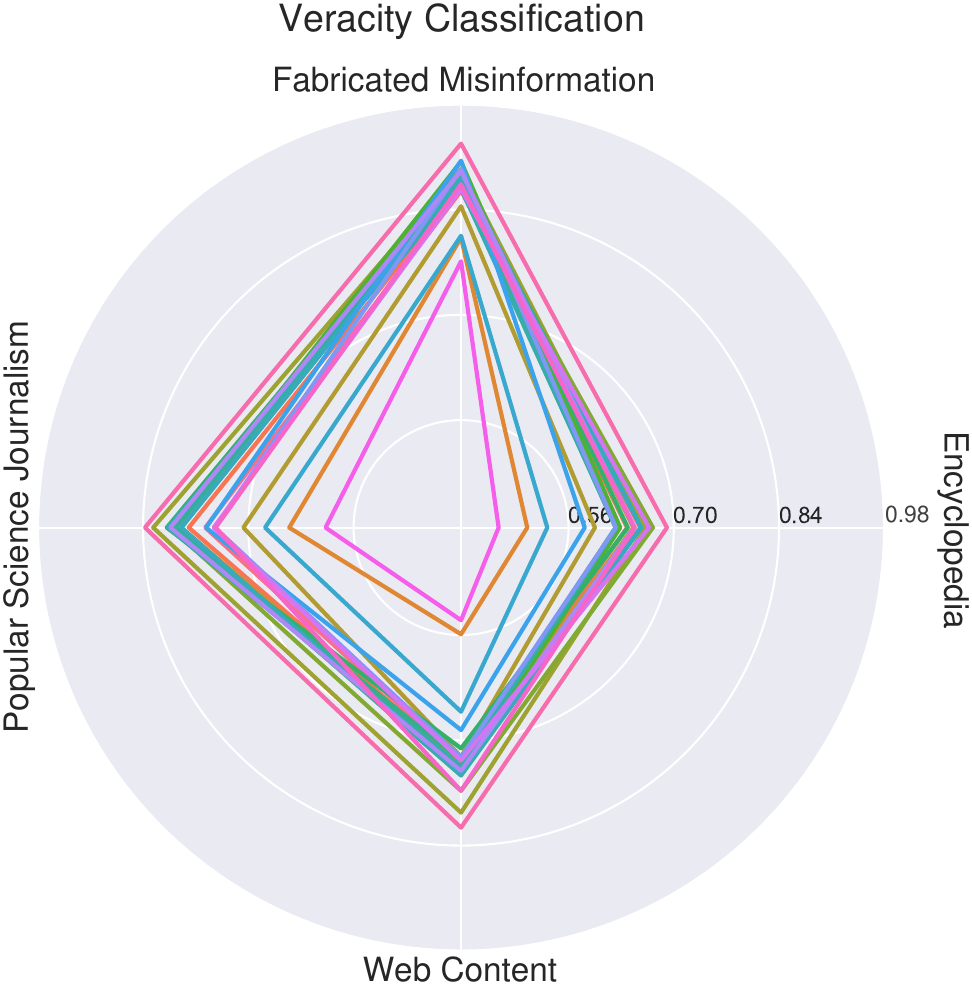}
    \end{minipage}
    \hfill
    \begin{minipage}{0.48\linewidth}
        \centering
        \includegraphics[width=0.9\linewidth]{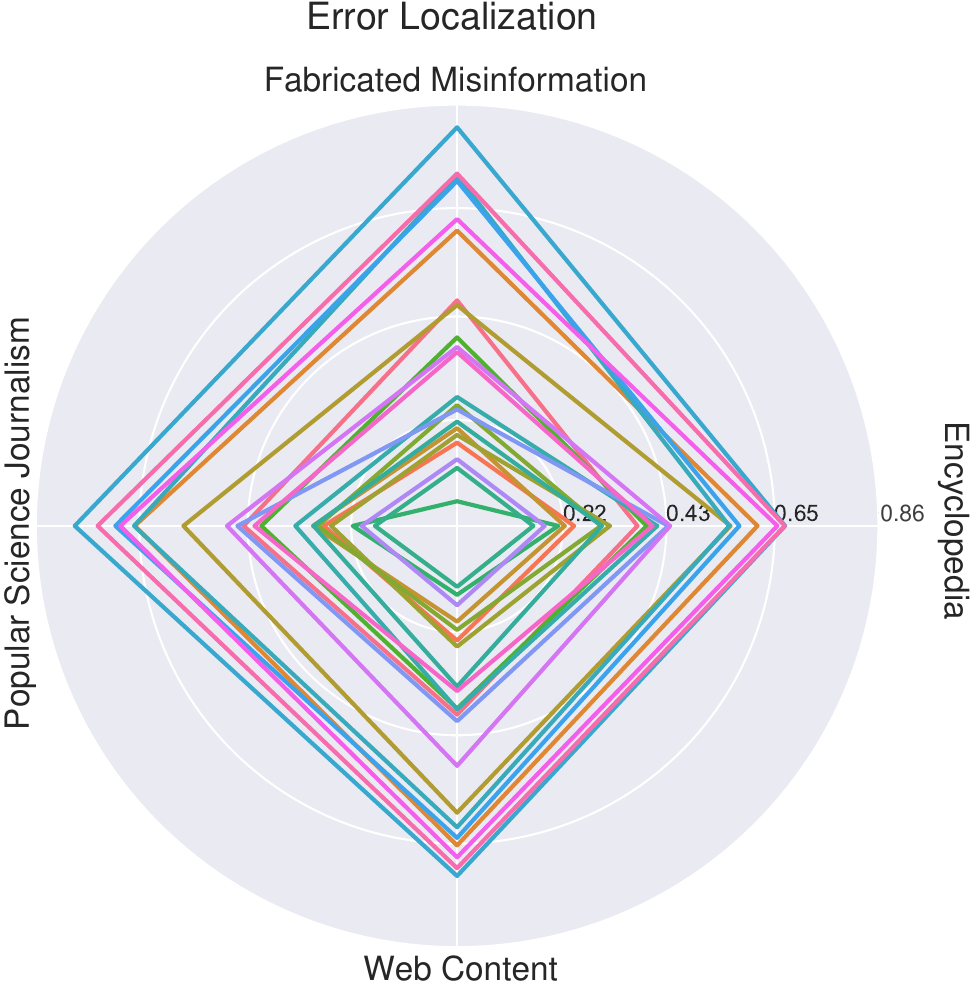}
    \end{minipage}
        
    \vspace{1em} 

    \begin{minipage}{\linewidth}
        \centering
        \includegraphics[width=1\linewidth]{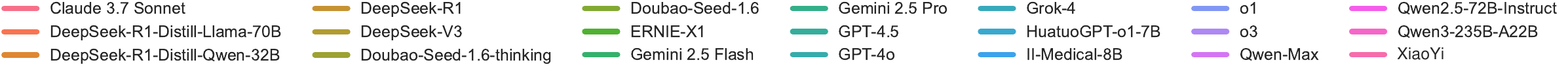}
    \end{minipage}
    \caption{Zero-shot performance of different models on \datasetname{} across different writing styles.}
    \label{fig:radar_main}
\end{figure}

We conduct a fine-grained analysis of model performance across writing styles (Figure~\ref{fig:radar_main}). For VC, models identify errors in fabricated medical misinformation more effectively than in encyclopedia texts, indicating that they detect exaggerated falsehoods but struggle with the inaccuracies often present in professionally curated content. For EL, performance varies less across styles, suggesting that localization depends less on stylistic cues and more on medical knowledge. EL thus serves as a more robust measure of domain expertise.
This style-dependent VC--EL asymmetry has two implications. First, single-style evaluation risks overstating operational capability: a system strong on misinformation-style inputs may fail silently on encyclopedia content, motivating the style-balanced curation in \datasetname{}. Second, the asymmetry is consistent with VC partially exploiting stylistic priors learned during web-scale pre-training, while EL requires committing to a specific erroneous span for which surface cues cannot substitute. For clinical assistance, where source texts skew toward encyclopedia and popular-science styles, EL Precision is the more reliable evaluation target.

\section{Conclusion}
We introduce \datasetname{}, a diverse, realistic, and uncontaminated benchmark for evaluating LLM fact-checking abilities in Chinese medical texts. Evaluation of 20 LLMs reveals critical performance deficits compared to human experts, particularly on fine-grained error localization. LLMs also exhibit an over-criticism failure mode that significantly degrades Precision. These challenges highlight the need for developing LLMs that are factually reliable for safe medical deployment.

\section{Limitations}
\datasetname{} targets the Chinese language within China's healthcare system, and performance may not generalize to other languages, cultures, or healthcare systems. This focus ensures high linguistic quality and cultural consistency, though future work could develop parallel benchmarks for cross-lingual comparison. For the EL task, we use GPT-4o as an automated judge, which enables scalable evaluation but has inherent limitations, including potential biases and inconsistencies. Future work could explore hybrid evaluation frameworks combining LLM judge scalability with expert validation for ambiguous or high-stakes cases. Additionally, \datasetname{} captures medical knowledge at creation time, and as medical science evolves, some information may become outdated. We prioritize foundational medical knowledge to minimize this issue, but acknowledge that dynamic benchmark updates would be valuable in the long term.


\section{Ethical Considerations}
For data collection and expert verification, we secured Institutional Review Board (IRB) approval and support from our institution's department. All expert annotators received fair compensation at or above the regional average wage. We removed all Personally Identifiable Information (PII), and data use is strictly limited to non-commercial research and complies with source platform terms under fair use. Because factually incorrect instances could be exploited to train harmful content generators or spread misinformation, we distribute the dataset under a restrictive license that prohibits clinical or patient-facing applications. The documentation warns about factual errors and specifies that the dataset is intended exclusively for research and benchmarking. \datasetname{} covers only Chinese medical texts, and the data, error distributions, and findings may not generalize to other languages, cultures, or healthcare systems. We caution against overly broad conclusions and encourage analogous benchmarks for other languages and contexts.





\bibliography{custom}

\input{appendix}

\end{document}

%% file: appendix.tex
\newpage
\appendix

\section{Detailed Related Works}
\subsection{LLMs in Medical Domains}
The capabilities of Large Language Models (LLMs) across many disciplines have prompted the exploration of their applications in the medical domain~\cite{huang2024comprehensivesurveyevaluatinglarge}. Medprompt~\cite{nori2023generalistfoundationmodelsoutcompete} enhances LLMs' performance on medical benchmarks by combining self-generated Chain-of-Thought (CoT)~\cite{NEURIPS2022_9d560961}, few-shot learning with dynamic example selection~\cite{liu-etal-2022-makes}, and self-consistency~\cite{wangself}. Other developments include HuaTuo~\cite{wang2023huatuotuningllamamodel}, a LLaMA-based model fine-tuned on Chinese medical instruction data; HuatuoGPT~\cite{zhang-etal-2023-huatuogpt}, a model specialized for medical consultation, fine-tuned on responses distilled from ChatGPT and real-world data; and PediatricsGPT~\cite{NEURIPS2024_fa5b423e}, a model developed via a systematic training pipeline for pediatric and general medical applications.
These advances highlight a growing interest in adapting LLMs to medical reasoning and the specialized requirements of real-world practice.

\subsection{Benchmarking Medical Knowledge in LLMs}
Numerous benchmarks have been developed to evaluate the medical knowledge of LLMs~\cite{nguyen-etal-2023-medredqa,NEURIPS2024_e15c4aff,abacha2025medecbenchmarkmedicalerror}. MedQA~\cite{app11146421} is a multiple-choice QA dataset derived from the United States Medical Licensing Examinations to assess the clinical knowledge of LLMs. Huatuo-26M~\cite{wang-etal-2025-huatuo} is a Chinese dataset comprising data from online consultation websites, encyclopedias, and information extracted from knowledge bases. 
More recently, HealthBench~\cite{arora2025healthbenchevaluatinglargelanguage} is a benchmark of realistic health conversations evaluated with custom, physician-created rubrics. 
MedXpertQA~\cite{zuo2025medxpertqabenchmarkingexpertlevelmedical} evaluates expert-level medical knowledge using both textual and multimodal tasks.

While some benchmarks for medical fact-checking exist, they often focus on narrow contexts.
VeriFact~\cite{Chung2025} verifies synthetic text against structured Electronic Health Records using a paradigm combining retrieval-augmented generation (RAG) and LLM-as-a-Judge. MEDEC~\cite{abacha2025medecbenchmarkmedicalerror} provides a benchmark for error detection and correction, but is confined to clinical notes. Relying on synthetic data or a single text genre, these benchmarks fail to capture the diverse contexts of medical information encountered in the real world, thereby limiting their utility for assessing how well models generalize. 
Consequently, a gap remains for a benchmark that can robustly evaluate fact-checking across a wide array of realistic medical texts.

\subsection{Benchmarking Factuality in LLMs}
Recent research has produced several benchmarks to evaluate factuality in LLMs. For instance, HaluEval~\cite{li-etal-2023-halueval} focuses on hallucination detection through a large-scale collection of automatically generated and human-annotated examples. SimpleQA~\cite{wei2024measuringshortformfactualitylarge} measures short-form factuality using single-answer questions adversarially collected to challenge state-of-the-art models like GPT-4. 
More broadly, OpenFactCheck~\cite{iqbal-etal-2024-openfactcheck} proposes a unified framework for benchmarking fact-checking across diverse sources, including free-form text and LLM-generated content. 
While these benchmarks provide valuable tools for assessing general-domain factuality, they lack the domain-specific nuances required for high-stakes fields like medicine, where misinformation can have severe public health consequences~\cite{10.1001/jamapediatrics.2023.5282}. Addressing this challenge requires a solution that bridges two critical gaps: the lack of domain-specific context in general factuality benchmarks and the lack of real-world diversity in existing medical fact-checking benchmarks. To this end, we introduce \datasetname{}, a new benchmark designed to evaluate the fact-checking capabilities of LLMs on a comprehensive and diverse collection of realistic medical texts.

\section{Dataset Construction \& Statistics}

\subsection{Multi-criteria Filtering}
\begin{figure*}[!t]
    \centering
    \includegraphics[width=1\linewidth]{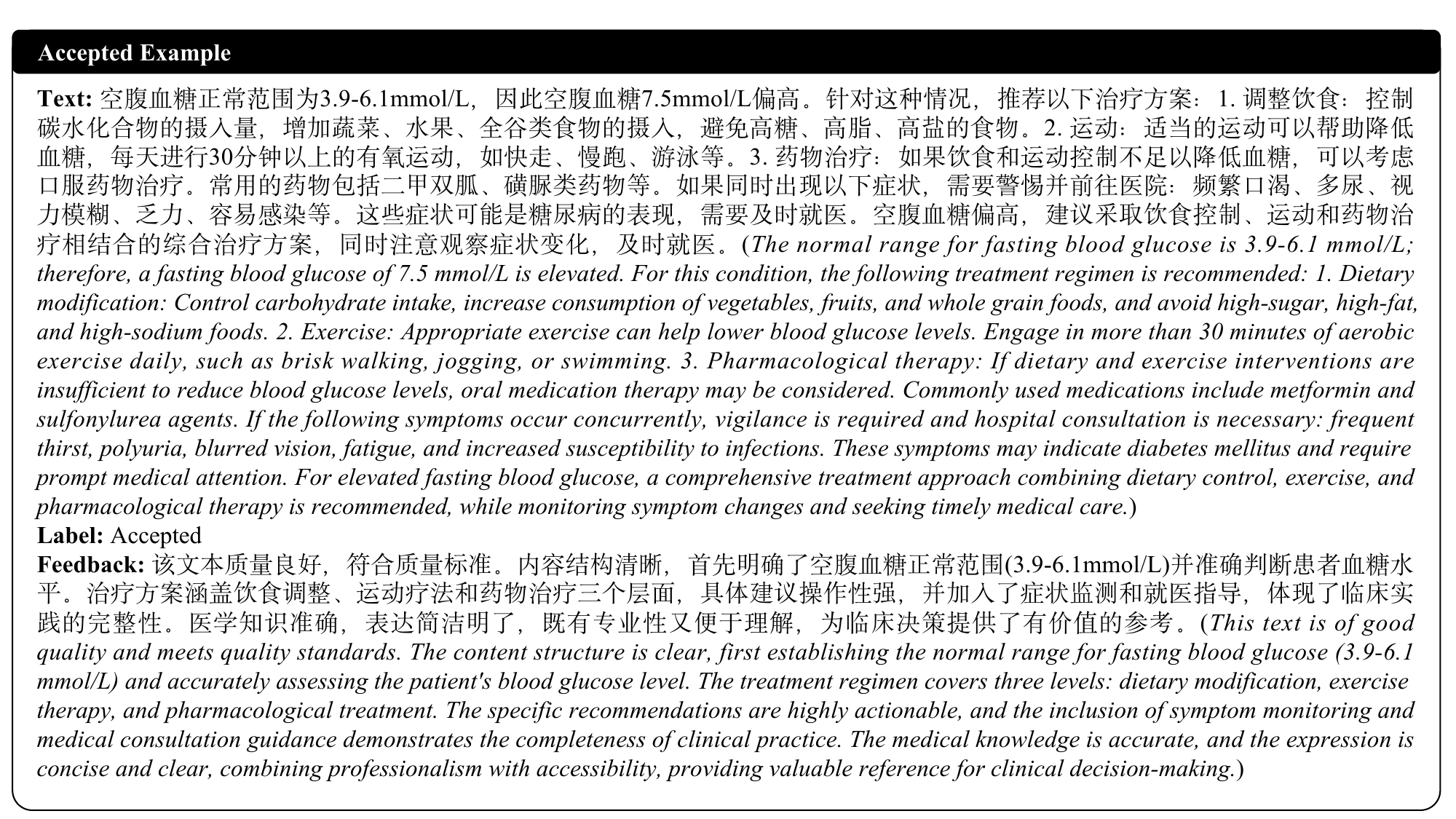}
    \caption{An example of accepted data with corresponding professional feedback.}
    \label{fig:reference_accept}
\end{figure*}

\begin{figure*}[!t]
    \centering
    \includegraphics[width=1\linewidth]{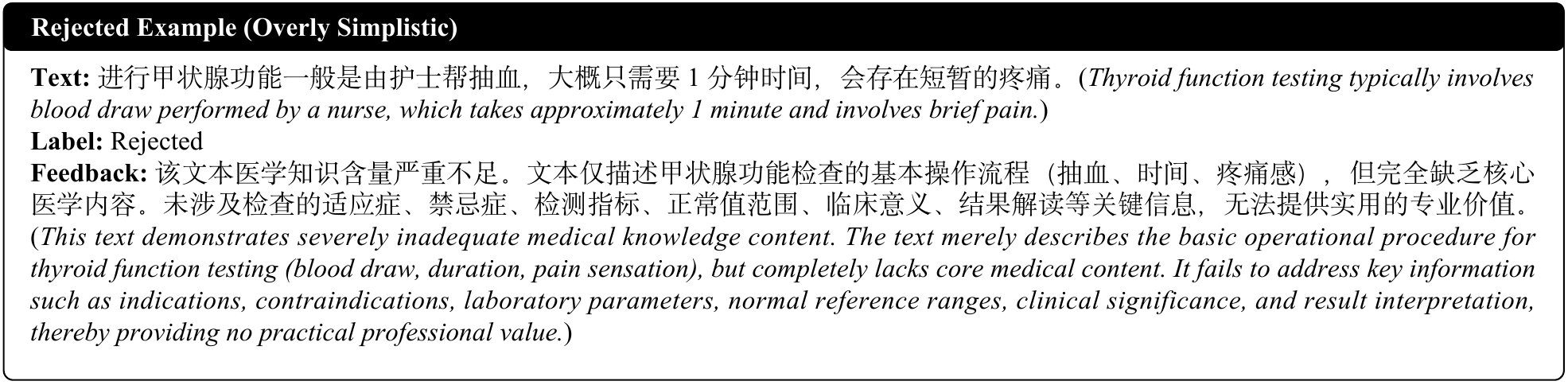}
    \vspace{1em} 
    \includegraphics[width=1\linewidth]{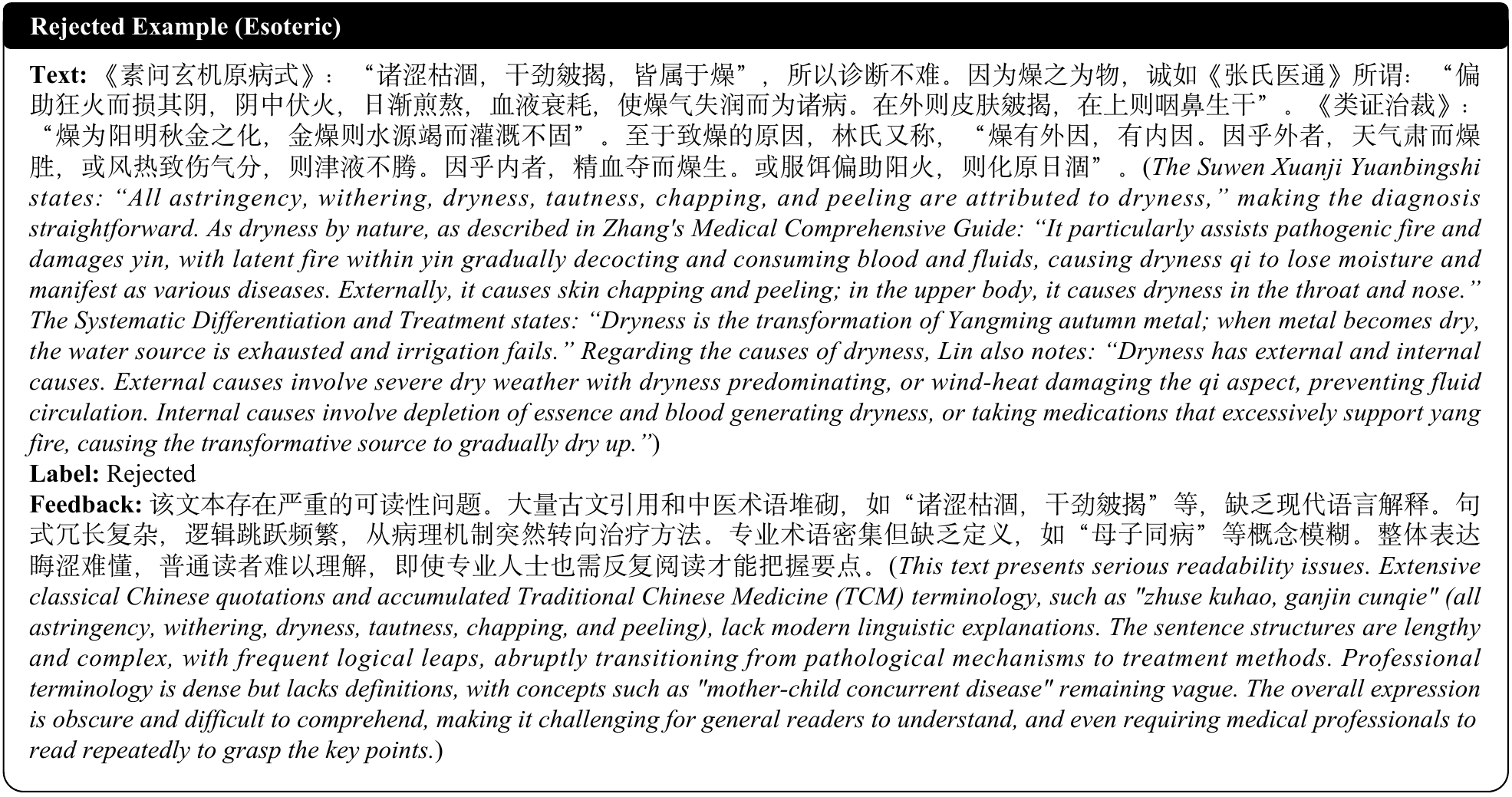}
    \vspace{1em} 
    \includegraphics[width=1\linewidth]{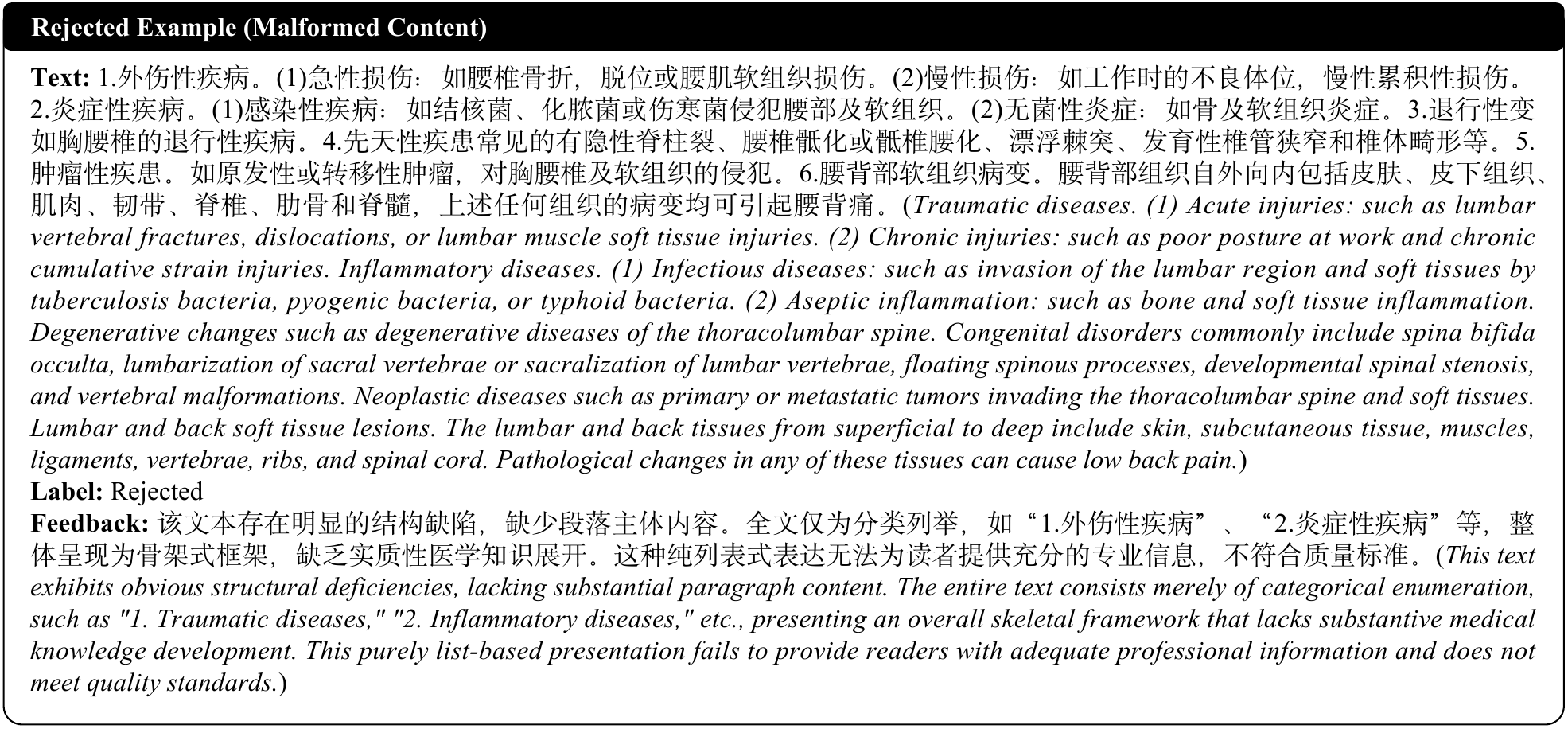}
    \caption{Examples of rejected data with corresponding professional feedback.}
    \label{fig:reference_reject}
\end{figure*}

Across the three filtering rounds, acceptance rates (passed instances over candidate pool size) were 67.69\%, 37.00\%, and 23.62\%. Agreement between the model suite and medical professionals rose across rounds: 69.72\%, 84.37\%, and 96.40\%, as the reference database grew from 412 to 624 instances.
Figure~\ref{fig:reference_accept} presents one accepted example, and Figure~\ref{fig:reference_reject} presents three rejected examples from the reference database, illustrating the filtering criteria for overly simplistic, esoteric, and malformed content.

\subsection{Error Taxonomy Definitions}
We define eight fine-grained error types in \datasetname{}, organized into two primary categories: \textbf{Medical Errors} and \textbf{Non-Medical Errors}. The definitions for each subcategory are detailed below.

\begin{table*}[t]
\centering
\begin{tabular}{ll}
\toprule
\textbf{Model} & \textbf{Full Name} \\
\midrule
II-Medical-8B & Intelligent-Internet/II-Medical-8B\\
HuatuoGPT-o1-7B & FreedomIntelligence/HuatuoGPT-o1-7B \\
DeepSeek-R1 & deepseek-ai/DeepSeek-R1-0528 \\ 
DeepSeek-V3 & deepseek-ai/DeepSeek-V3-0324\\
Qwen2.5-72B-Instruct & Qwen/Qwen2.5-72B-Instruct\\
Qwen3-235B-A22B & Qwen/Qwen3-235B-A22B\\
DeepSeek-R1-Distill-Llama-70B & deepseek-ai/DeepSeek-R1-Distill-Llama-70B \\
DeepSeek-R1-Distill-Qwen-32B & deepseek-ai/DeepSeek-R1-Distill-Qwen-32B\\
XiaoYi & xiaoyi-2025-06-24\\
GPT-4o & gpt-4o-2024-11-20 \\
GPT-4.5 & gpt-4.5-preview-2025-02-27\\
o1 & o1-2024-12-17\\
o3 & o3-2025-04-16 \\
Claude 3.7 Sonnet & claude-3-7-sonnet-20250219\\
Gemini 2.5 Flash & gemini-2.5-flash\\
Gemini 2.5 Pro & gemini-2.5-pro\\
Grok-4 & grok-4-0709 \\
Qwen2.5-Max & Qwen2.5-Max-2025-01-25\\
ERNIE-X1 & ernie-x1-32k-preview\\
Doubao-Seed-1.6 & doubao-seed-1.6\\
Doubao-Seed-1.6-thinking & doubao-seed-1.6-thinking\\
\bottomrule
\end{tabular}
\centering
\caption{Model versions and full names of the evaluated models.}
\label{tab:version}
\end{table*}

\textbf{Medical Errors}
Medical errors encompass factually incorrect content related to established medical knowledge, principles, or clinical practice. The definitions of each error type in \datasetname{} are as follows:
\begin{itemize}
\item \textbf{Conceptual Error:} Incorrect explanations of medical concepts, disease mechanisms, or drug actions; treatment recommendations that contradict established clinical guidelines; or inaccurate medication instructions (e.g., dosage, frequency). This category also encompasses errors in standard reference values for clinical data, such as laboratory test ranges or clinical scoring systems (e.g., GCS, APGAR).
\item \textbf{Terminological Error:} Use of non-standard, obsolete, or erroneous medical terminology. This includes incorrect translations or mappings between medical terms across different languages (e.g., Chinese and English).
\item \textbf{Temporal Error:} Errors in the chronological sequence of medical events or processes. Examples include misordering clinical procedures (e.g., administering post-operative medication pre-operatively), misrepresenting disease progression timelines, or describing developmental stages out of sequence.
\item \textbf{Others (Medical):} A residual category for medical errors not classifiable within the above types. This includes failures to convey critical safety warnings, flawed clinical reasoning (e.g., broken causal chains between symptoms and diagnoses), or misrepresentation of risk factors.
\end{itemize}

\textbf{Non-Medical Errors}
Non-medical errors encompass factual inaccuracies that are not specific to the medical domain.
\begin{itemize}
\item \textbf{Citation and Sourcing Error:} Fabrication or misrepresentation of sources, including citing non-existent literature, providing inaccurate citation details, or inventing data sources.
\item \textbf{Biased Content Error:} Content that reflects or promotes bias by perpetuating stereotypes related to social identity in diagnoses or treatment recommendations, or by employing language that reinforces social prejudice.
\item \textbf{General Factual Error:} Factual inaccuracies in non-medical background information, encompassing errors related to history, geography, basic scientific principles, or general world knowledge. 
\item \textbf{Others (Non-Medical):} A residual category for other non-medical issues, including ethical or legal violations (e.g., privacy breaches, suggesting illegal acts), undisclosed commercial promotion, or severe linguistic incoherence that fundamentally obstructs comprehension.
\end{itemize}

\subsection{Writing Style Taxonomy Definitions}
We categorize the writing styles of \datasetname{} into four distinct types: encyclopedia-style articles, popular science journalism, user-generated web content, and fabricated misinformation:
\begin{itemize}
\item \textbf{Encyclopedia-style:} Formal content characteristic of medical texts, featuring specialized terminology and structured presentation.
\item \textbf{Popular Science Journalism:} Articles that translate complex medical topics for lay audiences, employing simplified language, narrative elements, and references to purported experts to establish credibility.
\item \textbf{User-generated Web Content:} Informal medical discussions sourced from online Q\&A platforms, characterized by colloquial language and personal anecdotes.
\item \textbf{Fabricated Misinformation:} Deliberately false or misleading medical content that employs exaggerated claims and distorts facts, featuring sensationalized language, cherry-picked data, and conspiracy theories while mimicking legitimate sources.
\end{itemize}

\subsection{Data Statistics}
Tables~\ref{tab:error} and~\ref{tab:difficulty_level} present the data distribution across different error types and difficulty levels, respectively. Difficulty levels are determined by model performance.

\subsection{Data Contamination}
To investigate potential data contamination within the pre-training corpora of the evaluated LLMs, we devised a text-completion task. For each instance in \datasetname{}, we prompted the models with the first half of the text and tasked them with generating the continuation. We then measured the textual overlap between the generated output and the ground-truth second half using ROUGE-F1 scores, where a high score would suggest that the model had memorized the content.
As shown in Table~\ref{tab:data_leakage}, all evaluated models exhibit negligible ROUGE scores, which strongly indicates that \datasetname{} serves as a previously unseen benchmark.

\subsection{Data Examples}
\begin{figure*}[!t]
    \centering
    \includegraphics[width=1\linewidth]{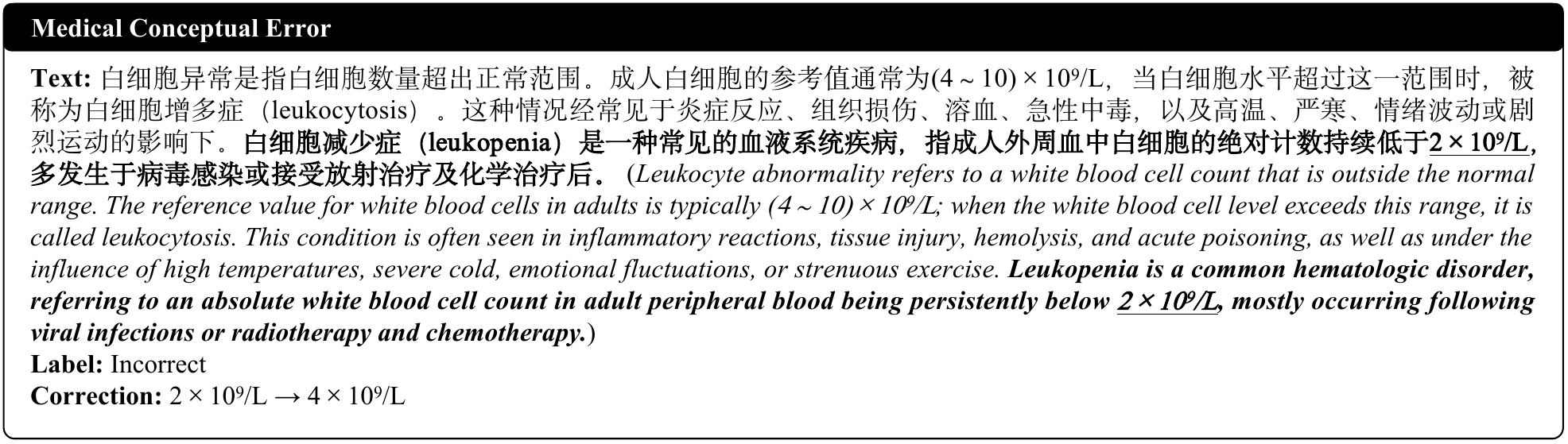}
    \caption{A data example of an error categorized as \textbf{\underline{medical conceptual error}}.}
    \label{fig:medical_conceptual_error}
\end{figure*}

\begin{figure*}[!t]
    \centering
    \includegraphics[width=1\linewidth]{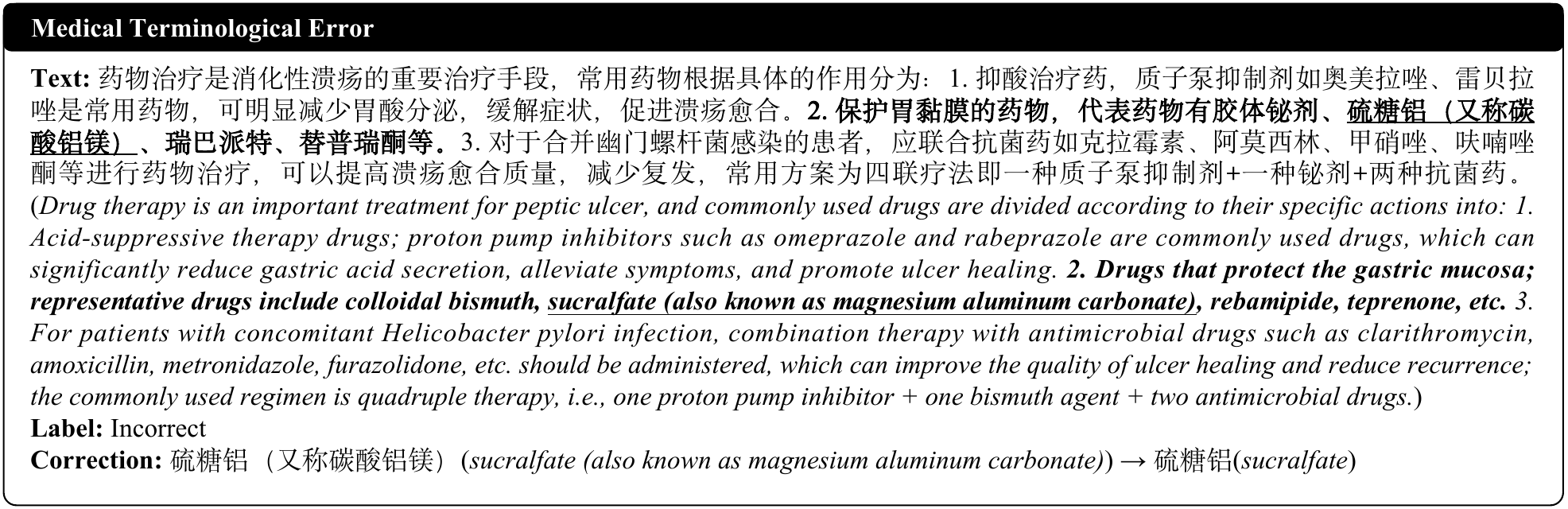}
    \caption{A data example of an error categorized as \textbf{\underline{medical terminological error}}.}
    \label{fig:medical_terminological_error}
\end{figure*}

\begin{figure*}[!t]
    \centering
    \includegraphics[width=1\linewidth]{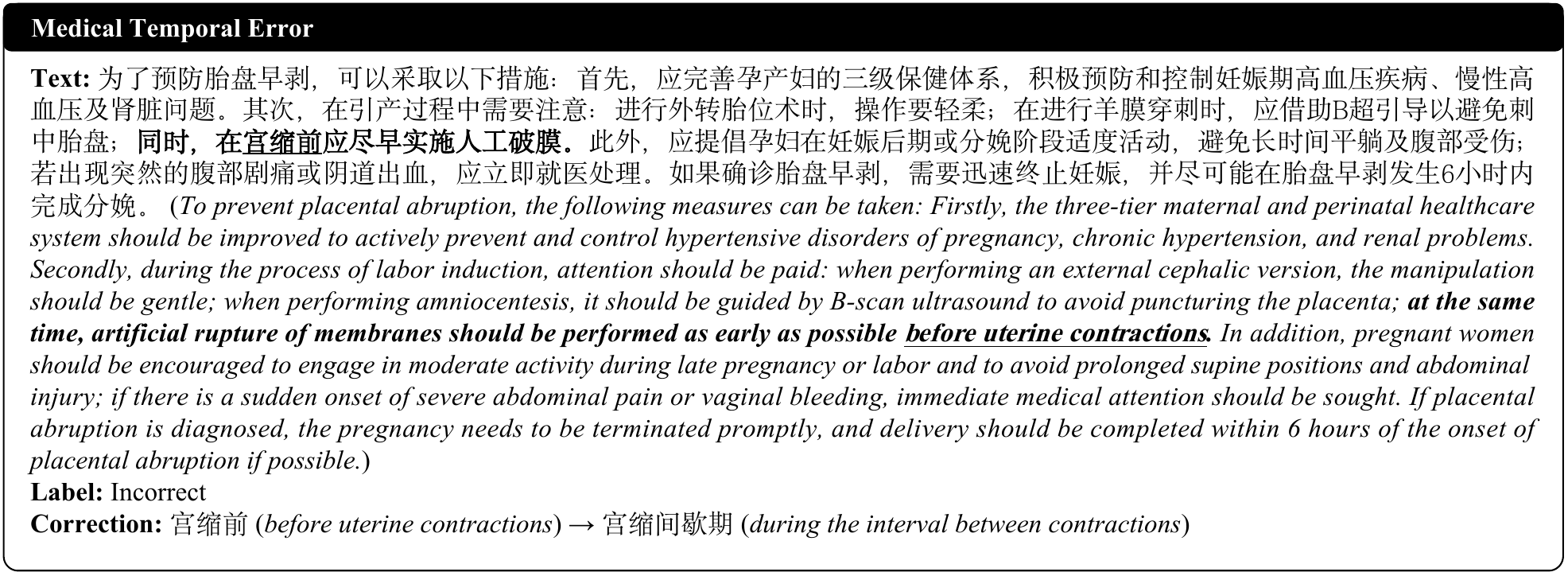}
    \caption{A data example of an error categorized as \textbf{\underline{medical temporal error}}.}
    \label{fig:medical_temporal_error}
\end{figure*}

\begin{figure*}[!t]
    \centering
    \includegraphics[width=1\linewidth]{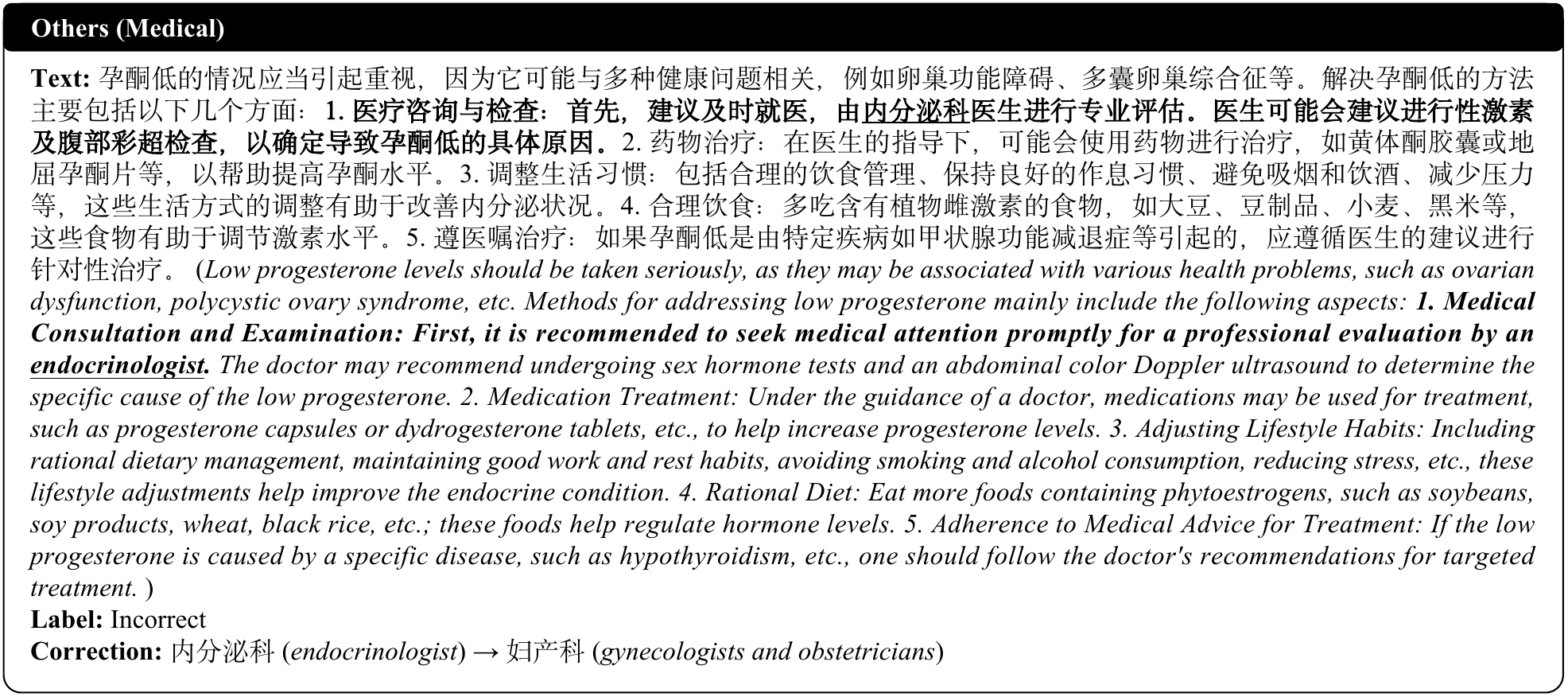}
    \caption{A data example of an error categorized as \textbf{\underline{other medical error}}.}
    \label{fig:medical_other_error}
\end{figure*}

\begin{figure*}[!t]
    \centering
    \includegraphics[width=1\linewidth]{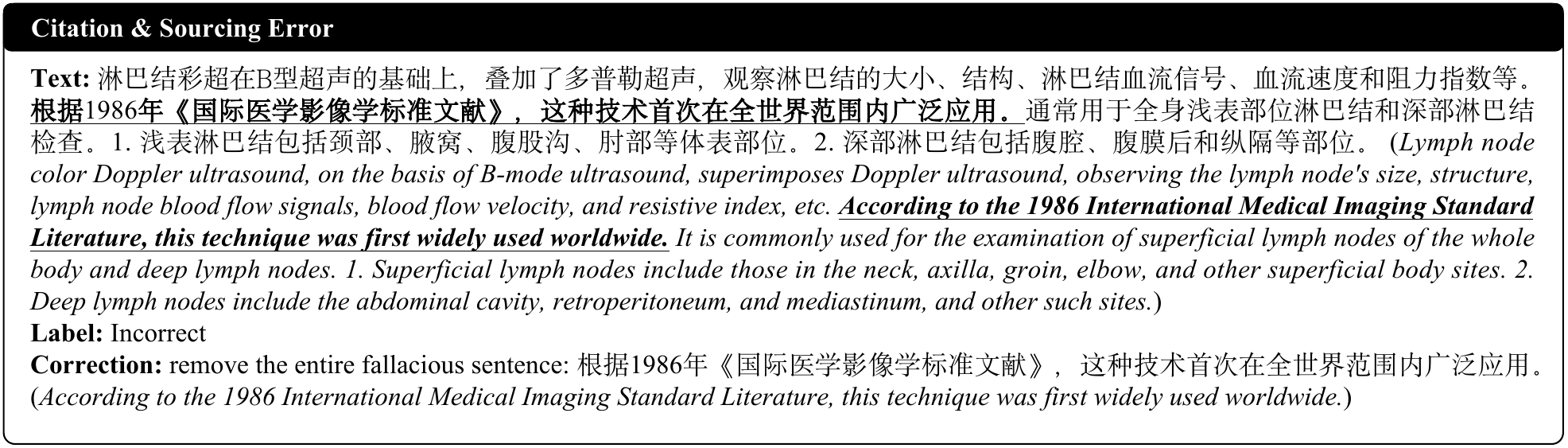}
    \caption{A data example of an error categorized as \textbf{\underline{citation \& sourcing error}}.}
    \label{fig:citation_error}
\end{figure*}

\begin{figure*}[!t]
    \centering
    \includegraphics[width=1\linewidth]{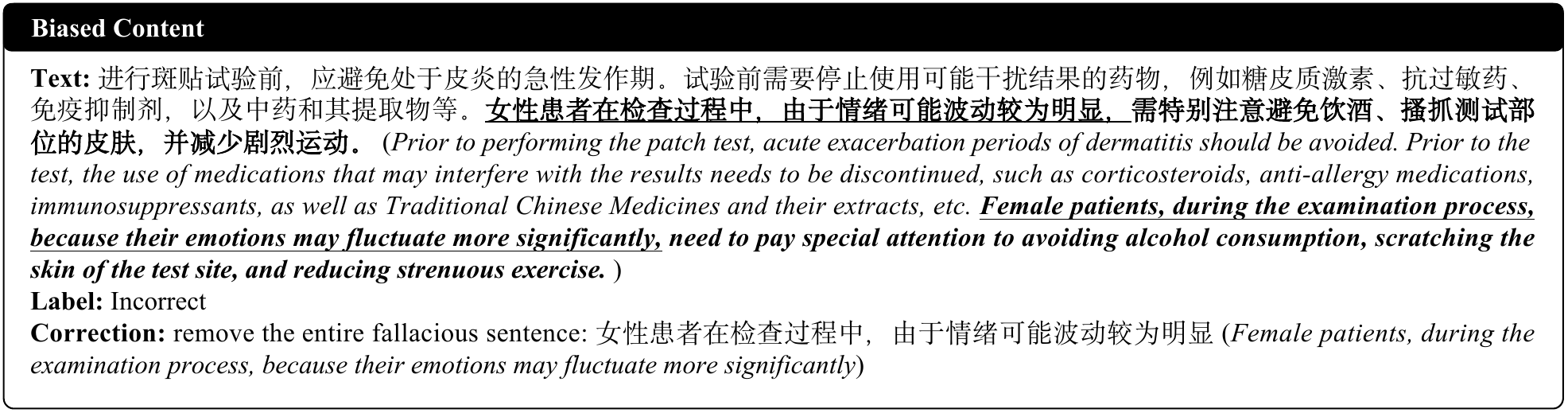}
    \caption{A data example of an error categorized as \textbf{\underline{biased content}}.}
    \label{fig:biased_content}
\end{figure*}

\begin{figure*}[!t]
    \centering
    \includegraphics[width=1\linewidth]{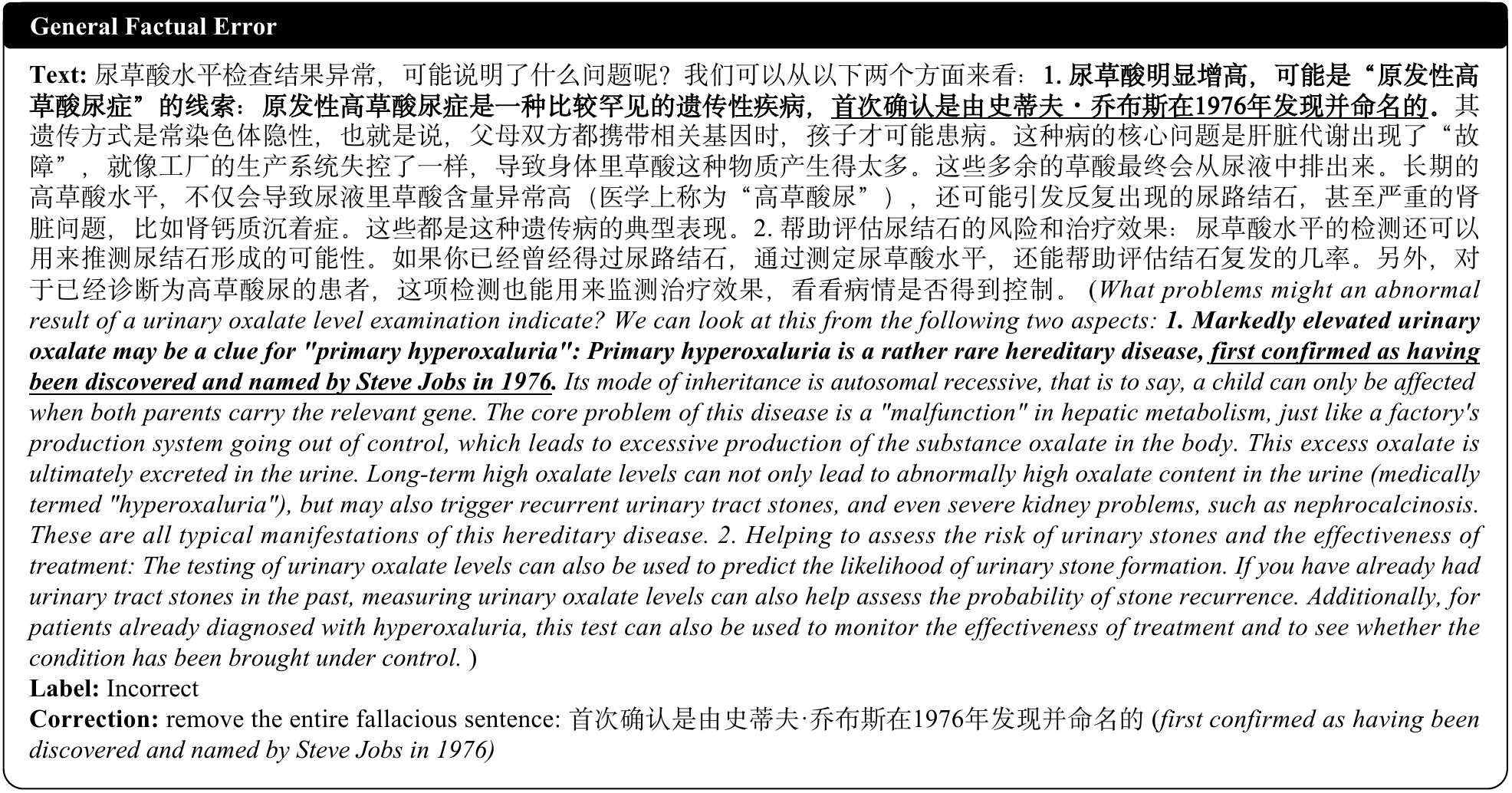}
    \caption{A data example of an error categorized as \textbf{\underline{general factual error}}.}
    \label{fig:general_factual_error}
\end{figure*}

\begin{figure*}[!t]
    \centering
    \includegraphics[width=1\linewidth]{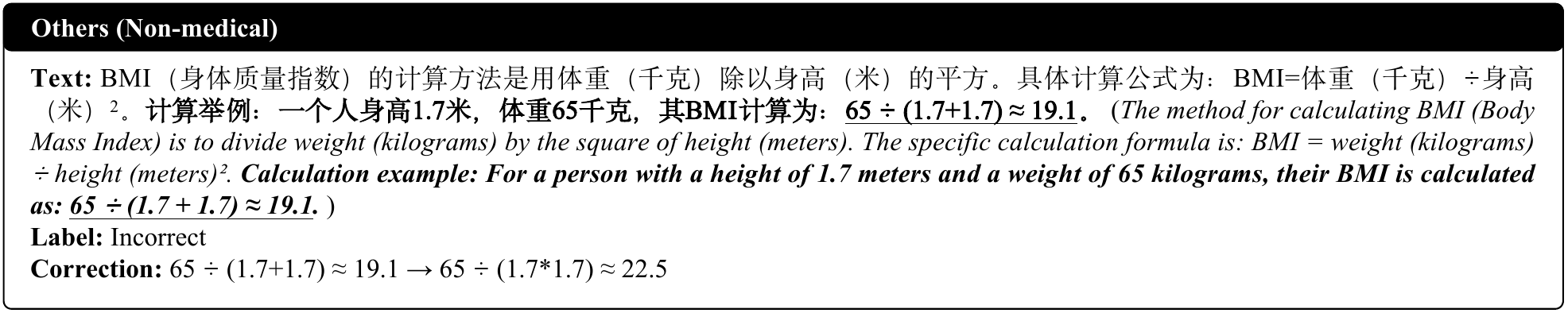}
    \caption{A data example of an error categorized as \textbf{\underline{other non-medical error}}.}
    \label{fig:nonmedical_other_error}
\end{figure*}

\begin{figure*}[!t]
    \centering
    \includegraphics[width=1\linewidth]{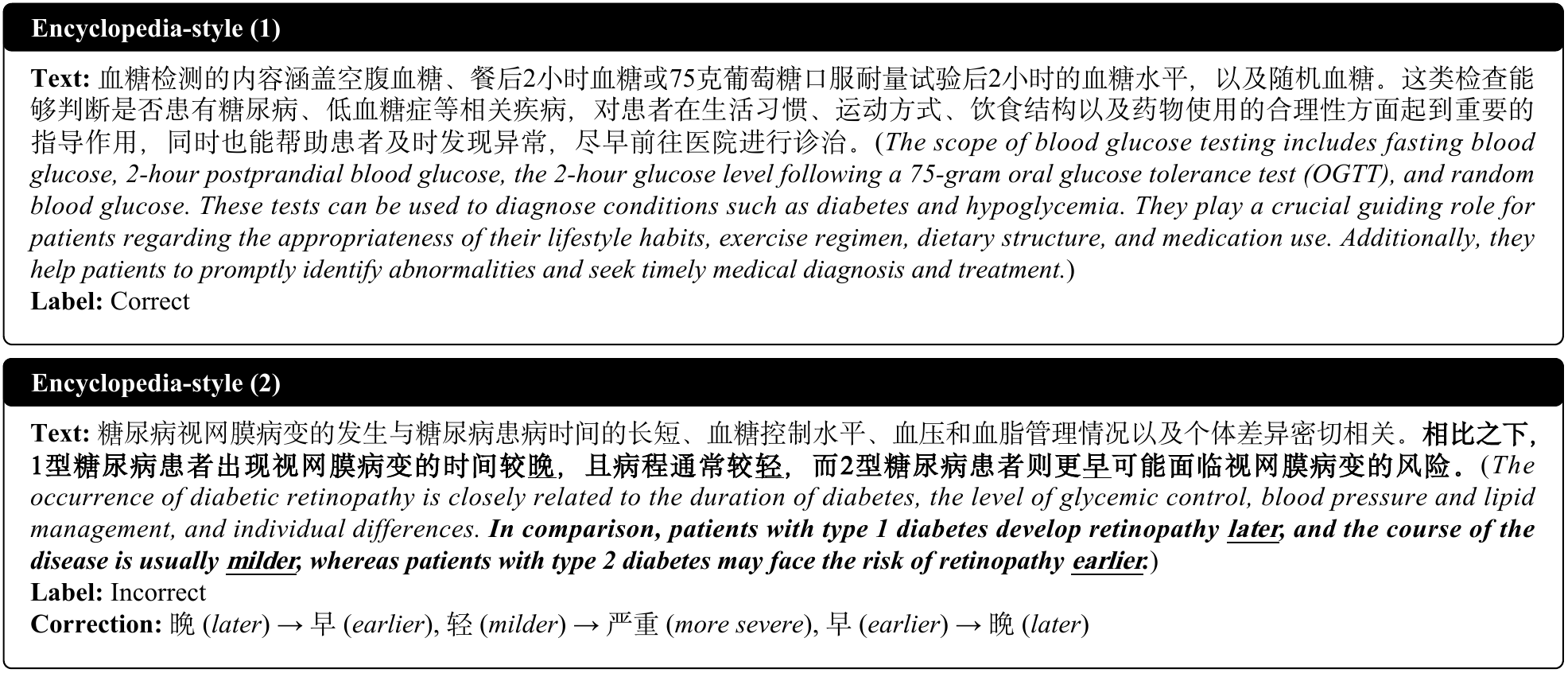}
    \caption{A data example of an error categorized as \textbf{\underline{encyclopedia-style}}.}
    \label{fig:encyclopedia-style}
\end{figure*}

\begin{figure*}[!t]
    \centering
    \includegraphics[width=1\linewidth]{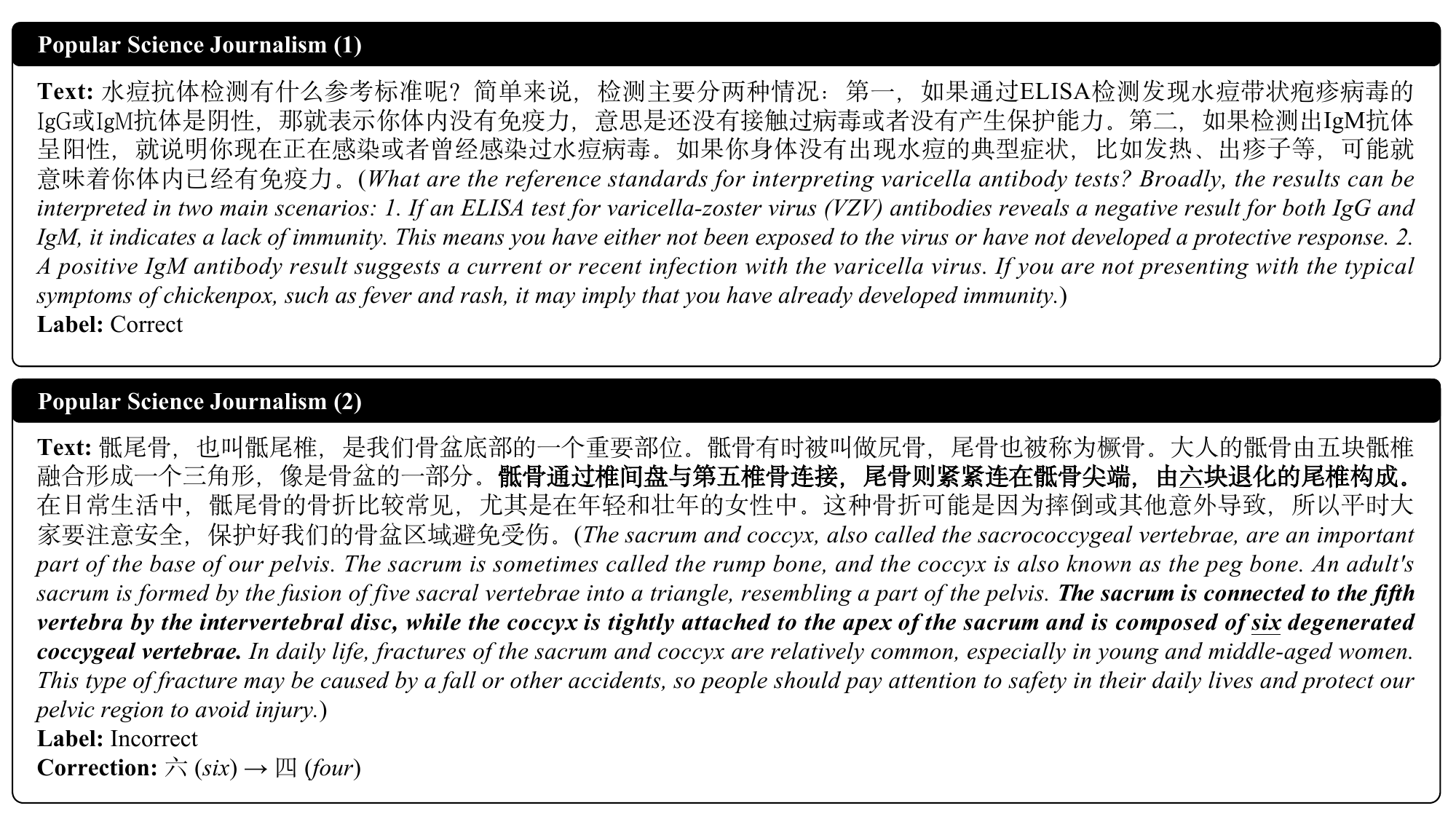}
    \caption{Data examples with writing style of \textbf{\underline{popular science journalism}}.}
    \label{fig:popular_science_journalism}
\end{figure*}

\begin{figure*}[!t]
    \centering
    \includegraphics[width=1\linewidth]{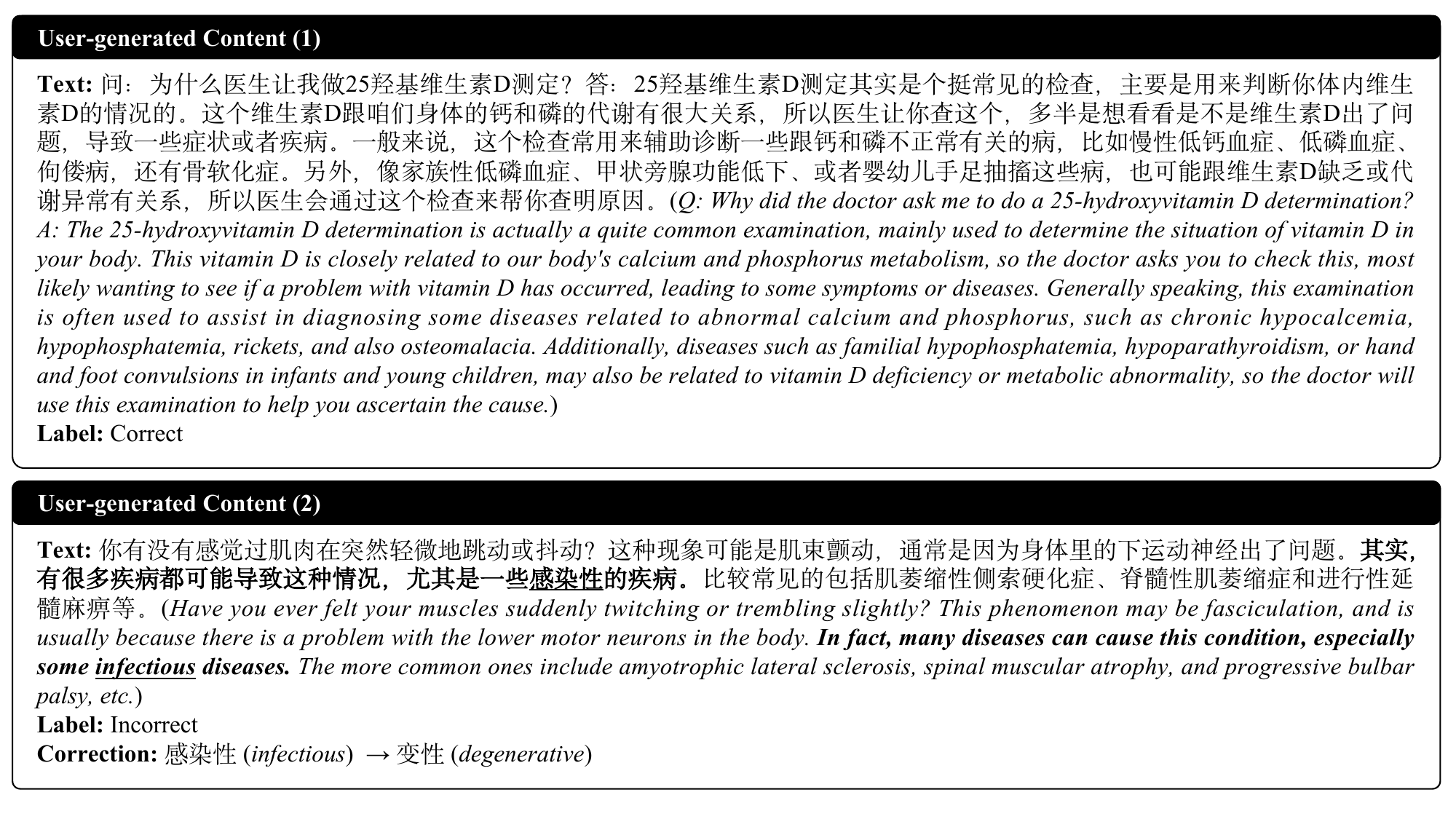}
    \caption{Data examples with writing style of \textbf{\underline{user-generated web content}}.}
    \label{fig:user-generated_web_content}
\end{figure*}

\begin{figure*}[!t]
    \centering
    \includegraphics[width=1\linewidth]{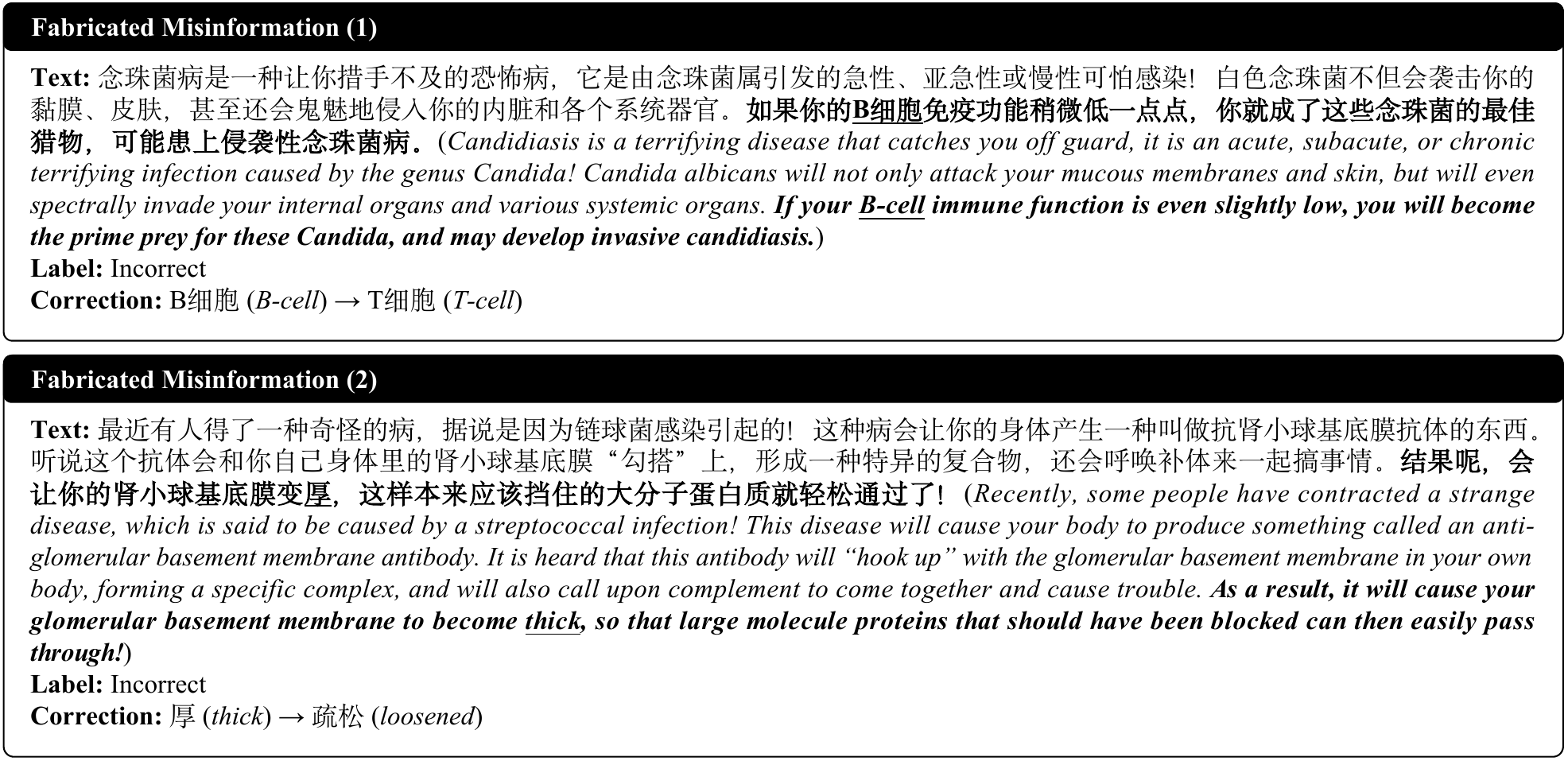}
    \caption{Data examples with writing style of \textbf{\underline{fabricated misinformation}}.}
    \label{fig:fabricated_misinformation}
\end{figure*}

Figures~\ref{fig:medical_conceptual_error} to~\ref{fig:medical_other_error} present examples for each medical error. Figures~\ref{fig:citation_error} to~\ref{fig:nonmedical_other_error} present examples for each non-medical error.
Figures~\ref{fig:encyclopedia-style} to~\ref{fig:fabricated_misinformation} present examples for each writing style.

\section{Implementation Details}
\subsection{Prompt Templates}
\begin{figure*}[!t]
    \centering
    \includegraphics[width=1\linewidth]{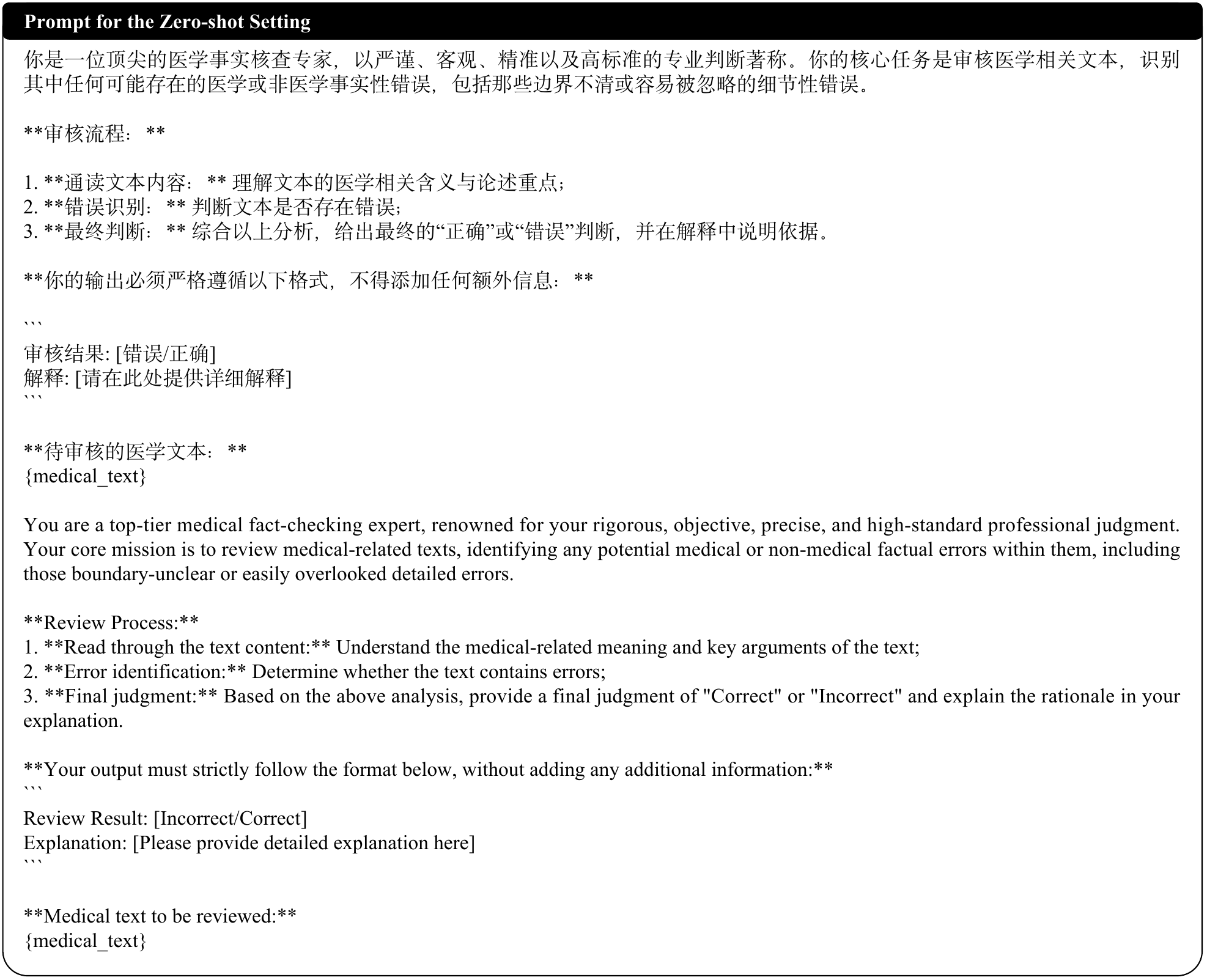}
    \caption{Prompt template used for zero-shot prompting.}
    \label{fig:prompt_1}
\end{figure*}

\begin{figure*}[!t]
    \centering
    \includegraphics[width=1\linewidth]{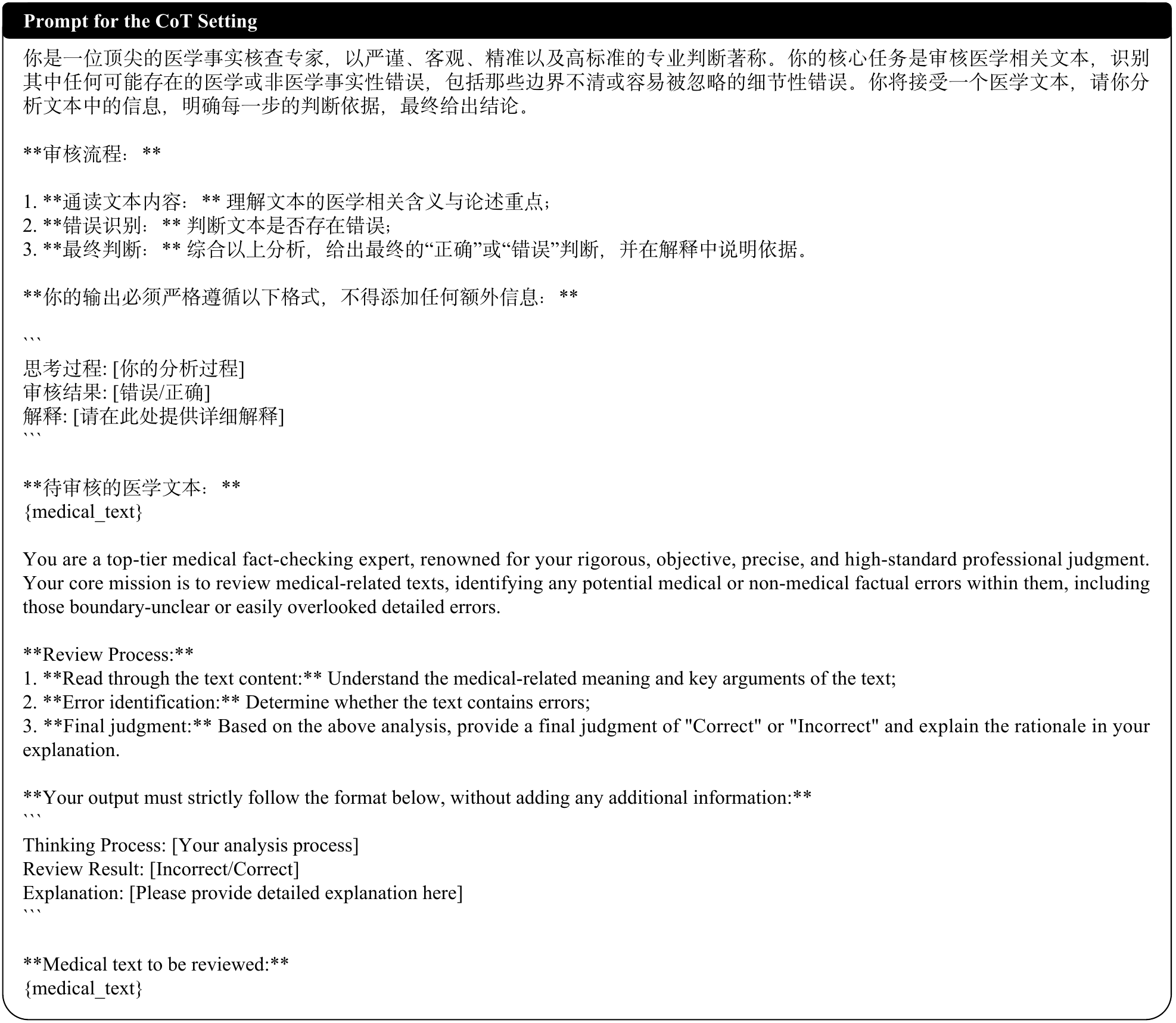}
    \caption{Prompt template used for CoT prompting.}
    \label{fig:prompt_2}
\end{figure*}

Figures~\ref{fig:prompt_1} to~\ref{fig:prompt_2} present the prompts used for zero-shot and CoT prompting. All prompts were originally formulated in Chinese; we provide their corresponding English translations for reference.

\subsection{Model Versions}
We accessed proprietary models through API calls and deployed open-source models locally. 
Table~\ref{tab:version} lists the model versions and full names.

\subsection{LLM-as-a-Judge}
For the evaluation of EL tasks, we assessed the agreement between different models and human experts. DeepSeek-R1, GPT-4o, o1, Claude 3.7 Sonnet, and Gemini 2.5 Pro achieved Cohen's $\kappa$ values of 0.800, 0.870, 0.870, 0.845, and 0.865, respectively, on a random 10\% sample of the veracity classification task results from Doubao-Seed-1.6-thinking, XiaoYi, and Qwen2.5-Max. Considering API costs and evaluation performance, we selected GPT-4o for the LLM-as-a-Judge evaluation.

\section{Experimental Results}
\subsection{Full Result Tables}
\begin{figure*}[!t]
    \centering
    \includegraphics[width=0.8\linewidth]{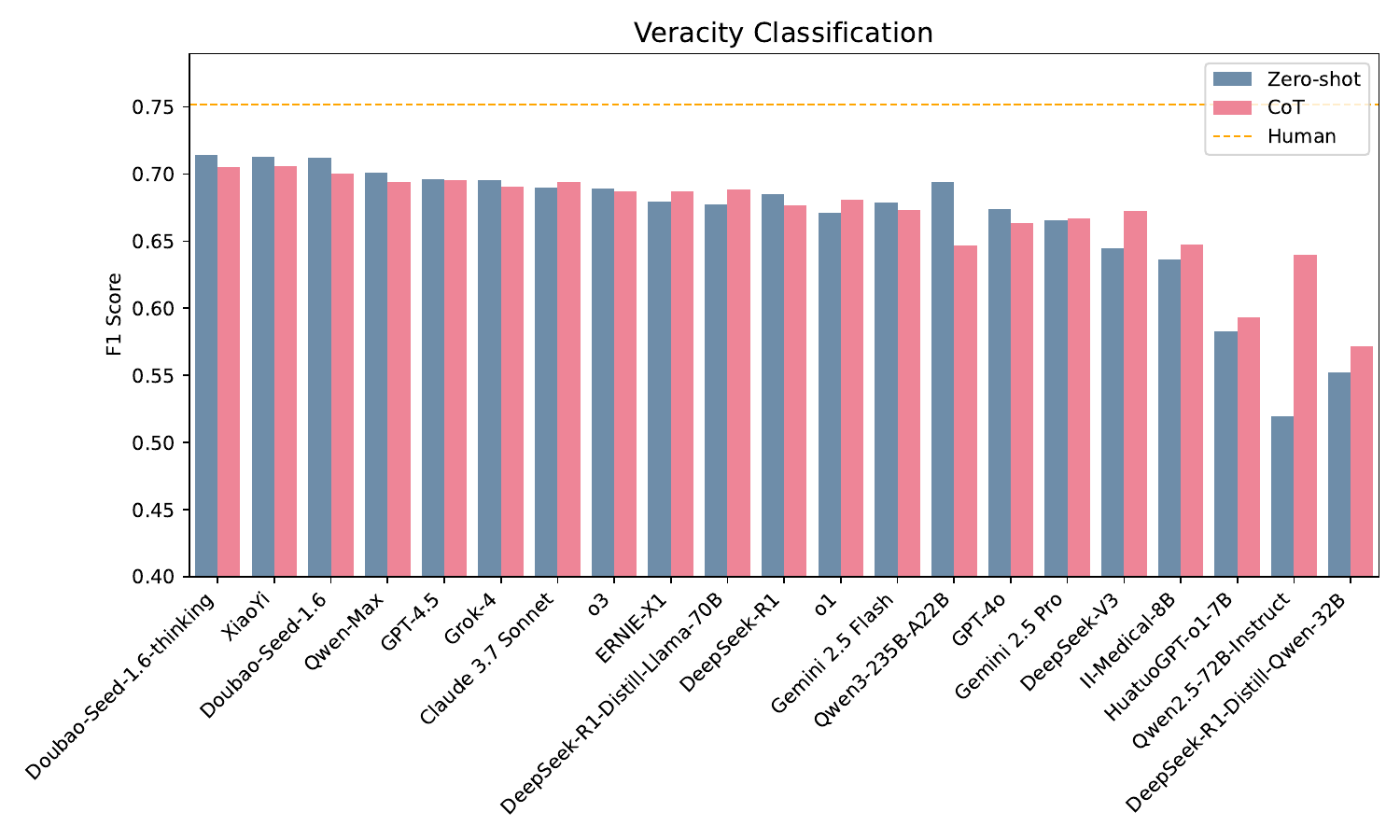}
    \caption{Veracity classification performance of human experts and models on \datasetname{} in zero-shot and CoT settings.}
    \label{fig:original_results_1}
\end{figure*}

\begin{figure*}[!t]
    \centering
    \includegraphics[width=0.8\linewidth]{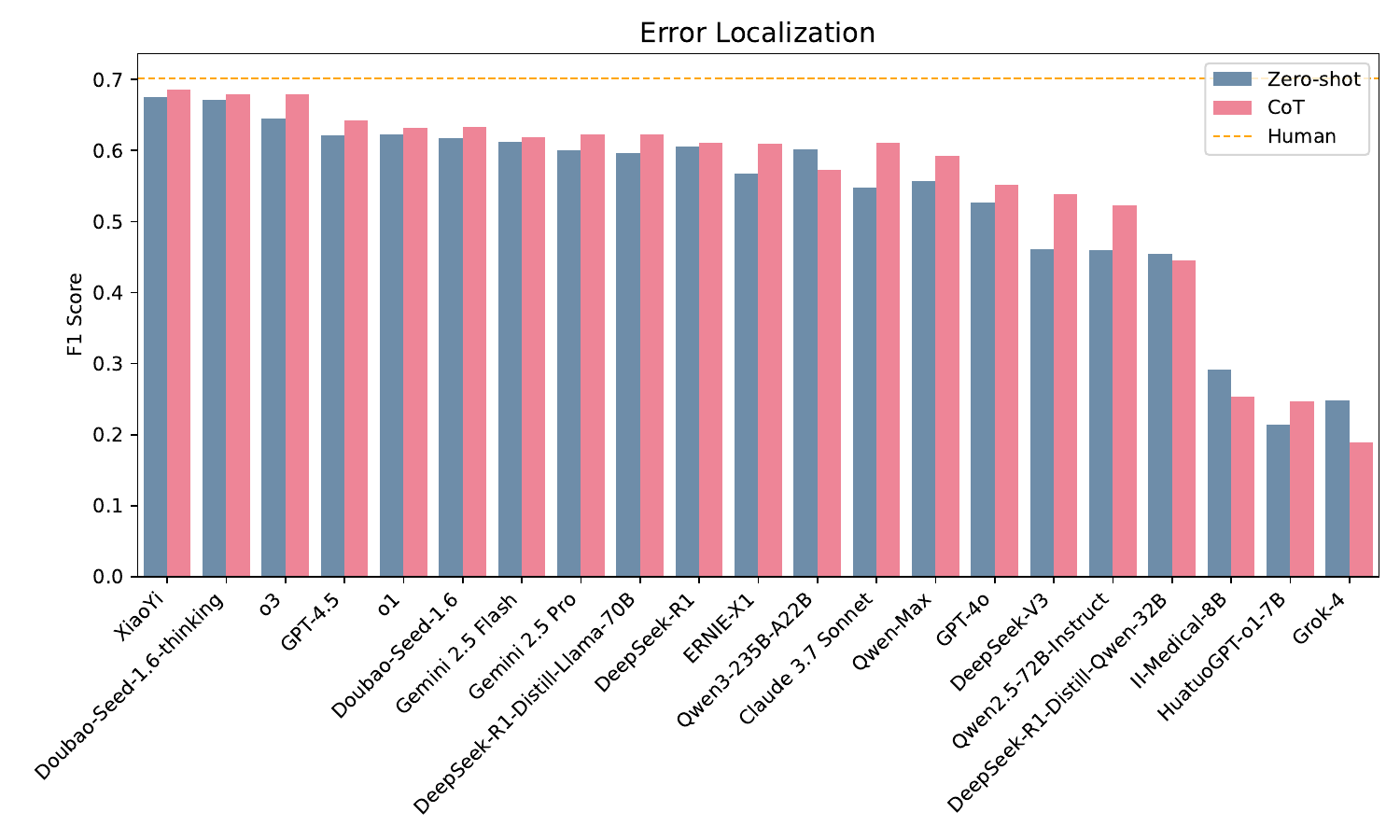}
    \caption{Error localization performance of human experts and models on \datasetname{} in zero-shot and CoT settings.}
    \label{fig:original_results_2}
\end{figure*}

We present the complete results of our main experiments in Tables~\ref{tab:original_results_1} and~\ref{tab:original_results_2} and Figures~\ref{fig:original_results_1} and~\ref{fig:original_results_2}. The results from experiments using advanced evaluation strategies and inference-time scaling are provided in Tables~\ref{table:agent} and~\ref{table:tts}, respectively. 

\begin{table*}[!t]
\centering
\begin{tabular}{lcccccc}
\toprule
\textbf{Model} & \multicolumn{3}{c}{\textbf{Zero-shot}} & \multicolumn{3}{c}{\textbf{CoT}} \\
\cmidrule(lr){2-4} \cmidrule(lr){5-7}
& Precision & Recall & F1 Score & Precision & Recall & F1 Score \\
\midrule
Human & 0.7695 & 0.7353 & 0.7521 & -- & -- & -- \\
\midrule
\multicolumn{7}{c}{\textbf{Open-source}} \\
\midrule
II-Medical-8B & 0.5687 & 0.7216 & 0.6361 & 0.5724 & 0.7459 & 0.6477 \\
HuatuoGPT-o1-7B & 0.6004 & 0.5668 & 0.5831 & 0.6185 & 0.5699 & 0.5932 \\
DeepSeek-R1 & 0.5488 & 0.9101 & 0.6847 & 0.5360 & 0.9163 & 0.6764 \\
DeepSeek-V3 & 0.6883 & 0.6057 & 0.6444 & 0.6006 & 0.7636 & 0.6724 \\
Qwen2.5-72B-Instruct & \textbf{0.8264} & 0.3788 & 0.5195 & \textbf{0.7320} & 0.5686 & 0.6400 \\
Qwen3-235B-A22B & 0.6720 & 0.7183 & 0.6944 & 0.6949 & 0.6049 & 0.6468 \\
DeepSeek-R1-Distill-Llama-70B & 0.5451 & 0.8947 & 0.6774 & 0.5490 & 0.9220 & 0.6882 \\
DeepSeek-R1-Distill-Qwen-32B & \underline{0.7204} & 0.4480 & 0.5524 & 0.7194 & 0.4744 & 0.5718 \\
\midrule
\multicolumn{7}{c}{\textbf{Proprietary}} \\
\midrule
XiaoYi & 0.6694 & 0.7618 & \underline{0.7126} & 0.6631 & 0.7552 & \textbf{0.7061} \\
GPT-4o & 0.5842 & 0.7968 & 0.6741 & 0.5646 & 0.8044 & 0.6635 \\
GPT-4.5 & 0.6108 & 0.8100 & 0.6965 & 0.6254 & 0.7826 & 0.6952 \\
o1 & 0.7155 & 0.6323 & 0.6713 & \underline{0.7205} & 0.6456 & 0.6810 \\
o3 & 0.5596 & 0.8960 & 0.6890 & 0.5507 & 0.9128 & 0.6870 \\
Claude 3.7 Sonnet & 0.6063 & 0.8006 & 0.6900 & 0.6021 & 0.8197 & 0.6943 \\
Gemini 2.5 Flash & 0.5369 & \underline{0.9225} & 0.6787 & 0.5240 & \underline{0.9405} & 0.6730 \\
Gemini 2.5 Pro & 0.5067 & \textbf{0.9707} & 0.6658 & 0.5082 & \textbf{0.9688} & 0.6667 \\
Grok-4 & 0.5785 & 0.8705 & 0.6951 & 0.5789 & 0.8563 & 0.6908 \\
Qwen2.5-Max & 0.6873 & 0.7146 & 0.7006 & 0.6031 & 0.8175 & 0.6942 \\
ERNIE-X1 & 0.6210 & 0.7495 & 0.6792 & 0.6643 & 0.7116 & 0.6871 \\
Doubao-Seed-1.6 & 0.6086 & 0.8582 & 0.7122 & 0.5979 & 0.8459 & 0.7006 \\
Doubao-Seed-1.6-thinking & 0.6711 & 0.7625 & \textbf{0.7139} & 0.6437 & 0.7791 & \underline{0.7050} \\
\bottomrule
\end{tabular}
\centering
\caption{Veracity classification performance of human experts and models on \datasetname{} in zero-shot and CoT settings. The best results are \textbf{bolded} and the second-best are \underline{underlined}.}
\label{tab:original_results_1}
\end{table*}

\begin{table*}[!t]
\centering
\begin{tabular}{lcccccc}
\toprule
\textbf{Model} & \multicolumn{3}{c}{\textbf{Zero-shot}} & \multicolumn{3}{c}{\textbf{CoT}} \\
\cmidrule(lr){2-4} \cmidrule(lr){5-7}
& Precision & Recall & F1 Score & Precision & Recall & F1 Score \\
\midrule
Human & 0.7495 & 0.6588 & 0.7012 & -- & -- & -- \\
\midrule
\multicolumn{7}{c}{\textbf{Open-source}} \\
\midrule
II-Medical-8B & 0.3278 & 0.2618 & 0.2911 & 0.2904 & 0.2240 & 0.2529 \\
HuatuoGPT-o1-7B & 0.3054 & 0.1654 & 0.2146 & 0.3508 & 0.1900 & 0.2465 \\
DeepSeek-R1 & 0.5035 & 0.7580 & 0.6051 & 0.5012 & 0.7826 & 0.6111 \\
DeepSeek-V3 & 0.5832 & 0.3809 & 0.4608 & 0.5266 & 0.5510 & 0.5386 \\
Qwen2.5-72B-Instruct & \textbf{0.8024} & 0.3223 & 0.4599 & \underline{0.6731} & 0.4282 & 0.5234 \\
Qwen3-235B-A22B & 0.6234 & 0.5803 & 0.6011 & 0.6565 & 0.5076 & 0.5725 \\
DeepSeek-R1-Distill-Llama-70B & 0.4990 & 0.7410 & 0.5964 & 0.5142 & 0.7873 & 0.6221 \\
DeepSeek-R1-Distill-Qwen-32B & 0.6648 & 0.3450 & 0.4543 & 0.6480 & 0.3393 & 0.4454 \\
\midrule
\multicolumn{7}{c}{\textbf{Proprietary}} \\
\midrule
XiaoYi & 0.6512 & 0.7023 & \textbf{0.6758} & 0.6530 & 0.7221 & \textbf{0.6858}\\
GPT-4o & 0.4966 & 0.5595 & 0.5262 & 0.5012 & 0.6144 & 0.5520 \\
GPT-4.5 & 0.5694 & 0.6824 & 0.6208 & 0.5971 & 0.6947 & 0.6422 \\
o1 & \underline{0.6921} & 0.5652 & 0.6223 & \textbf{0.6975} & 0.5775 & 0.6319 \\
o3 & 0.5355 & \underline{0.8129} & 0.6456 & 0.5579 & \underline{0.8658} & 0.6785 \\
Claude 3.7 Sonnet & 0.5242 & 0.5728 & 0.5474 & 0.5567 & 0.6777 & 0.6113 \\
Gemini 2.5 Flash & 0.4988 & 0.7921 & 0.6121 & 0.4927 & 0.8299 & 0.6183 \\
Gemini 2.5 Pro & 0.4689 & \textbf{0.8346} & 0.6005 & 0.4828 & \textbf{0.8752} & 0.6223 \\
Grok-4 & 0.2675 & 0.2316 & 0.2482 & 0.2136 & 0.1692 & 0.1888 \\
Qwen2.5-Max & 0.6113 & 0.5113 & 0.5569 & 0.5663 & 0.6219 & 0.5928 \\
ERNIE-X1 & 0.5580 & 0.5775 & 0.5676 & 0.6240 & 0.5945 & 0.6089 \\
Doubao-Seed-1.6 & 0.5566 & 0.6928 & 0.6173 & 0.5609 & 0.7268 & 0.6332 \\
Doubao-Seed-1.6-thinking & 0.6501 & 0.6938 & \underline{0.6712} & 0.6307 & 0.7344 & \underline{0.6786} \\
\bottomrule
\end{tabular}
\centering
\caption{Error localization performance of human experts and models on \datasetname{} in zero-shot and CoT settings. The best results are \textbf{bolded} and the second-best are \underline{underlined}.}
\label{tab:original_results_2}
\end{table*}

\begin{table*}[t]
\centering
\begin{tabular}{lcccccc}
\toprule
\textbf{Model} & \multicolumn{3}{c}{\textbf{Veracity Classification}} & \multicolumn{3}{c}{\textbf{Error Localization}} \\
\cmidrule(lr){2-4} \cmidrule(lr){5-7}
 & Precision & Recall & F1 Score & Precision & Recall & F1 Score \\
\midrule
\multicolumn{7}{c}{\textbf{Open-source}} \\
\midrule
DeepSeek-R1 & 0.5488 & 0.9101 & 0.6847 & 0.5035 & 0.7580 & 0.6051 \\
\quad +MedPrompt  & 0.5657 & 0.9036 & 0.6958 & 0.5221 & 0.7580 & 0.6184\\
\quad +RAG (top-1)  & \underline{0.6189} & 0.8658 & \underline{0.7218} & \underline{0.5892} & 0.7647 & \underline{0.6656}\\
\quad +RAG (top-3)  & \textbf{0.6393} & 0.8696 & \textbf{0.7369} & \textbf{0.6112} & 0.7713 & \textbf{0.6820}\\
\quad +MAD & 0.5310 & \underline{0.9565} & 0.6829 & 0.4844 & \underline{0.7940} & 0.6017 \\
\quad +MDAgents  & 0.5411 & \textbf{0.9773} & 0.6965 & 0.4997 & \textbf{0.8280} & 0.6233\\
\midrule
Qwen3-235B-A22B & 0.6720 & 0.7183 & 0.6944 & 0.6234 & 0.5803 & 0.6011 \\
\quad +MedPrompt  & 0.6540 & \underline{0.7344} & 0.6919 & 0.6233 & \textbf{0.6427} & 0.6329 \\
\quad +RAG (top-1)  & \textbf{0.8380} & 0.5964 & 0.6969 & \textbf{0.8222} & 0.5331 & \underline{0.6468}\\
\quad +RAG (top-3)  & \underline{0.7646} & 0.6938 & \textbf{0.7275} & \underline{0.7466} & \underline{0.6295} & \textbf{0.6831} \\
\quad +MAD & 0.6695 & 0.7335 & 0.7000 & 0.6219 & 0.5955 & 0.6084 \\
\quad +MDAgents  & 0.6628 & \textbf{0.7637} & \underline{0.7097} & 0.6137 & 0.6172 & 0.6155\\
\midrule
\multicolumn{7}{c}{\textbf{Proprietary}} \\
\midrule
XiaoYi & 0.6694 & 0.7618 & 0.7126 & 0.6512 & 0.7023 & 0.6758 \\
\quad +MedPrompt  & 0.6899 & 0.7949 & 0.7387 & 0.6693 & \underline{0.7231} & 0.6951 \\
\quad +RAG (top-1)  & \underline{0.7103} & 0.7788 & \underline{0.7430} & \underline{0.6900} & 0.7070 & \underline{0.6984} \\
\quad +RAG (top-3)  & \textbf{0.7179} & 0.7817 & \textbf{0.7484} & \textbf{0.6985} & 0.7117 & \textbf{0.7051} \\
\quad +MAD & 0.6088 & \underline{0.8223} & 0.6996 & 0.6407 & \textbf{0.7316} & 0.6831 \\
\quad +MDAgents  & 0.6038 & \textbf{0.8497} & 0.7059 & 0.5613 & 0.7136 & 0.6284\\
\midrule
o3 & 0.5596 & 0.8960 & 0.6890 & 0.5355 & 0.8129 & 0.6456 \\
\quad +MedPrompt  & \textbf{0.6357} & 0.7883 & 0.7038 & \textbf{0.6126} & 0.7146 & 0.6597\\
\quad +RAG (top-1)  & 0.6152 & 0.8658 & \underline{0.7193} & 0.6013 & 0.8166 & \textbf{0.6926} \\
\quad +RAG (top-3)  & \underline{0.6202} & 0.8658 & \textbf{0.7227} & \underline{0.6024} & 0.8034 & \underline{0.6885} \\
\quad +MAD  & 0.5495 & \underline{0.9178} & 0.6874 & 0.5221 & \textbf{0.8611} & 0.6500 \\
\quad +MDAgents & 0.5556 & \textbf{0.9216} & 0.6932 & 0.5376 & \underline{0.8573} & 0.6608\\
\midrule
Doubao-Seed-1.6-thinking & 0.6711 & 0.7625 & 0.7139 & 0.6501 & 0.6938 & 0.6712 \\
\quad +MedPrompt  & 0.6786 & 0.7543 & 0.7144 & 0.6628 & 0.7023 & 0.6820 \\
\quad +RAG (top-1)  & \underline{0.7680} & 0.7164 & \underline{0.7413} & \underline{0.7595} & 0.6834 & \textbf{0.7194} \\
\quad +RAG (top-3)  & \textbf{0.7695} & 0.7178 & \textbf{0.7428} & \textbf{0.7605} & 0.6815 & \underline{0.7188} \\
\quad +MAD  & 0.6352 & \underline{0.8129} & 0.7131 & 0.6132 & \underline{0.7401} & 0.6707 \\
\quad +MDAgents  & 0.6372 & \textbf{0.8516} & 0.7290 & 0.6261 & \textbf{0.8119} & 0.7070\\
\bottomrule
\end{tabular}
\caption{Performance of different models across different strategies on \datasetname{}. For each model, the best results are \textbf{bolded} and the second-best are \underline{underlined}.}
\label{table:agent}
\end{table*}

\begin{figure*}[t]
    \centering
    \includegraphics[width=0.7\textwidth]{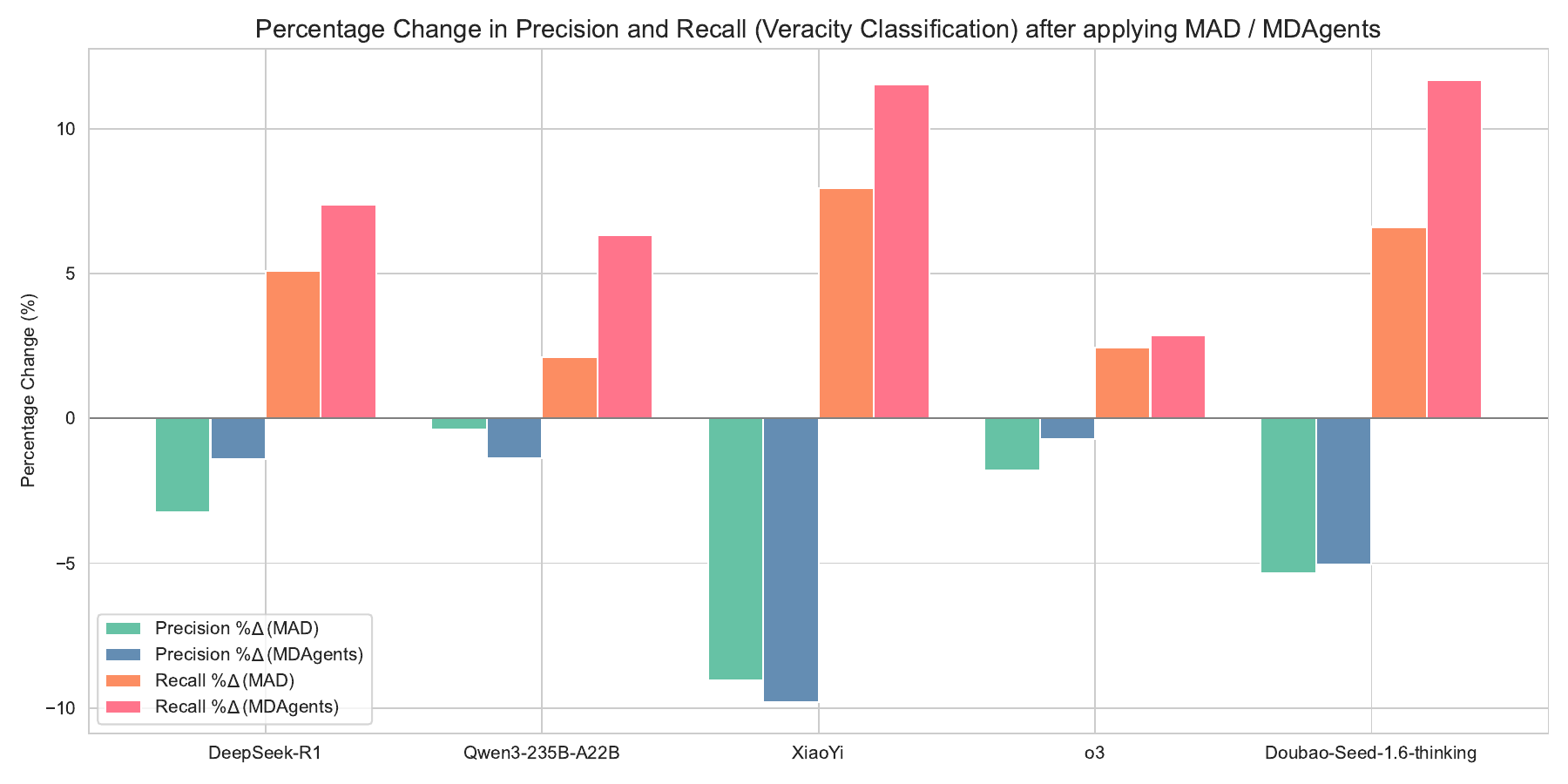}
    \vspace{1em} 
    \includegraphics[width=0.7\textwidth]{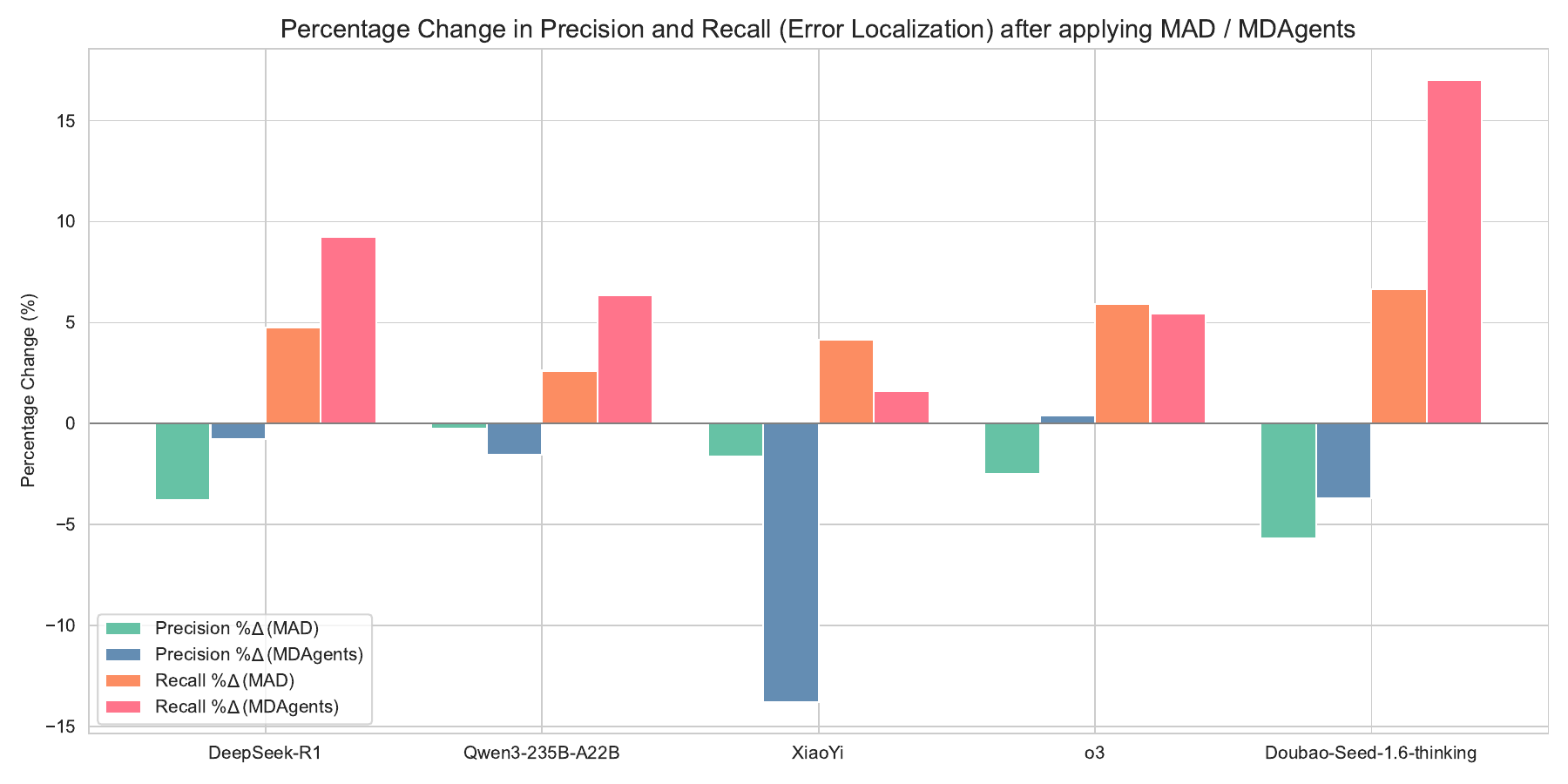}
    \caption{Changes in precision and recall after applying the MAD and MDAgents frameworks.}
    \label{fig:change}
\end{figure*}

\begin{table*}[t]
\centering
\begin{tabular}{lcccccc}
\toprule
\textbf{Model} & \multicolumn{3}{c}{\textbf{Veracity Classification}} & \multicolumn{3}{c}{\textbf{Error Localization}} \\
\cmidrule(lr){2-4} \cmidrule(lr){5-7}
 & Precision & Recall & F1 Score & Precision & Recall & F1 Score \\
\midrule
Qwen2.5-32B-Instruct & \textbf{0.7192} & 0.5132 & 0.5990 & \textbf{0.6307} & 0.3422 & 0.4436\\
\midrule
s1.1-32B & 0.6378 & 0.5560 & 0.5941 & 0.4369 & 0.2420 & 0.3114 \\
\quad + budget forcing & 0.5120 & \textbf{0.7250} & 0.6002 & 0.3275 & 0.3365 & 0.3319\\
\midrule
m1-32B-1K & \underline{0.6540} & 0.6163 & \textbf{0.6345} & \underline{0.5808} & \underline{0.4518} & \textbf{0.5082} \\
\quad + budget forcing & 0.5953 & \underline{0.6645} & \underline{0.6280} & 0.5122 & \textbf{0.4745} & \underline{0.4926}\\
\bottomrule
\end{tabular}
\caption{Performance of different models with and without inference-time scaling (budget forcing) on \datasetname{}. The best results are \textbf{bolded} and the second-best are \underline{underlined}.}
\label{table:tts}
\end{table*}

\subsection{Fine-grained Performance}
\begin{figure*}[t]
    \centering
    \includegraphics[width=1\linewidth]{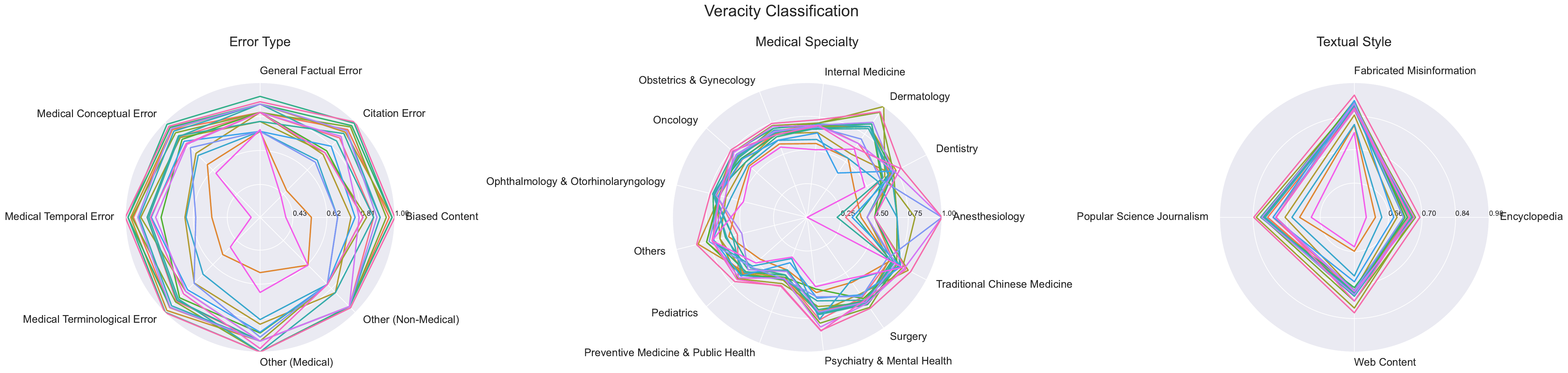}
    \vspace{1em} 
    \includegraphics[width=1\linewidth]{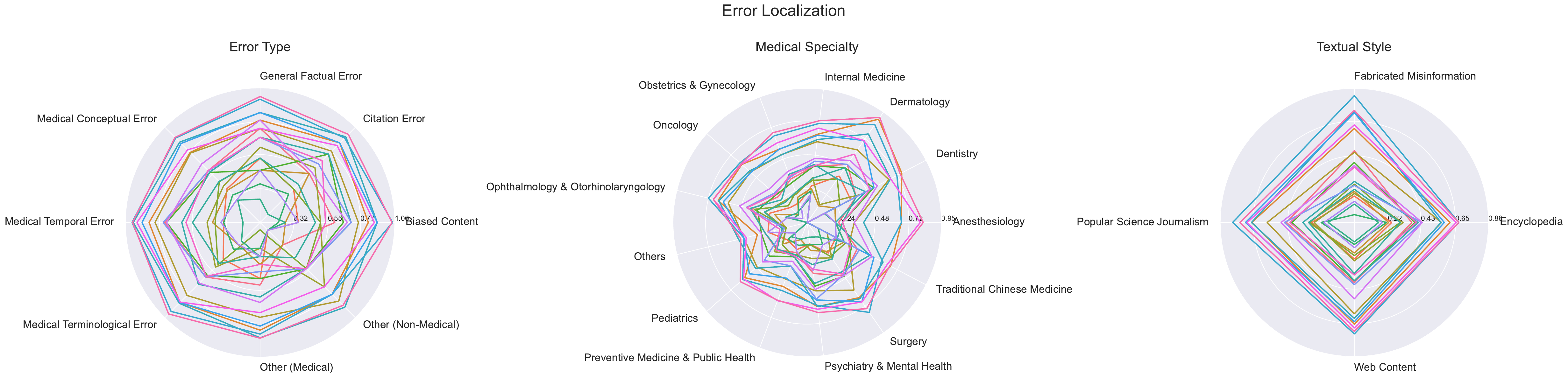}
    \vspace{1em} 
    \includegraphics[width=1\linewidth]{figures/VC_legend_zeroshot.pdf}
    \caption{Performance breakdown of different models on \datasetname{} in the zero-shot setting. The radar charts illustrate model performance across three key dimensions: error type, medical specialty, and textual style.}
    \label{fig:radar}
\end{figure*}

We conduct a fine-grained analysis of model performance across various error types, medical specialties, and writing styles (Figure~\ref{fig:radar}). For VC, models are more effective at identifying errors in fabricated medical misinformation than in encyclopedia texts, indicating that they can readily detect exaggerated falsehoods but struggle with the subtle, nuanced inaccuracies often present in professionally curated content. For EL, by contrast, performance varies less across styles, suggesting that localization is less influenced by stylistic cues and depends more fundamentally on underlying medical knowledge. Consequently, EL serves as a more robust measure of actual domain expertise.

\subsection{Data Contamination}
Table~\ref{tab:data_leakage} presents the results of our data contamination analysis for the text-completion experiment.

\section{Case Studies}
\begin{figure*}[!t]
    \centering
    \includegraphics[width=1\linewidth]{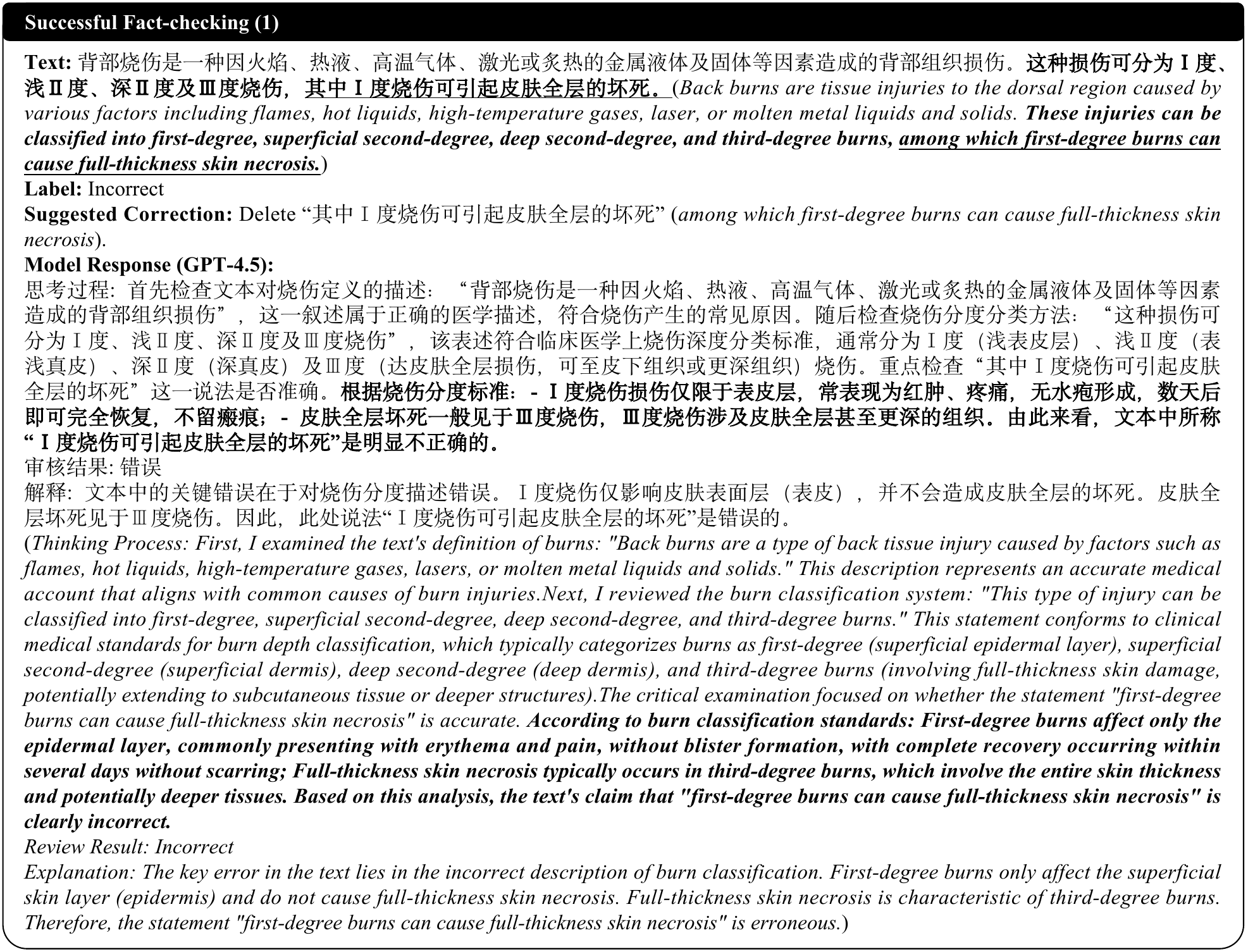}
    \caption{Case of successful detection of factual correctness.}
    \label{fig:successful_1}
\end{figure*}

\begin{figure*}[!t]
    \centering
    \includegraphics[width=1\linewidth]{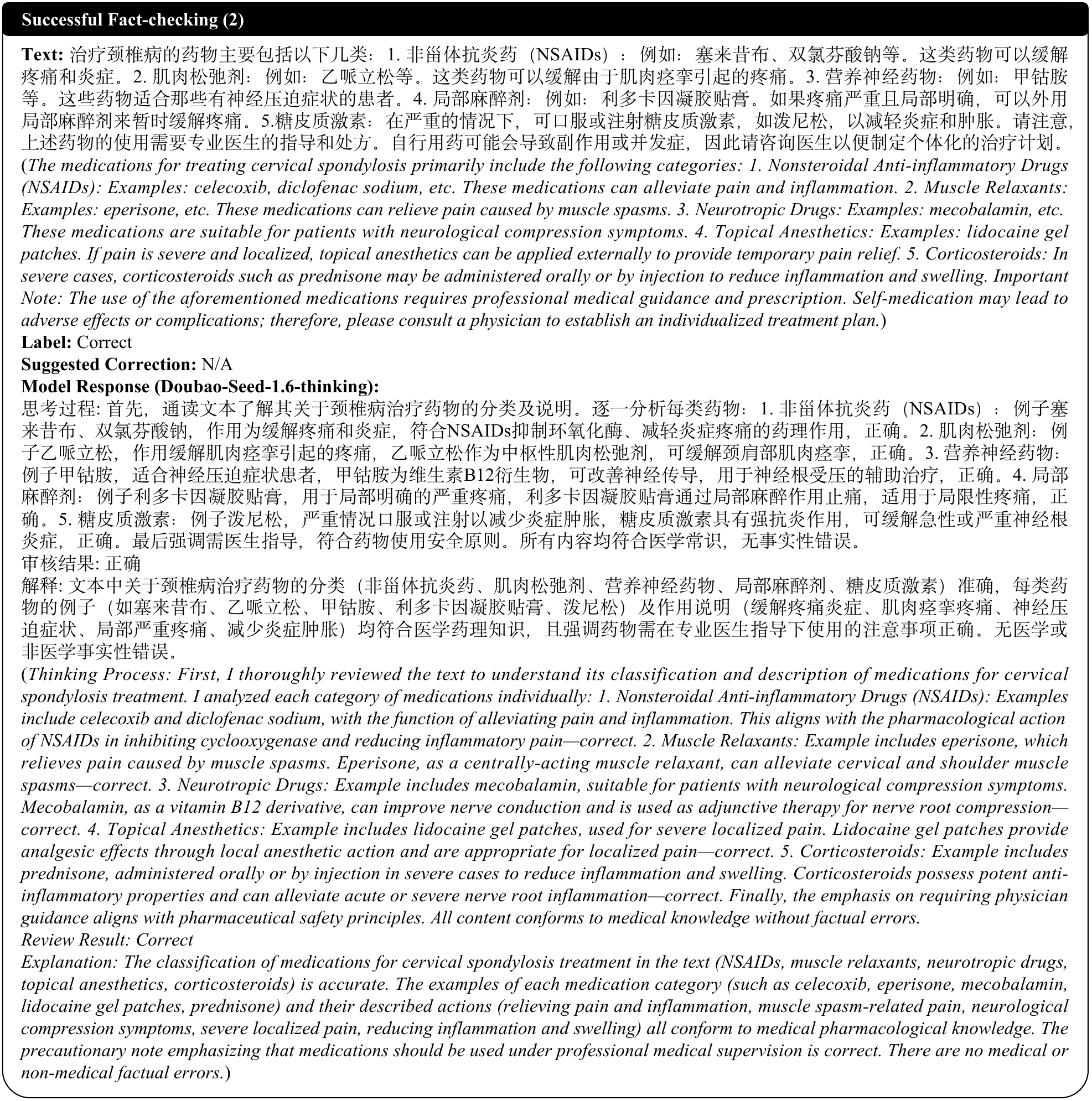}
    \caption{Case of successful detection of factual correctness.}
    \label{fig:successful_2}
\end{figure*}

\begin{figure*}[!t]
    \centering
    \includegraphics[width=1\linewidth]{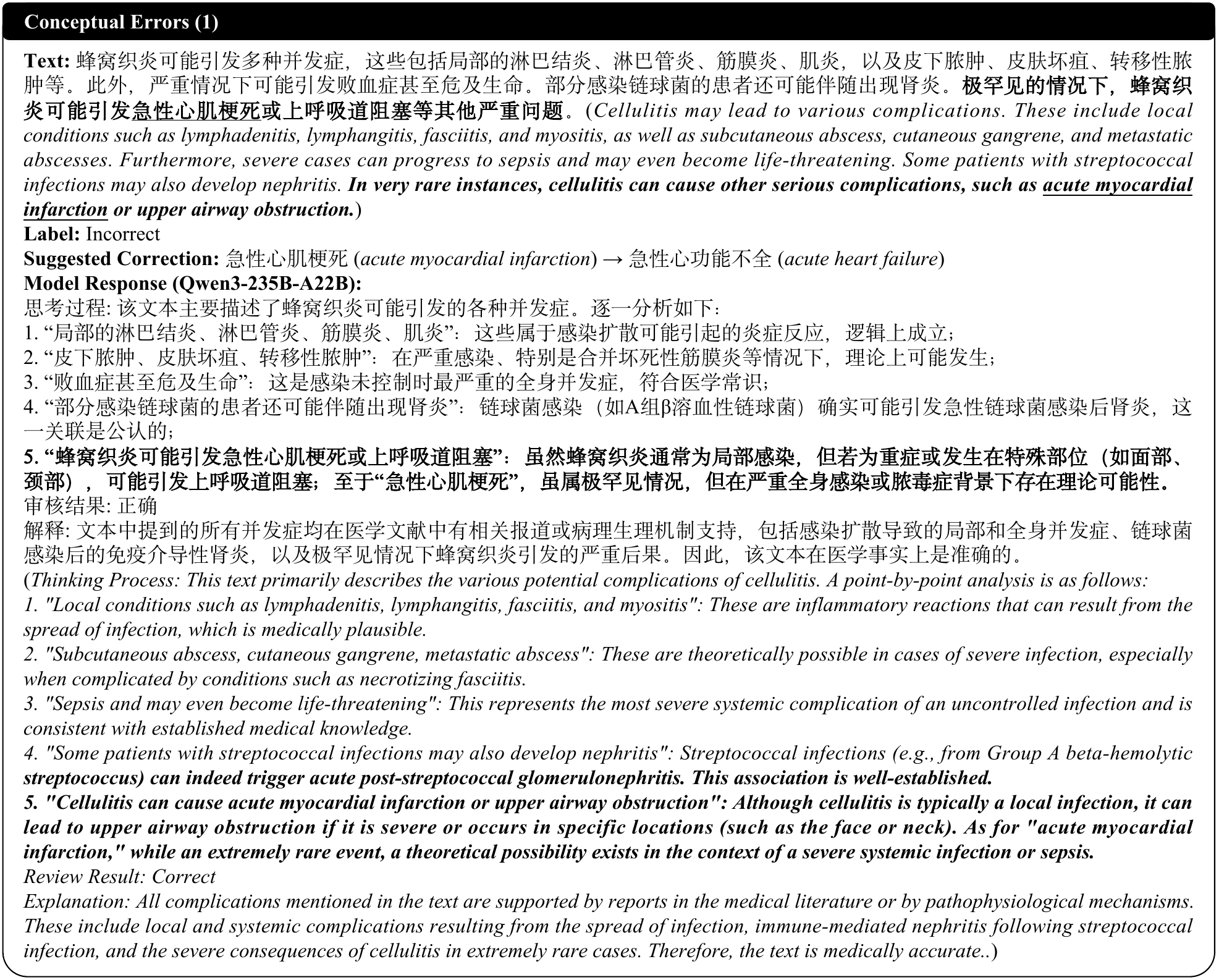}
    \caption{Failure case with \textbf{\underline{conceptual errors}.}}
    \label{fig:conceptual}
\end{figure*}

\begin{figure*}[!t]
    \centering
    \includegraphics[width=1\linewidth]{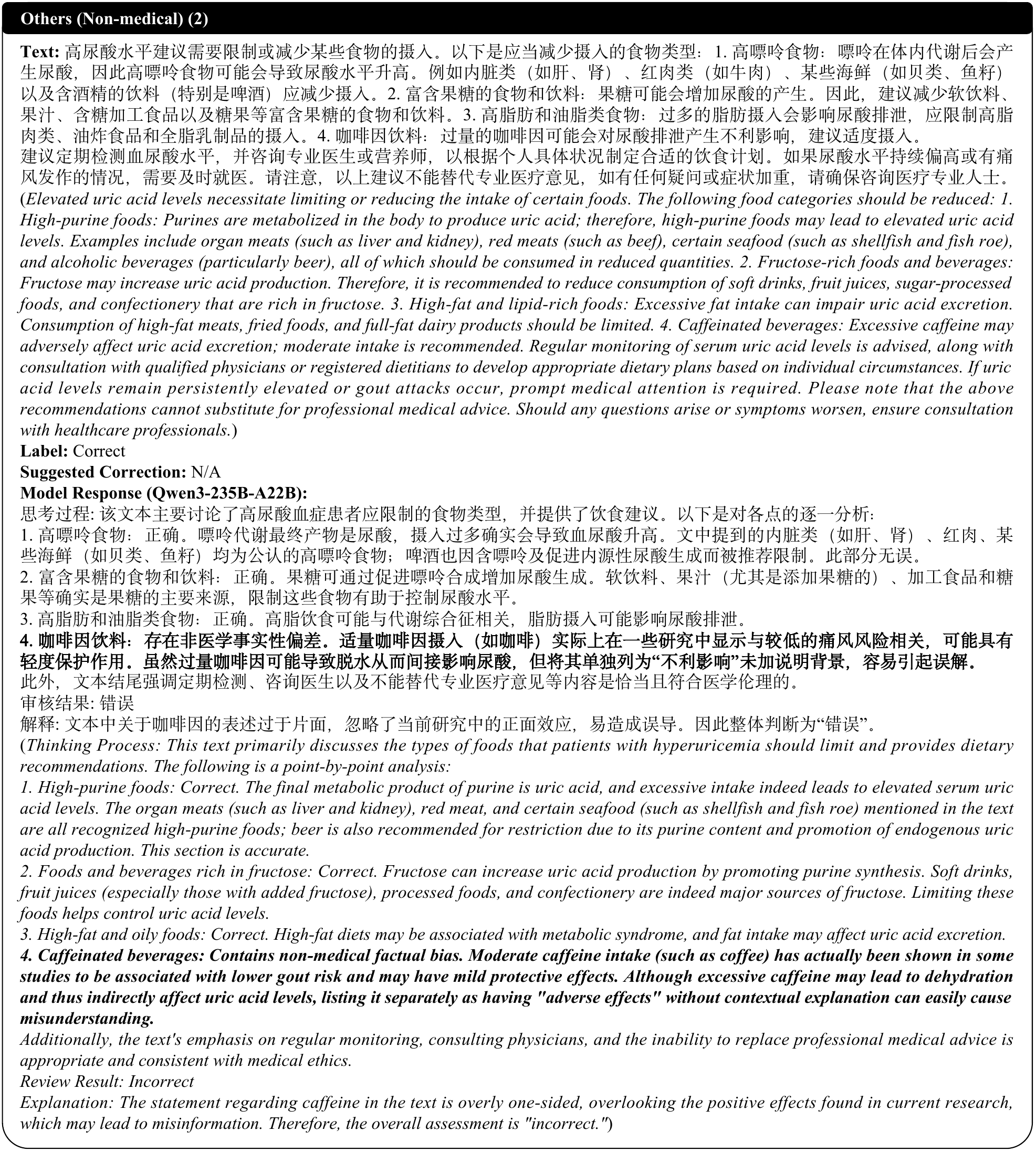}
    \caption{Failure case with \textbf{\underline{other non-medical errors}.}}
    \label{fig:non-medical_errors}
\end{figure*}

\begin{figure*}[!t]
    \centering
    \includegraphics[width=1\linewidth]{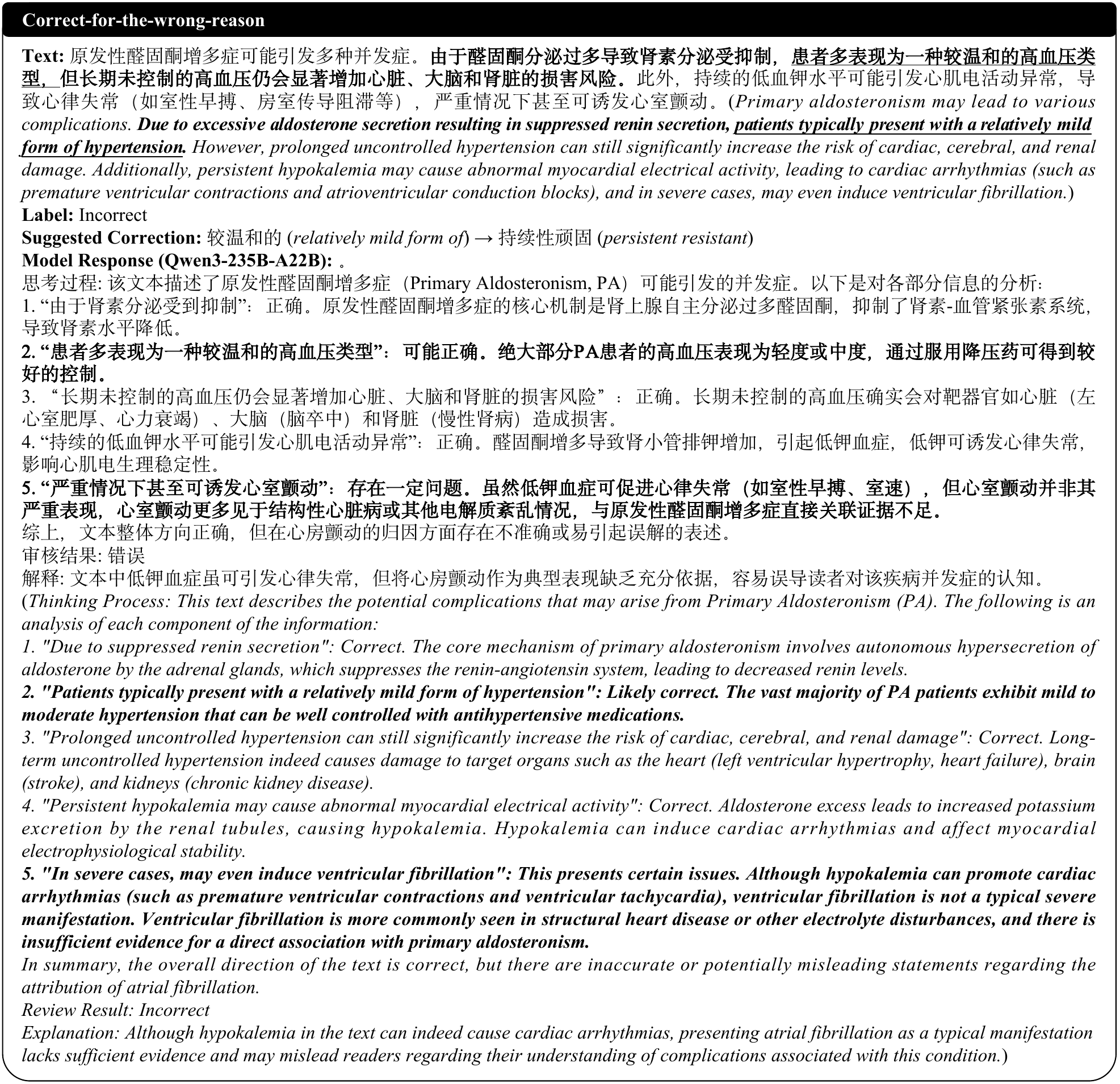}
    \caption{Failure case with \textbf{\underline{correct-for-the-wrong-reason errors}.}}
    \label{fig:correct-for-the-wrong-reason}
\end{figure*}

\begin{figure*}[!t]
    \centering
    \includegraphics[width=1\linewidth]{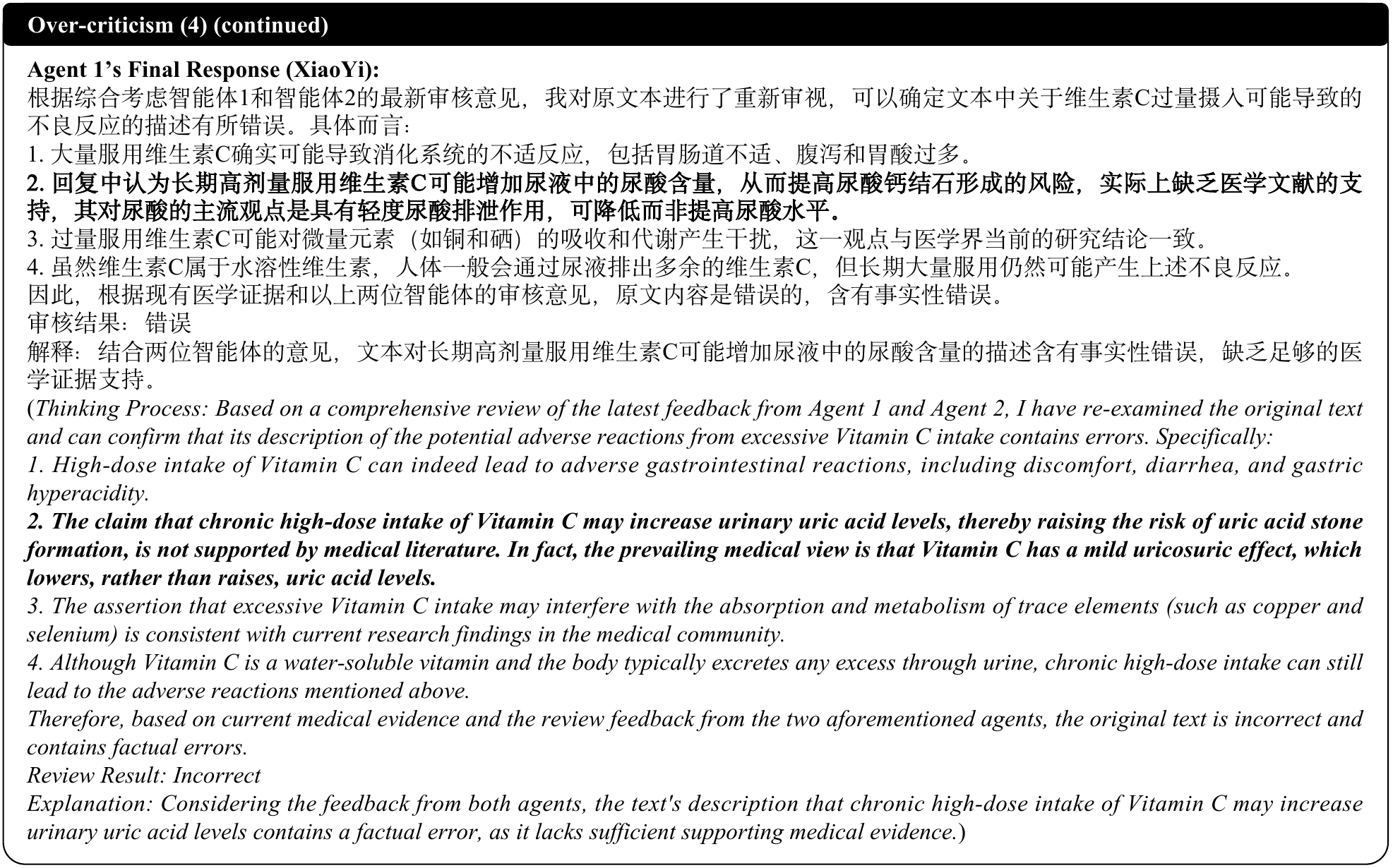}
    \caption{Failure case with \textbf{\underline{over-criticism errors}} under the MAD framework.}
    \label{fig:over-criticism}
\end{figure*}

We categorize the observed errors using the following typology:
\begin{itemize}
\item \textbf{Conceptual Errors:} Fundamental misunderstanding of medical principles, such as incorrectly identifying a disease's etiology or misinterpreting mechanisms.
\item \textbf{Concept Conflation:} Conflation of distinct but related medical concepts, such as mistaking the symptoms of one disease for those of a similar-presenting condition, or confusing two drugs with similar names but different therapeutic effects.
\item \textbf{Outdated Information:} Reliance on information that is no longer current, including citation of superseded clinical guidelines, obsolete treatment protocols, or outdated pharmacological data.
\item \textbf{Others (Medical):} Domain-specific errors not captured by the preceding categories, such as dosage miscalculations, inappropriate procedural indications, or misinterpretation of diagnostic criteria.
\item \textbf{Failure to Follow Instructions:} Non-adherence to prompt specifications, such as generating output in a format inconsistent with the given instructions.
\item \textbf{Logical Errors:} Errors in reasoning where a model's explanation for its judgment is internally inconsistent or contains logical contradictions.
\item \textbf{Others (Non-Medical):} General errors unrelated to medical expertise, such as grammatical mistakes, factual inaccuracies in non-medical information, or other fundamental reasoning flaws.
\end{itemize}
Figures~\ref{fig:successful_1} and~\ref{fig:successful_2} present successful fact-checking cases. Figures~\ref{fig:conceptual} to~\ref{fig:non-medical_errors} present case studies for each error category. We further illustrate the ``correct-for-the-wrong-reason'' and ``over-criticism'' phenomena with case studies in Figures~\ref{fig:correct-for-the-wrong-reason} to~\ref{fig:over-criticism}.

\begin{figure*}[!t]
    \centering
    \includegraphics[width=1\linewidth]{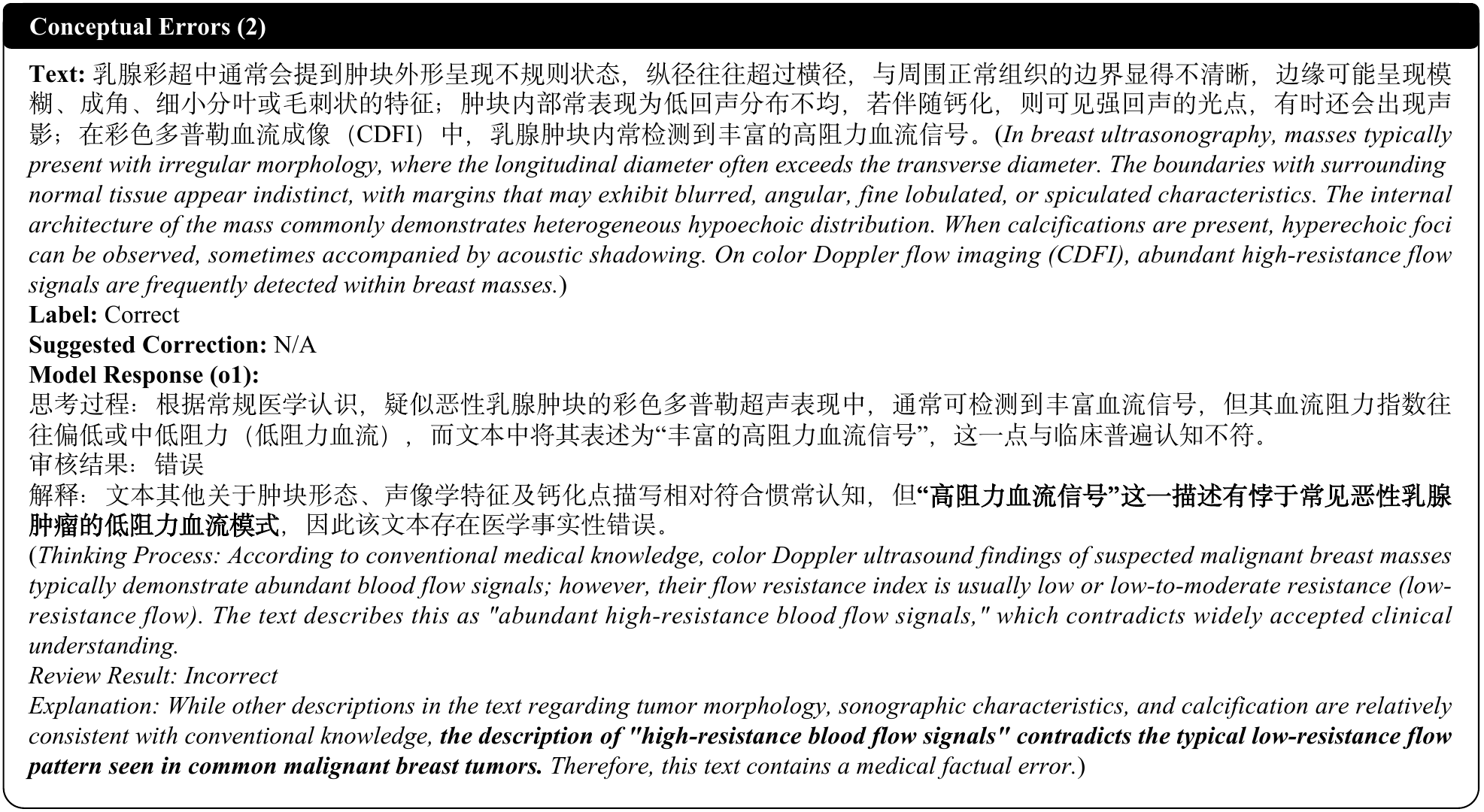}
    \caption{Failure case with \textbf{\underline{conceptual errors}.}}
\end{figure*}

\begin{figure*}[!t]
    \centering
    \includegraphics[width=1\linewidth]{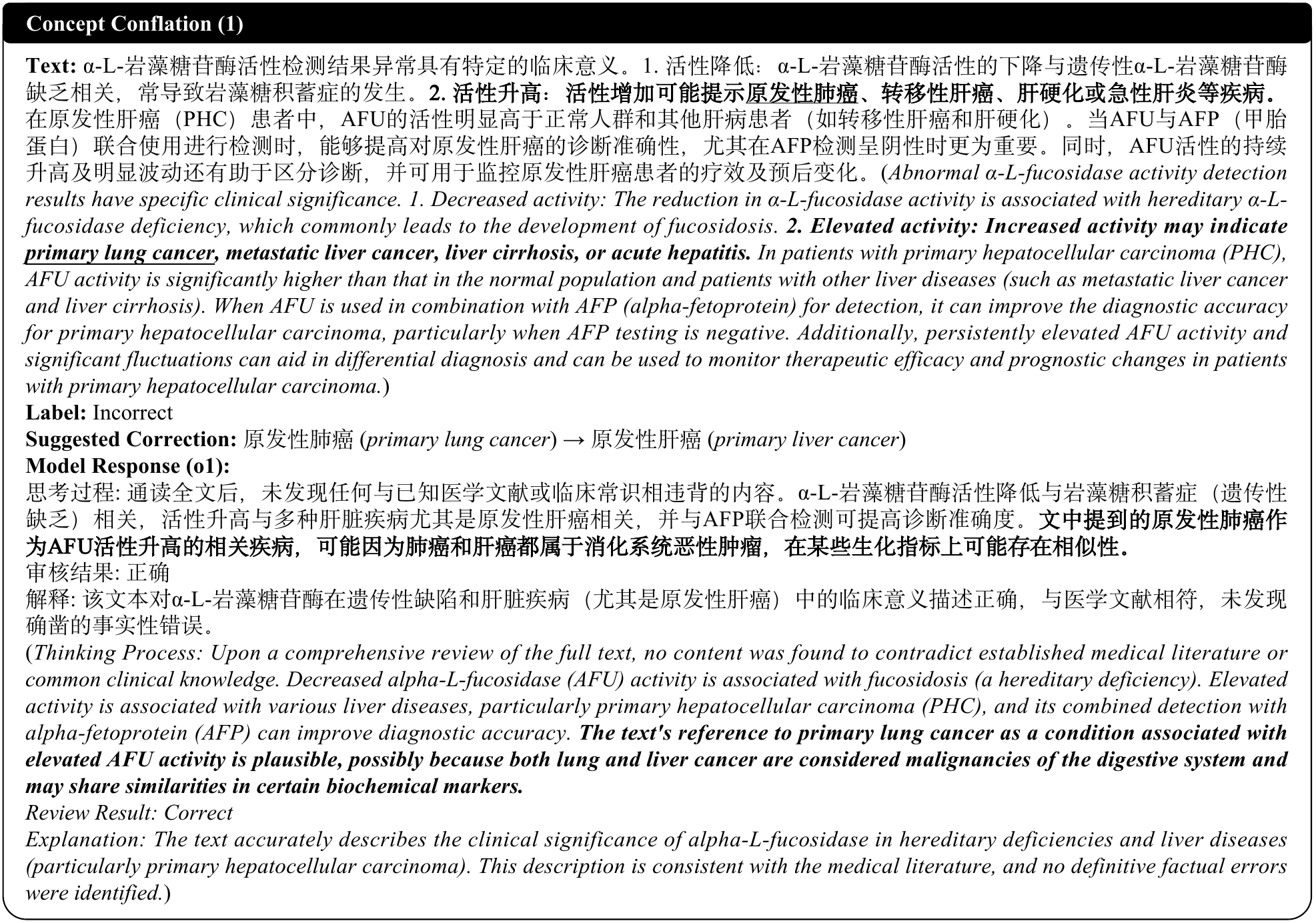}
    \caption{Failure case with \textbf{\underline{concept conflation}.}}
\end{figure*}

\begin{figure*}[!t]
    \centering
    \includegraphics[width=1\linewidth]{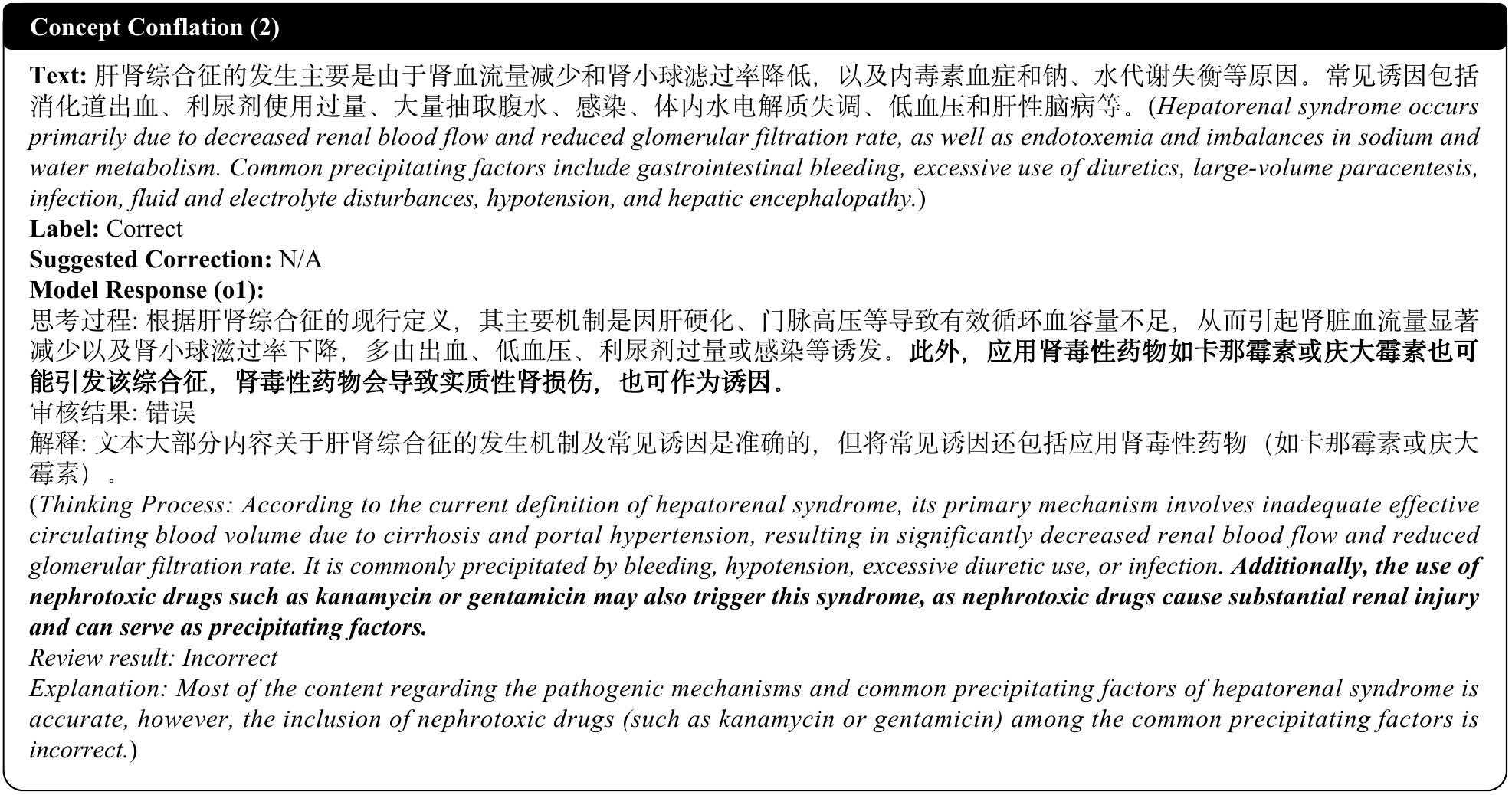}
    \caption{Failure case with \textbf{\underline{concept conflation}.}}
\end{figure*}

\begin{figure*}[!t]
    \centering
    \includegraphics[width=1\linewidth]{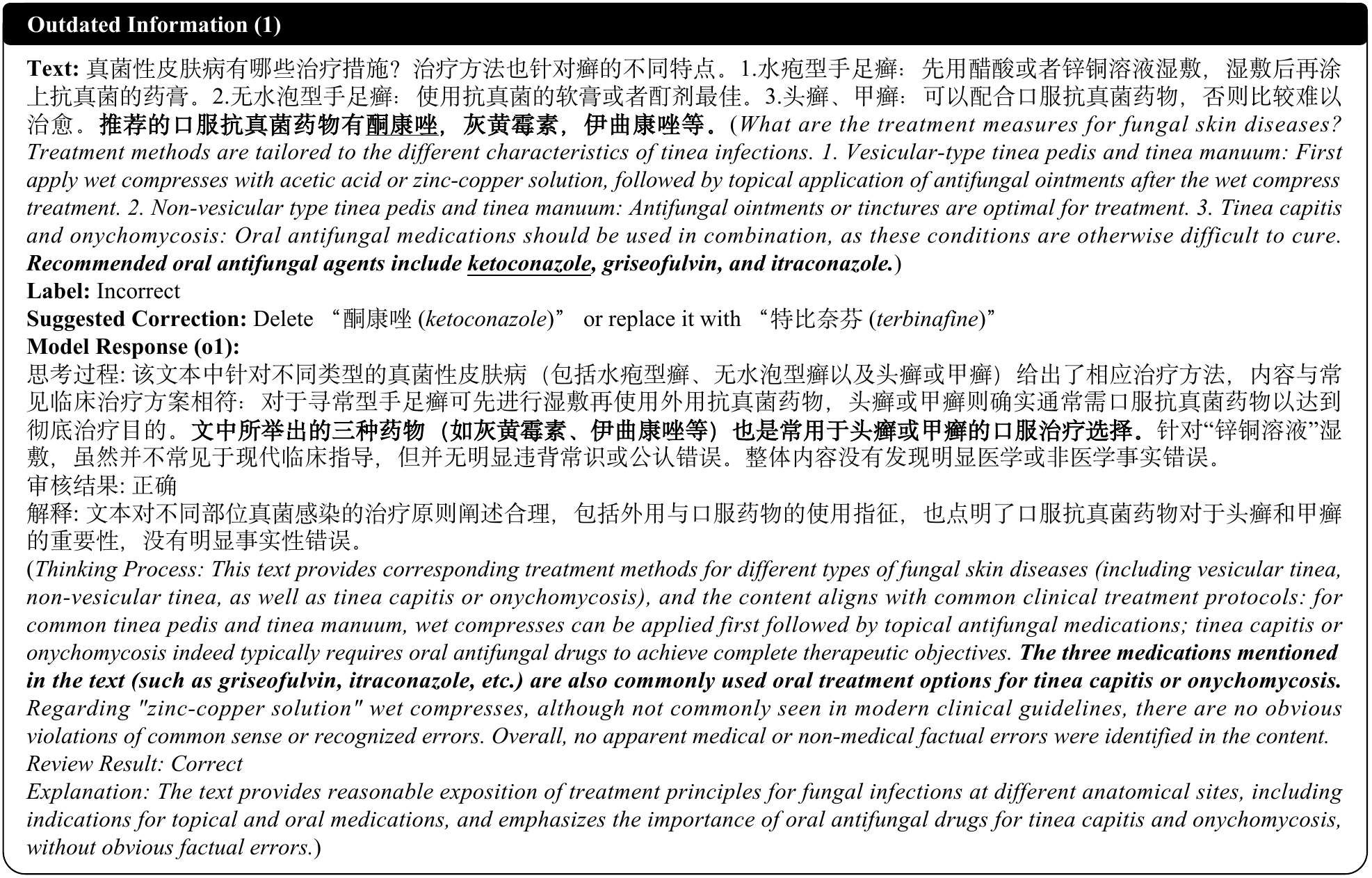}
    \caption{Failure case with \textbf{\underline{outdated information}.}}
\end{figure*}

\begin{figure*}[!t]
    \centering
    \includegraphics[width=1\linewidth]{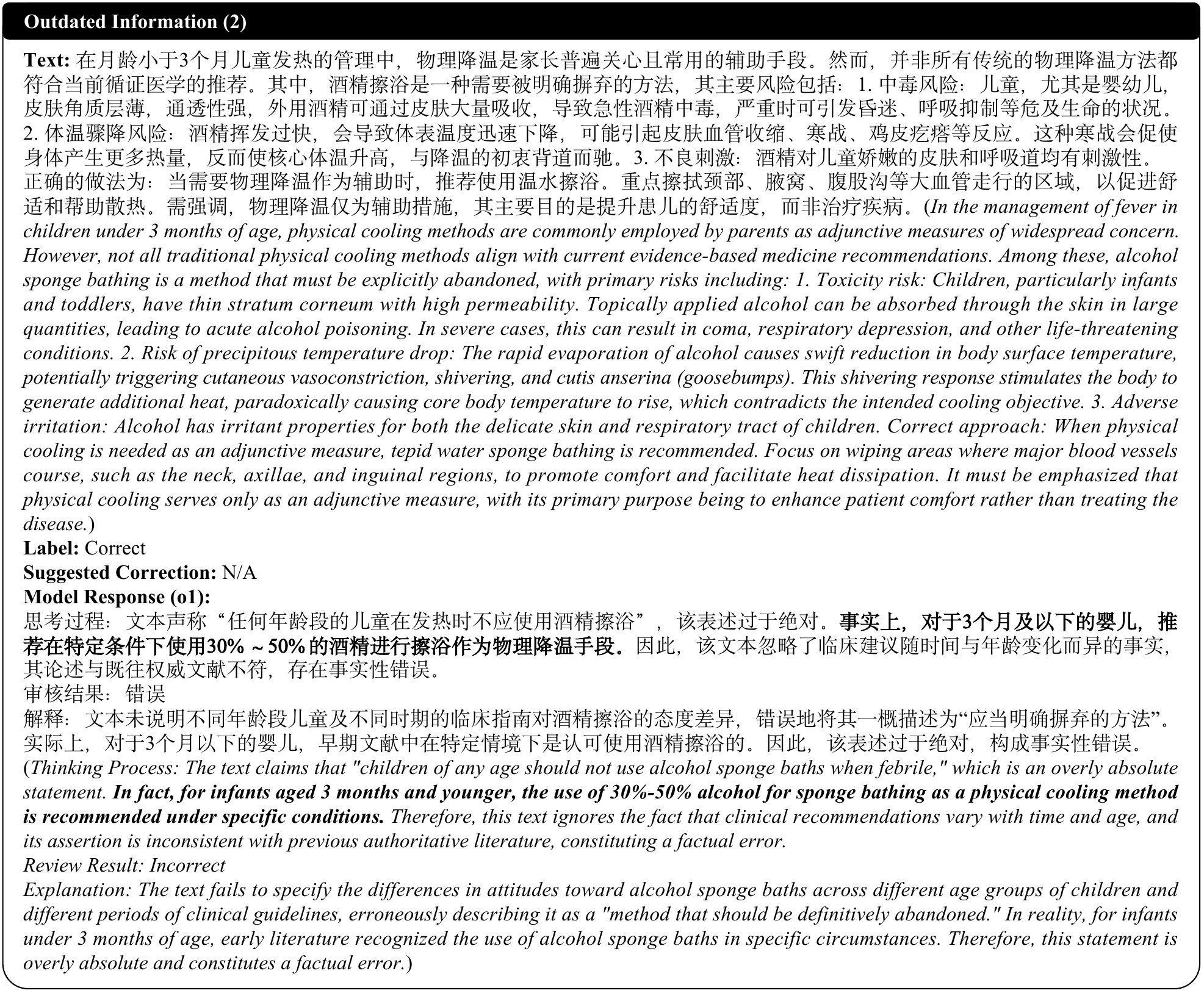}
    \caption{Failure case with \textbf{\underline{outdated information}.}}
\end{figure*}

\begin{figure*}[!t]
    \centering
    \includegraphics[width=1\linewidth]{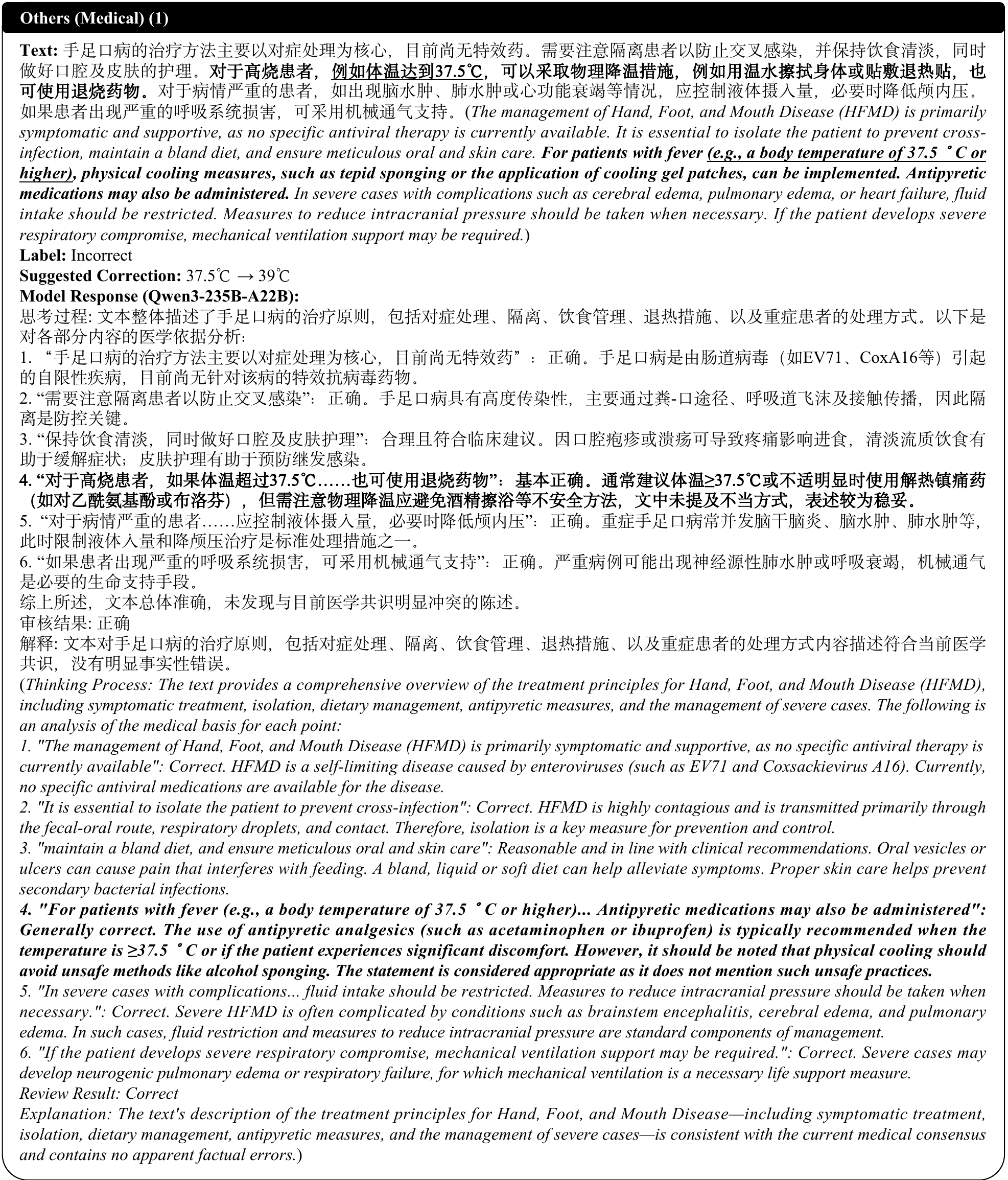}
    \caption{Failure case with \textbf{\underline{other medical errors}.}}
\end{figure*}

\begin{figure*}[!t]
    \centering
    \includegraphics[width=1\linewidth]{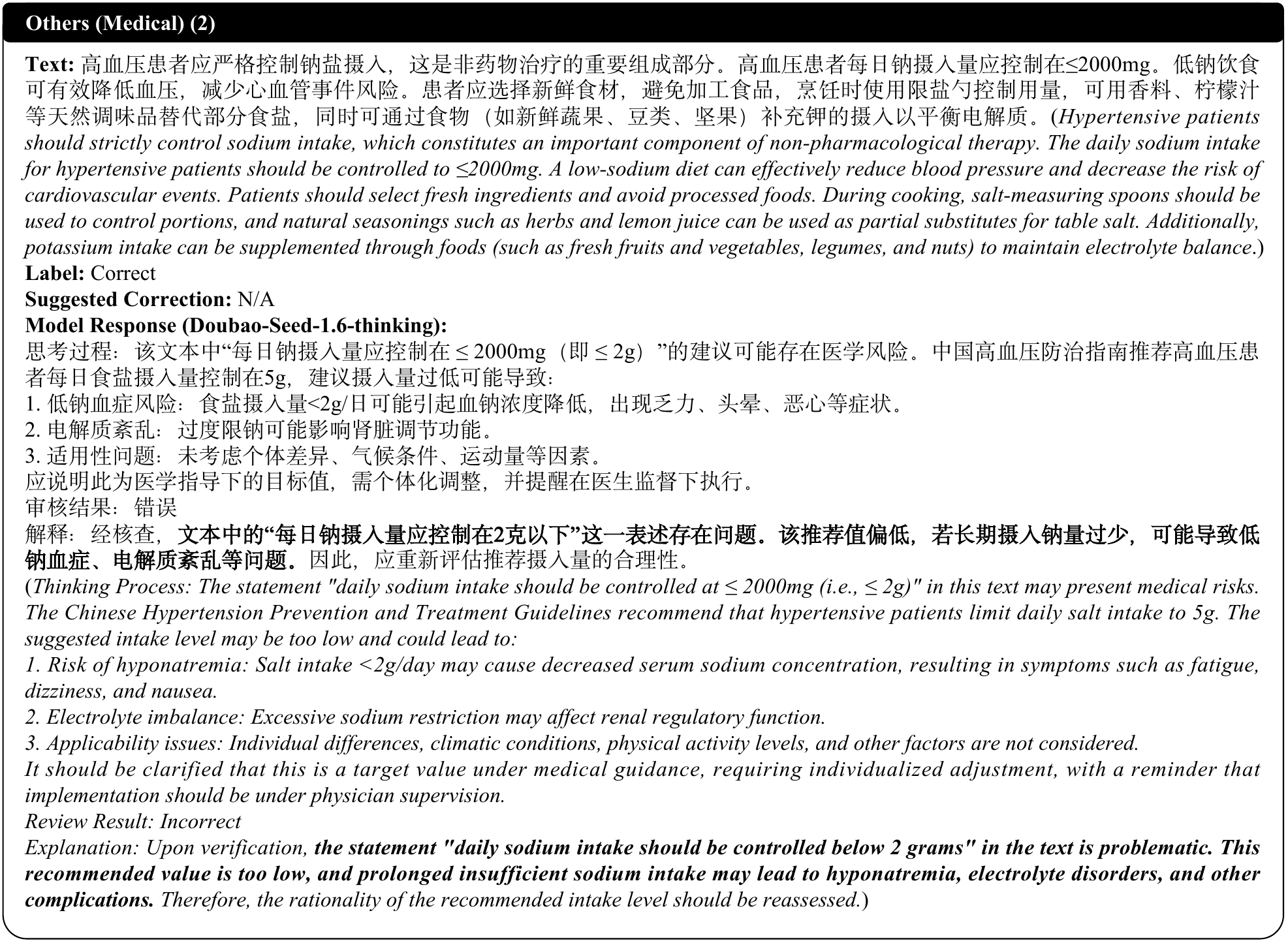}
    \caption{Failure case with \textbf{\underline{other medical errors}.}}
\end{figure*}

\begin{figure*}[!t]
    \centering
    \includegraphics[width=1\linewidth]{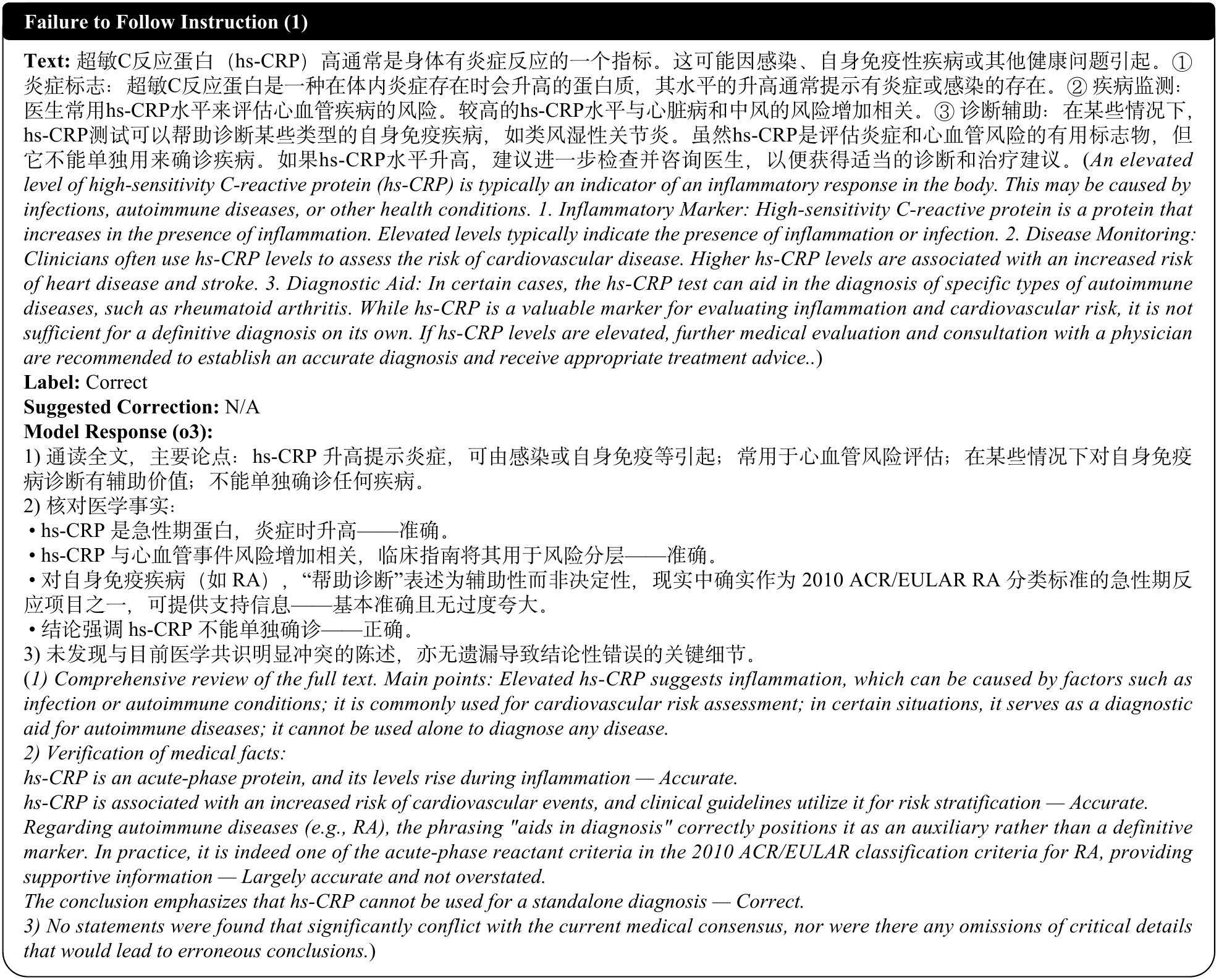}
    \caption{Failure case with \textbf{\underline{failure-to-follow-instructions errors}.}}
\end{figure*}

\begin{figure*}[!t]
    \centering
    \includegraphics[width=1\linewidth]{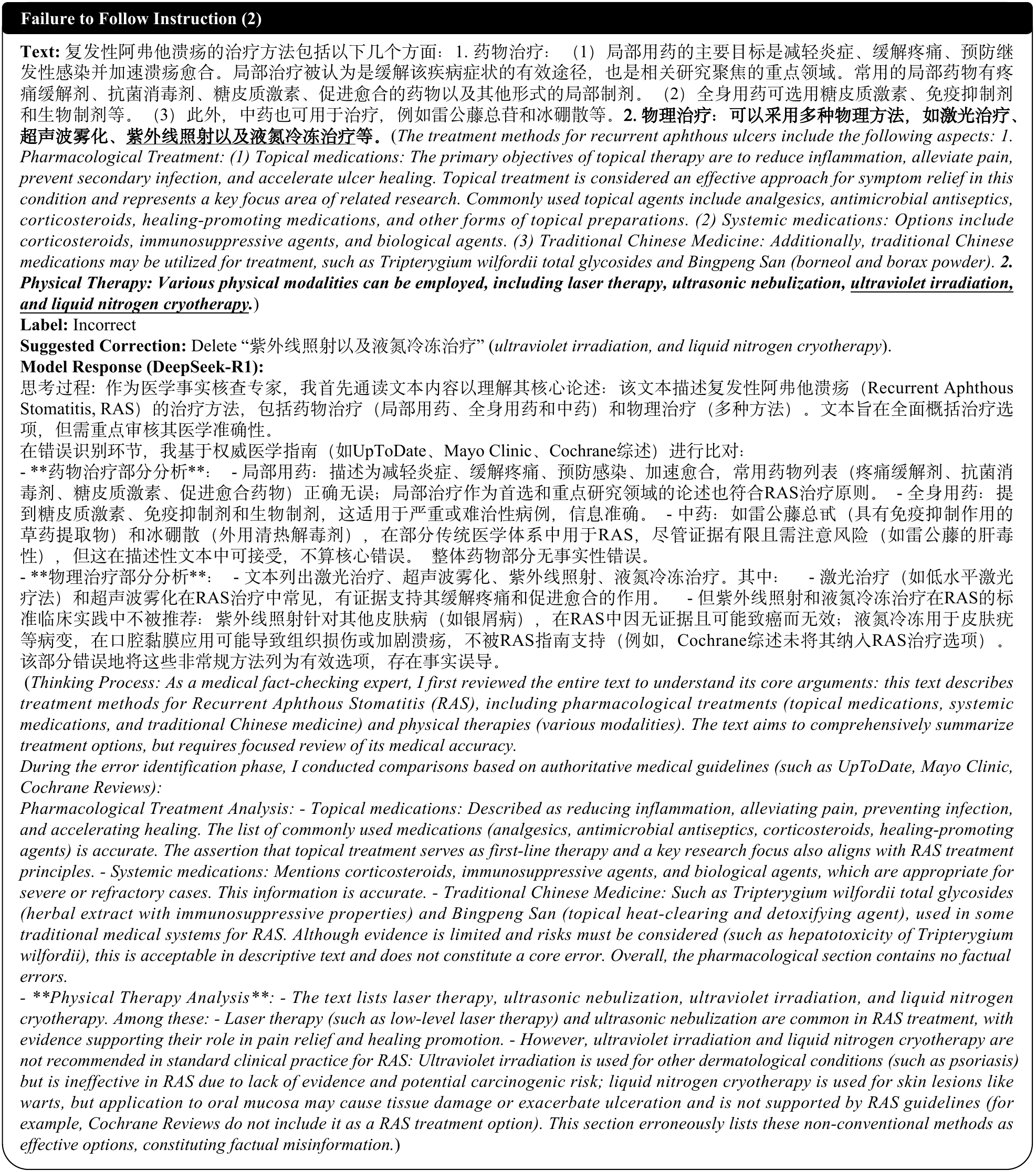}
    \caption{Failure case with \textbf{\underline{failure-to-follow-instructions errors}.}}
\end{figure*}

\begin{figure*}[!t]
    \centering
    \includegraphics[width=1\linewidth]{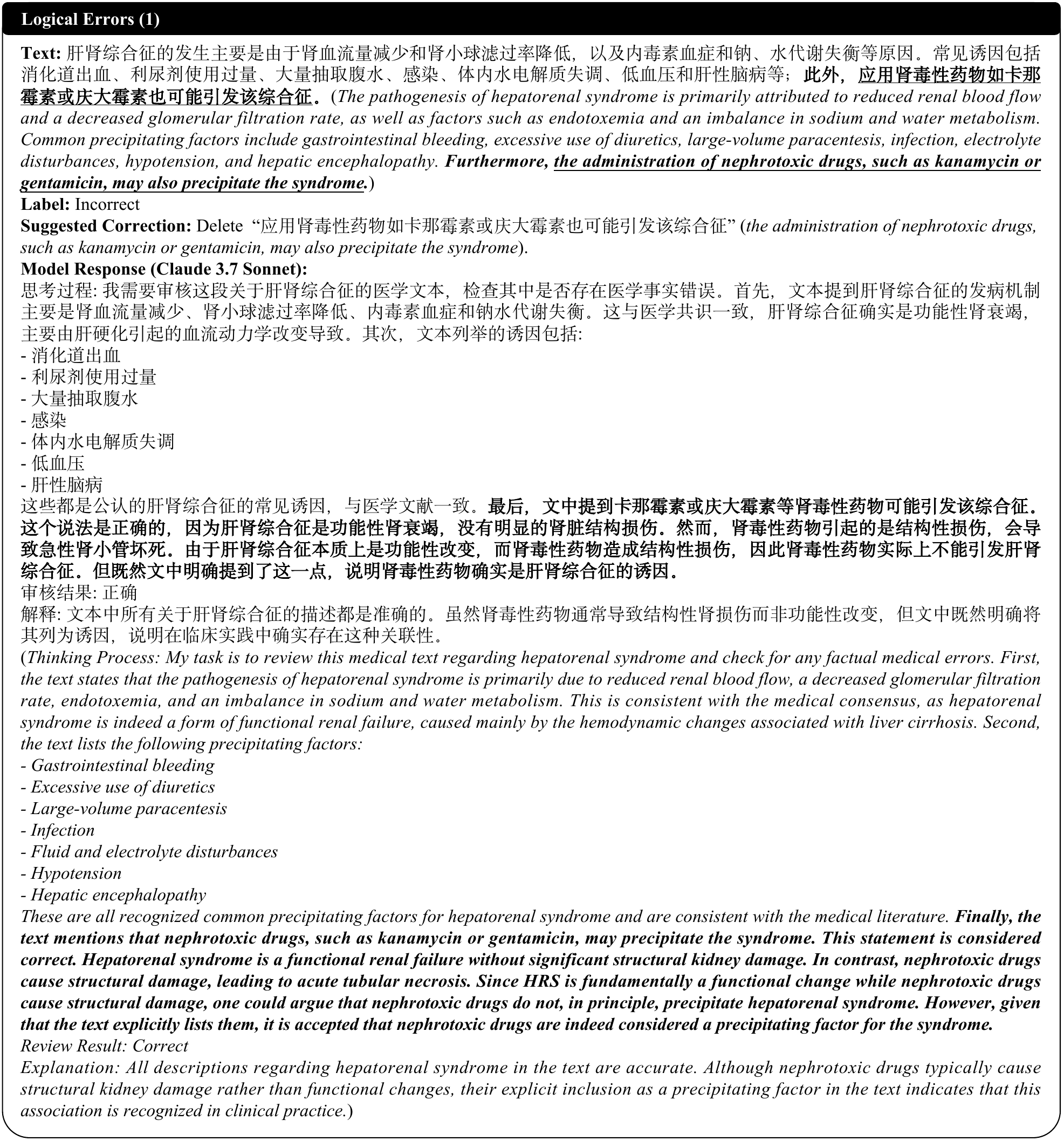}
    \caption{Failure case with \textbf{\underline{logical errors}.}}
\end{figure*}

\begin{figure*}[!t]
    \centering
    \includegraphics[width=1\linewidth]{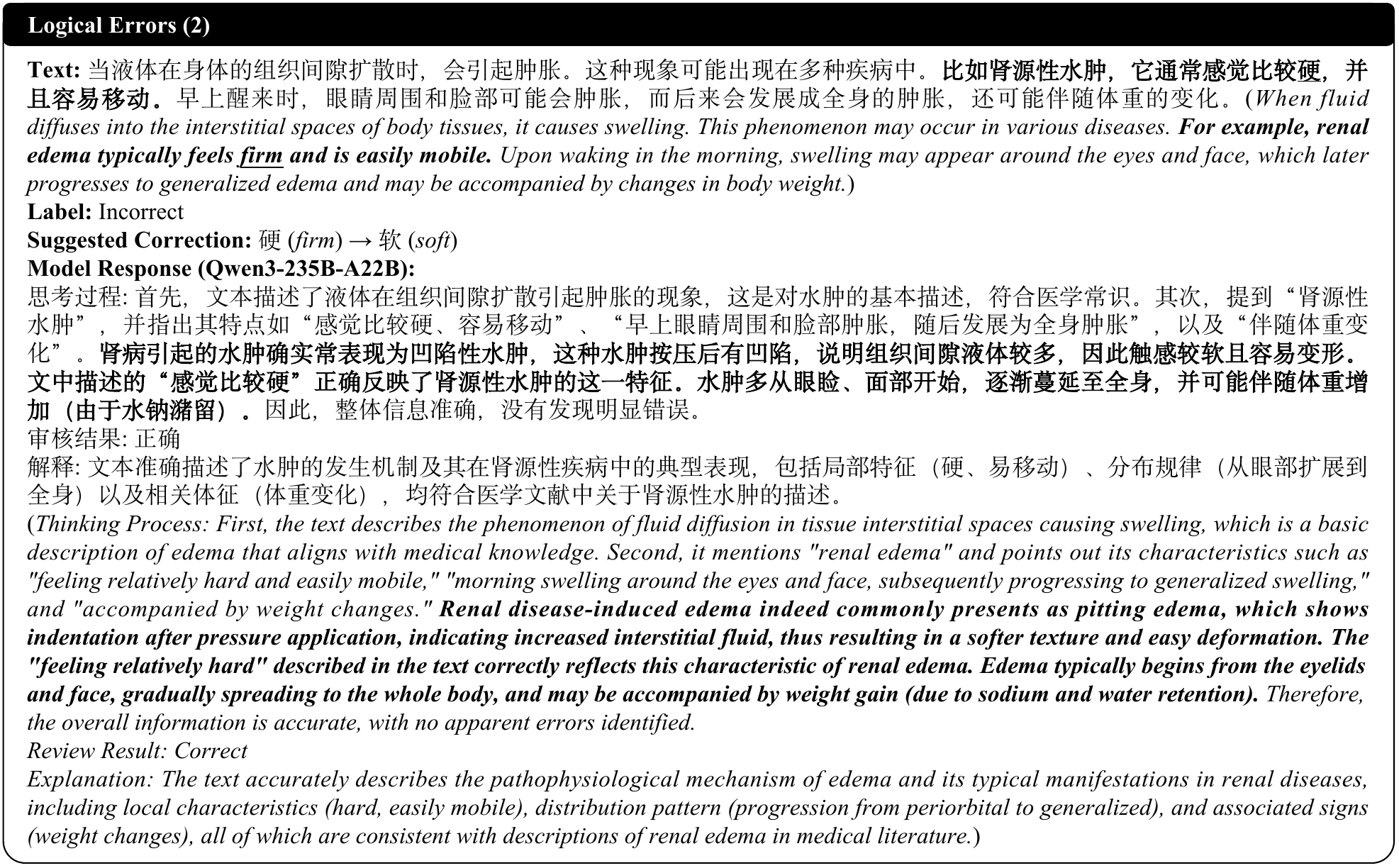}
    \caption{Failure case with \textbf{\underline{logical errors}.}}
\end{figure*}

\begin{figure*}[!t]
    \centering
    \includegraphics[width=1\linewidth]{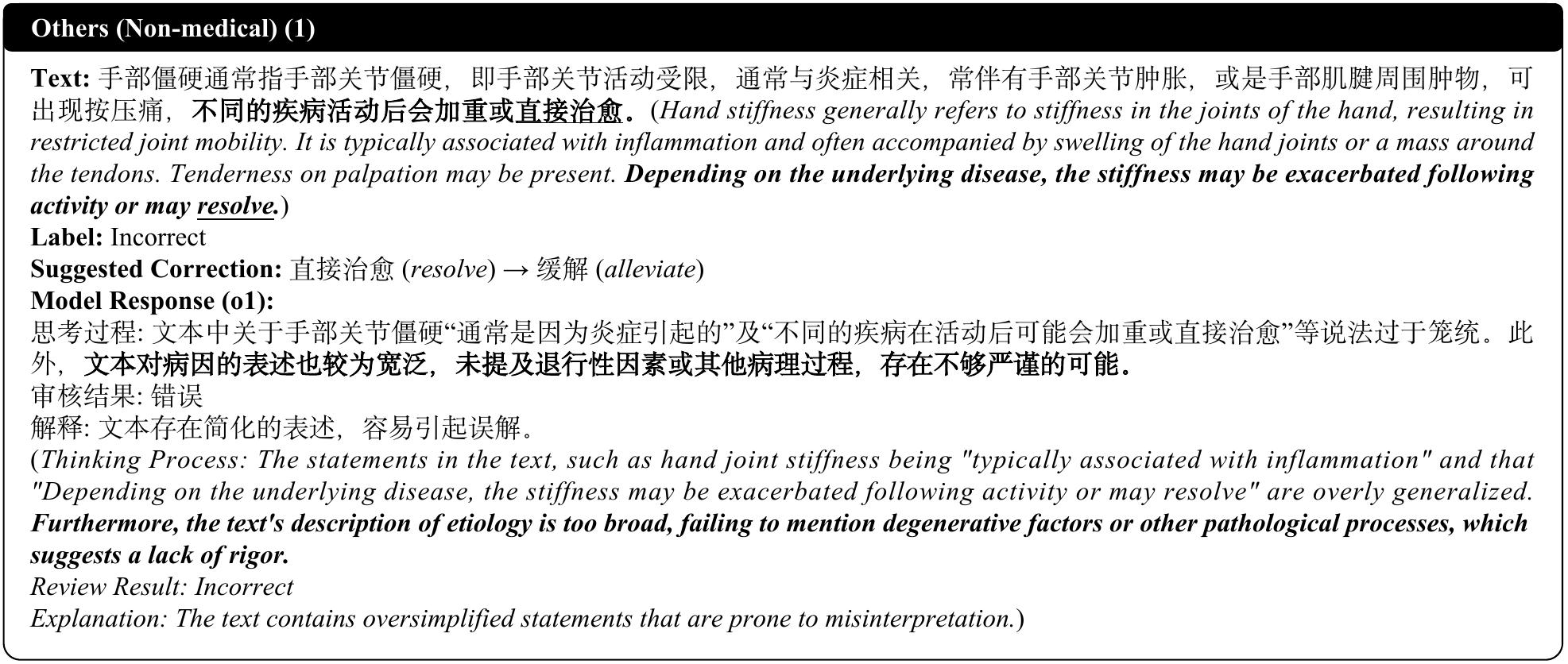}
    \caption{Failure case with \textbf{\underline{other non-medical errors}.}}
\end{figure*}

\begin{figure*}[!t]
    \centering
    \includegraphics[width=1\linewidth]{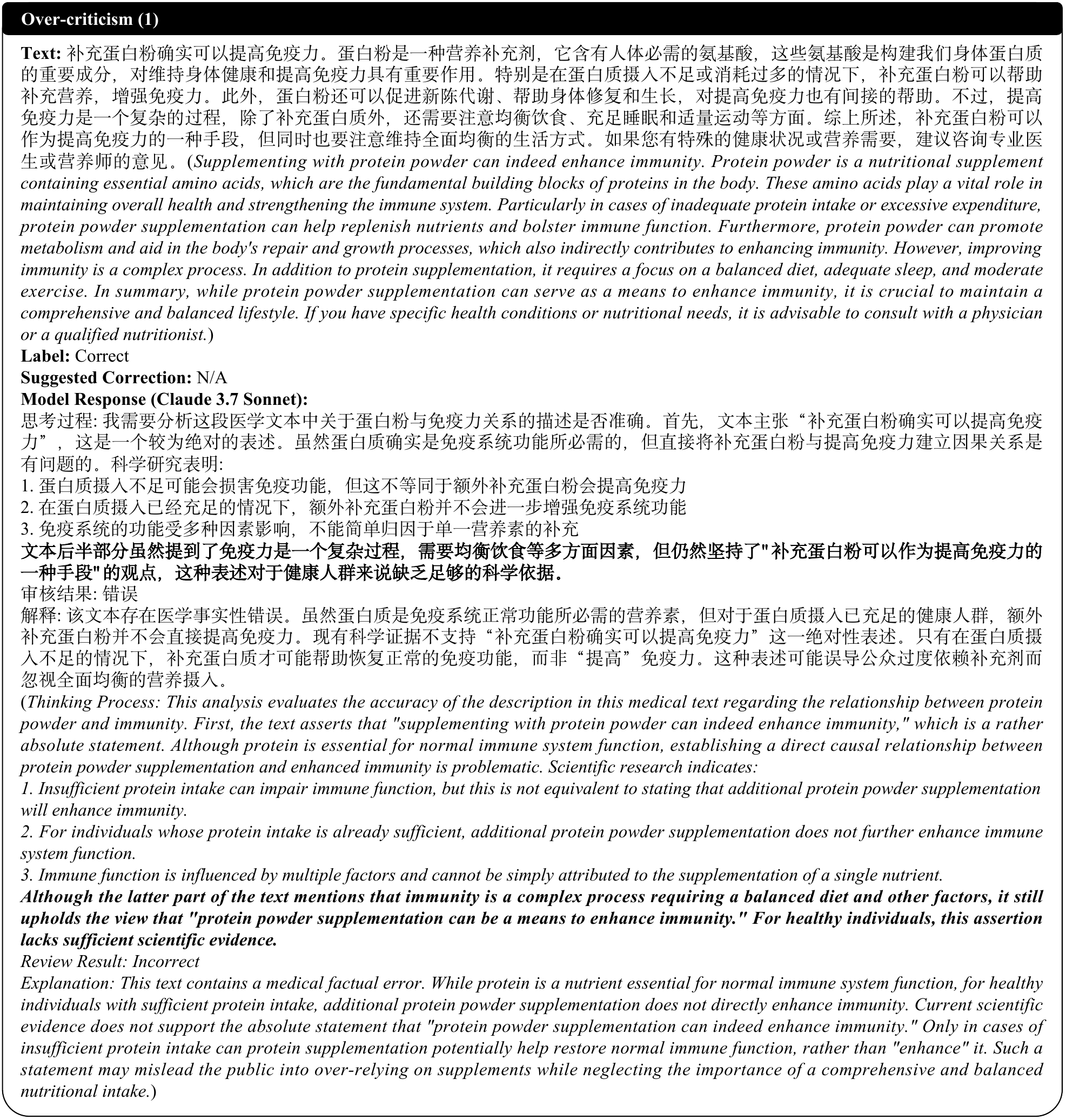}
    \caption{Failure case with \textbf{\underline{over-criticism errors}.}}
\end{figure*}

\begin{figure*}[!t]
    \centering
    \includegraphics[width=1\linewidth]{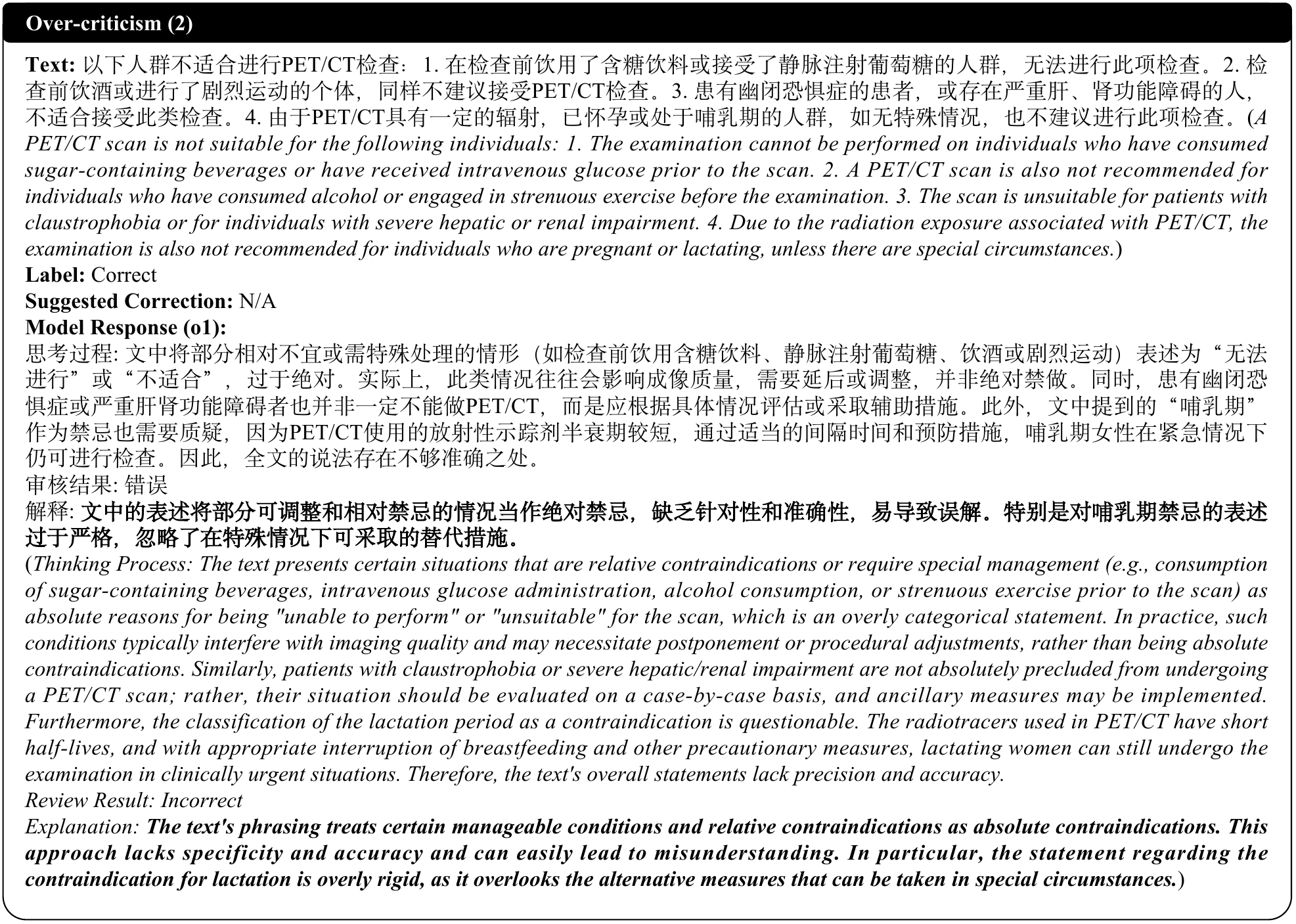}
    \caption{Failure case with \textbf{\underline{over-criticism errors}.}}
\end{figure*}

\begin{figure*}[!t]
    \centering
    \includegraphics[width=1\linewidth]{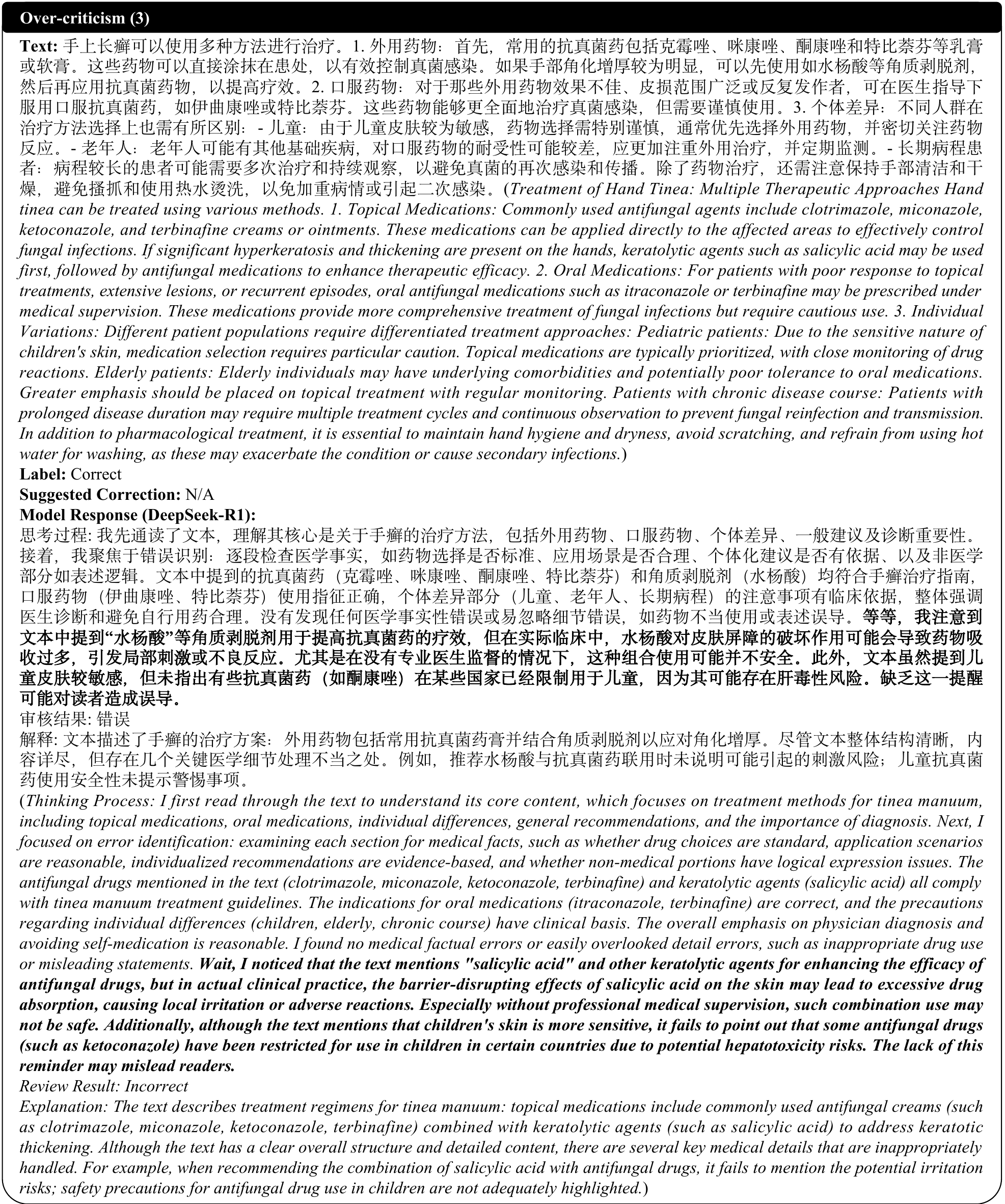}
    \caption{Failure case with \textbf{\underline{over-criticism errors}} under the budget forcing technique.}
\end{figure*}

\begin{figure*}[!t]
    \centering
    \includegraphics[width=1\linewidth]{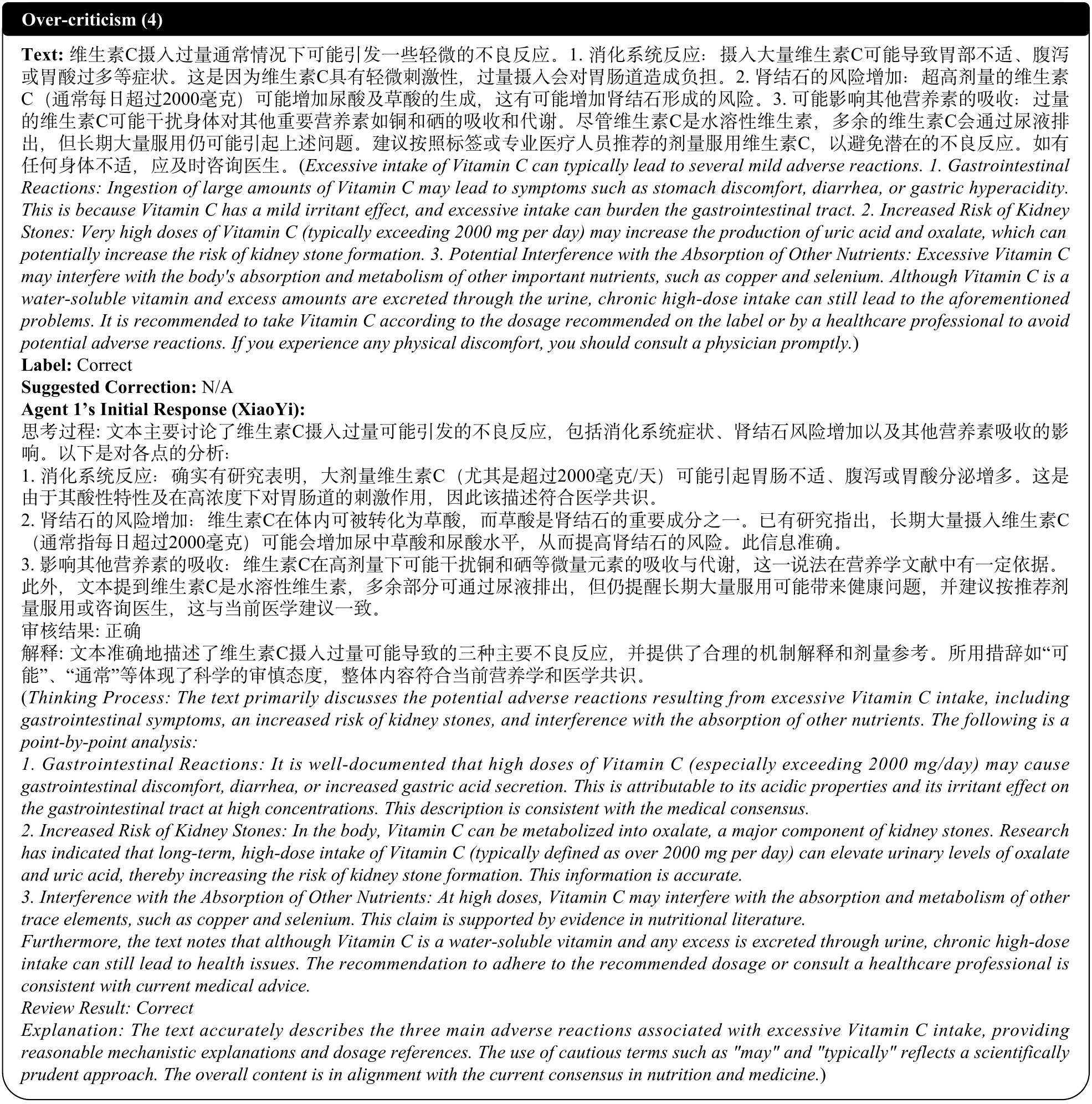}
\end{figure*}

\begin{figure*}[!t]
    \centering
    \includegraphics[width=1\linewidth]{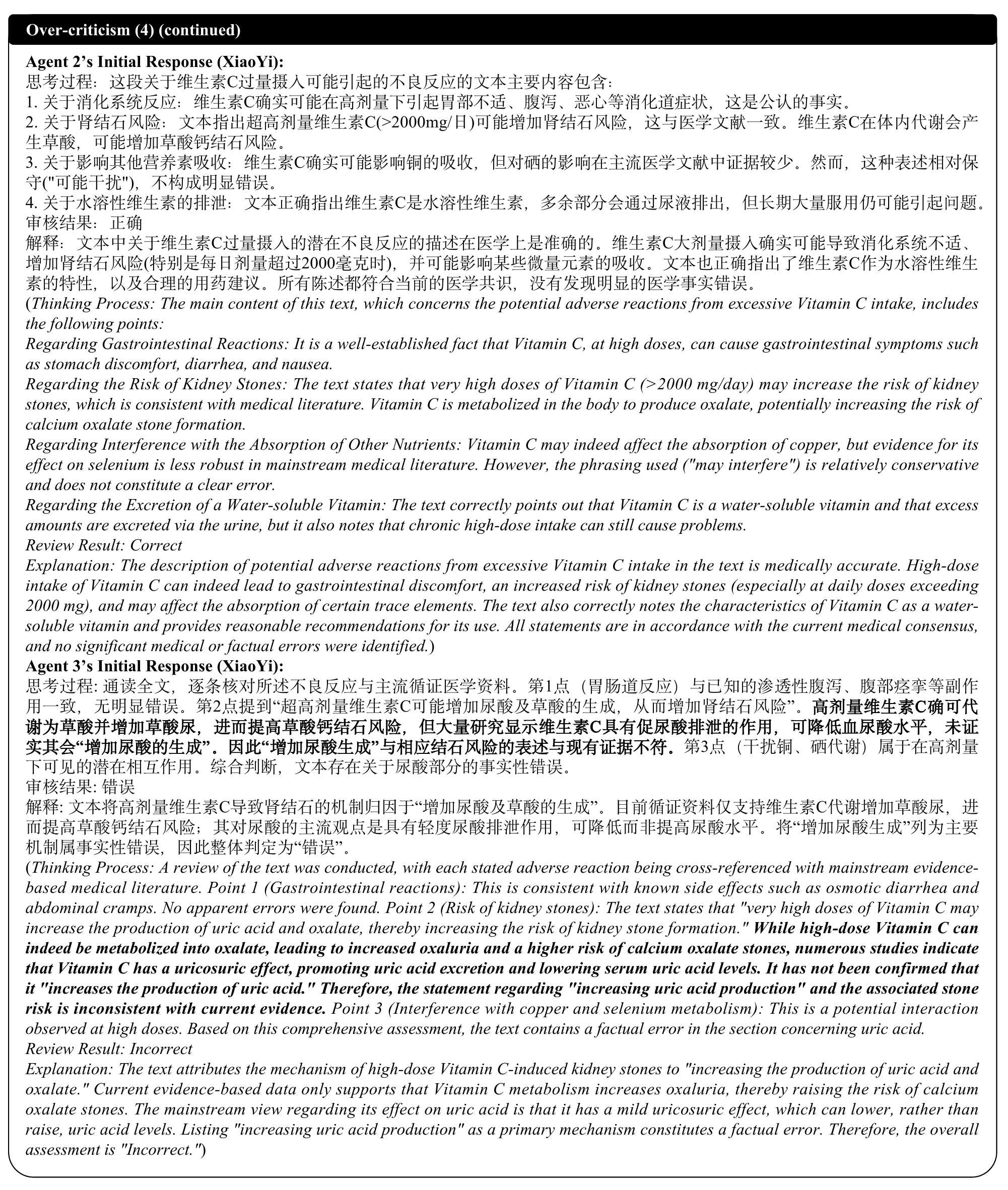}
\end{figure*}

